\documentclass[acmsmall,nonacm]{acmart}
\renewcommand\footnotetextcopyrightpermission[1]{} % removes footnote with conference information in first column
\usepackage{booktabs}   %% For formal tables:
\usepackage{subcaption} %% For complex figures with subfigures/subcaptions
\usepackage{booktabs} % For formal tables
\PassOptionsToPackage{usenames, svgnames}{xcolor}

\usepackage{url}
\usepackage{bbm}

\usepackage{amsmath, amsthm, amsfonts}
\usepackage{thmtools}
\usepackage{mathtools}

\usepackage{graphicx}
\usepackage[hypcap=true]{subcaption}
\usepackage{tcolorbox}
\usepackage{adjustbox}

\usepackage{multirow}
\usepackage{longtable}

\usepackage[inline]{enumitem}
\usepackage{booktabs}
\usepackage[super]{nth}
\usepackage{bm}
\usepackage{pifont}

\usepackage{relsize, scalerel}
\usepackage{twoopt, etoolbox}
\usepackage{setspace}
\usepackage{xspace}
\usepackage{hyphenat}
\usepackage[xspace]{ellipsis}
\usepackage{listings}
\usepackage{lipsum}% http://ctan.org/pkg/lipsum

\usepackage{tikz}
\usetikzlibrary{fit, positioning, arrows.meta, calc}

\usepackage{algorithm}
\usepackage[noend]{algpseudocode}
\usepackage{listings}
\usepackage{varwidth}
\usepackage{hhline}
\newcommand{\keywordc}[1]{{\color[rgb]{0.8,0.3,0.1} #1}}

\newcommand{\stc}[1]{{\color[rgb]{0.1,0.1,0.7} #1}}
\lstalias[]{csharp}[Sharp]{C}
\lstdefinelanguage{dsl}{
	morekeywords = {language,feature,using,semantics,learners,int,string,@input,@start,values,let,in,@ref,@values,Tuple,@extern,@output,namespace,bool,std,@id,grammar},
	otherkeywords = {:=,=>,:,[],=},
	sensitive = true,
	morecomment = [l]{//},
	morestring = [b]',
}
\definecolor{linenum-gray}{HTML}{888888}
\renewcommand{\algorithmiccomment}[1]{ {\itshape \color[rgb]{0,0.6,0} \textsc{//} #1}}
\newcommand{\NoNumber}{\def\alglinenumber##1{}}
\newcommand{\WithNumber}{\def\alglinenumber##1{\sf\scriptsize\color{linenum-gray}##1:\hspace{-11pt}}}
\newcommand{\Functionx}[2]{\NoNumber \Function{#1}{#2} \addtocounter{ALG@line}{-1} \WithNumber}

\usepackage{url}
\theoremstyle{remark}

\theoremstyle{plain}
\newtheorem{defn}{Definition}
\newtheorem{lem}{Lemma}
\newtheorem{thm}{Theorem}
\newtheorem{example}{Example}
\newtheorem{remark}{Remark}

\newcommand{\imp}{\textsc{Imp}}

\newcommand{\implin}{\textsc{Linear}}
\newcommand{\impconst}{\textsc{Const}}
\newcommand{\lang}{\mathcal{L}}
\newcommand{\return}{\textbf{\textsf{return}}}

\newcommand{\ifc}{\textbf{\textsf{if}}}
\newcommand{\thenc}{\textbf{\textsf{then}}}
\newcommand{\elsec}{\textbf{\textsf{else}}}
\newcommand{\selftune}{\ensuremath{\mathsf{PBR}}\xspace}
\newcommand{\sketch}{\ensuremath{\mathsf{Sketch}}\xspace}
\newcommand{\Nevergrad}{\ensuremath{\mathsf{Nevergrad}}\xspace}
\newcommand{\decisionservice}{\ensuremath{\mathsf{DecisionService}}\xspace}
\newcommand{\smartchoices}{\ensuremath{\mathsf{SmartChoices}}\xspace}
\newcommand{\prose}{\ensuremath{\mathsf{PROSE}}\xspace}

\newcommand{\good}[1]{{\textbf{\color[rgb]{0,0.6,0}#1}}}

\newcommand{\bad}[1]{{\textbf{\color[rgb]{0.6,0,0}#1}}}
\newcommand{\paramval}{\ensuremath{a}\xspace}
\newcommand{\paramvals}{\ensuremath{\mathbf{\paramval}}\xspace}
\newcommand{\featval}{\ensuremath{x}\xspace}
\newcommand{\featvals}{\ensuremath{\mathbf{\featval}}\xspace}
\newcommand{\x}{\ensuremath{\mathbf{\featval}}\xspace}
\newcommand{\z}{\ensuremath{\mathbf{z}}\xspace}
\newcommand{\weight}{\ensuremath{w}\xspace}
\newcommand{\weights}{\ensuremath{\mathbf{\weight}}\xspace}
\newcommand{\w}{\ensuremath{\mathbf{\weight}}\xspace}
\newcommand{\numfeat}{\ensuremath{d}\xspace}
\newcommand{\numweights}{\ensuremath{d}\xspace}
\newcommand{\numparam}{\ensuremath{m}\xspace}

\newcommand{\pred}{\ensuremath{f}\xspace}
\newcommand{\numcontext}{\ensuremath{p}\xspace}

\newcommand{\Fermat}{\textsc{Fermat}\xspace}
\newcommand{\Thermostat}{\texttt{Thermostat}\xspace}
\newcommand{\Aircraft}{\texttt{Aircraft}\xspace}

\newcommand{\assignr}{\texttt{PBR\textbf{.AssignReward}}}
\newcommand{\decfun}{\texttt{PBR\textbf{.DecisionFunction}}}
\newcommand{\Create}{\texttt{\textbf{Create}}\xspace}
\newcommand{\Connect}{\texttt{\textbf{Connect}}\xspace}
\newcommand{\Predict}{\texttt{\textbf{Predict}}\xspace}
\newcommand{\GetExprTree}{\texttt{\textbf{GetExprTree}}\xspace}
\newcommand{\Refresh}{\texttt{\textbf{Refresh}}\xspace}

\newcommand{\treemodel}{\mathbf{W}, \boldsymbol{\Theta}}
\newcommand{\treemodelopt}{\mathbf{W}^*, \boldsymbol{\Theta}^*}
\newcommand{\treemodelW}{\mathbf{W}}
\newcommand{\treemodelTheta}{\boldsymbol{\Theta}}
\newcommand{\btheta}{\boldsymbol{\theta}}

\newcommand{\ent}{\textsc{EntropyNet}}
\newcommand{\entdesc}{\text{entropy net}}
\newcommand{\fent}{f_{\emph{\textsc{Net}}}}

\newcommand{\netmodel}{\ensuremath{\widetilde{\mathbf{W}}}}
\newcommand{\netmodelw}{\ensuremath{\widetilde{\mathbf{w}}}}

\newcommand{\arch}{Shallow Nets}

\newcommand{\netw}{\widetilde{\mathbf{w}}}
\usepackage{xcolor}
\begin{document}
	
	%% Title information
	\title{Programming by Rewards}
	\subtitle{Synthesizing programs using black-box rewards}

%	%% when present, will be used in
%	%% header instead of Full Title.
%	\titlenote{with title note}             %% \titlenote is optional;
%	%% can be repeated if necessary;
%	%% contents suppressed with 'anonymous'
%	\subtitle{Subtitle}                     %% \subtitle is optional
%	\subtitlenote{with subtitle note}       %% \subtitlenote is optional;
%	%% can be repeated if necessary;
%	%% contents suppressed with 'anonymous'

	%% Author information
	%% Contents and number of authors suppressed with 'anonymous'.
	%% Each author should be introduced by \author, followed by
	%% \authornote (optional), \orcid (optional), \affiliation, and
	%% \email.
	%% An author may have multiple affiliations and/or emails; repeat the
	%% appropriate command.
	%% Many elements are not rendered, but should be provided for metadata
	%% extraction tools.

\author{Nagarajan Natarajan} 
\affiliation{\institution{Microsoft Research} \country{IN}} 
\email{nagarajn@microsoft.com}
\author{Ajaykrishna Karthikeyan} 
\affiliation{\institution{Microsoft Research} \country{IN}} 
\email{t-ajka@microsoft.com}
\author{Prateek Jain} 
\affiliation{\institution{Microsoft Research} \country{IN}} 
\email{prajain@microsoft.com}
\author{Ivan Radi{\v{c}}ek} 
\affiliation{\institution{Microsoft} \country{Austria}} 
\email{ivradice@microsoft.com}
\author{Sriram Rajamani} 
\affiliation{\institution{Microsoft Research} \country{IN}} 
\email{sriram@microsoft.com}
\author{Sumit Gulwani} 
\affiliation{\institution{Microsoft} \country{USA}} 
\email{sumitg@microsoft.com}
\author{Johannes Gehrke}
\affiliation{\institution{Microsoft Research} \country{USA}} 
\email{johannes@microsoft.com}

% \ccsdesc[300]{Computer systems organization~Redundancy}
% \ccsdesc{Computer systems organization~Robotics}
% \ccsdesc[100]{Networks~Network reliability}

%%
%% Keywords. The author(s) should pick words that accurately describe
%% the work being presented. Separate the keywords with commas.

%% A "teaser" image appears between the author and affiliation
%% information and the body of the document, and typically spans the
%% page.
% \begin{teaserfigure}
%   \includegraphics[width=\textwidth]{sampleteaser}
%   \caption{Seattle Mariners at Spring Training, 2010.}
%   \Description{Enjoying the baseball game from the third-base
%   seats. Ichiro Suzuki preparing to bat.}
%   \label{fig:teaser}
% \end{teaserfigure}

%%
%% This command processes the author and affiliation and title
%% information and builds the first part of the formatted document.
%
\begin{abstract}
We formalize and study ``programming by rewards'' (PBR), a new approach for specifying and synthesizing subroutines for optimizing some quantitative metric such as performance, resource utilization, or correctness over a benchmark. A PBR specification consists of (1) input features $\x$, and (2) a reward function $r$, modeled as a black-box component (which we can only run), that assigns a reward for each execution. The goal of the synthesizer is to synthesize a {\em decision function} $f$ which transforms the features to a decision value for the black-box component so as to maximize the expected reward 
$E[r \circ f (\x)]$ for executing decisions $f(\x)$ for various values of $\x$.

We consider a space of decision functions in a DSL of loop-free if-then-else programs, which can branch on linear functions of the input features in a tree-structure and compute a linear function of the inputs in the leaves of the tree. We find that this DSL captures decision functions that are manually written in practice by programmers.  Our technical contribution is the use of continuous-optimization techniques to perform synthesis of such decision functions as if-then-else programs. We also show that the framework is theoretically-founded ---in cases when the rewards satisfy nice properties, the synthesized code is optimal in a precise sense. 

PBR hits a sweet-spot between program synthesis techniques that require the entire system $r \circ f$ as a white-box, and reinforcement learning (RL) techniques that   treat the entire system $r \circ f$ as a black-box. PBR takes a middle path treating $f$ as a white-box, thereby exploiting the structure of $f$ to get better accuracy and faster convergence, and treating $r$ as a black-box, thereby scaling to large real-world systems. Our algorithms are provably more accurate and sample efficient than existing synthesis-based and reinforcement learning-based techniques under certain assumptions. 

We have leveraged PBR to synthesize non-trivial decision functions related to search and ranking heuristics in the PROSE codebase (an industrial strength program synthesis framework) and achieve competitive results to manually written procedures over multiple man years of tuning. We present empirical evaluation against other baseline techniques over real-world case studies (including PROSE) as well on simple synthetic benchmarks. 
%We also demonstrate these advantages empirically by evaluating our implementation of PBR and comparing it with both program synthesis and RL approaches on both simple synthetic examples as well as real-word examples.
%jg01
%existing systems like sketching~\cite{sketchtool} and decision service~\cite{agarwal2016multiworld}. 
%Software and services are controlled by many parameters, which are either hard-coded in configuration files, or set using custom logic in manually written decision functions in the code. PBR allows developers to automatically synthesize decision functions that transform input features to output decisions. The output decisions are then fed into a black-box component, which can be arbitrarily complicated. The execution of the black-box can then be observed by a rewards module, which gives a real valued reward for the decision taken. The goal of PBR is to synthesize a decision function that maximizes the expected reward.  

\end{abstract}

\maketitle
\thispagestyle{empty}

\section{Introduction}
\label{sect:intro}
%Program synthesis systems typically rely on kinds of specifications that humans can naturally provide. Certain types of specification are natural to specific domains, and consequently the program synthesis systems work with domain-specific languages. In domains like data formatting/processing, it is natural to convey intent in the form of enumerative input-output examples. In the information retrieval domain  (e.g. SQL queries),  often it is desirable to specify natural language input. In the program verification domain, the spec is often succinct and encodes some notion/test of satisfiability. All these systems take in a spec and produce some valid program that can carry out the specific user-intended task.

Consider the following scenario, which routinely arises while writing software. A developer wants to write a sub-routine to decide how to set some threshold parameter, such as timeout value, before executing a software component (say a database system or a networking system). First, they may not apriori know what the threshold needs to be for a particular input --- because fundamentally there may not be any "right" threshold for a given input, but it may depend on the myriad program variables in complex ways that eventually affects the execution of the software. Suppose the developer makes a decision to set the threshold to some value $\tau$. After the component finishes executing, the developer may be able to measure some non-functional metric such as latency or throughput or resource utilization to get feedback on whether the threshold $\tau$ was a "good" or "bad" choice. Such feedback can be given as a "reward" (using the terminology of reinforcement learning) to improve future choices for the threshold $\tau$.

The performance and functionality of large scale software is dependent on many such thresholds, which we call as {\em decision values} or more succinctly, {\em decisions}.  Typically, decisions are tuned to suitable values depending on the variables that represent the state of the program (such as size of internal queues) as well as the state of the environment (such as number of requests received per second), and the number of such dependencies can be very large, or "high-dimensional", to use terminology of machine learning. Often, such decisions are set in the code, using custom logic, as shown in the example code in Figure~\ref{fig:code-samples-prose}. We call functions with such custom logic as {\em decision functions}. Decision functions also generalize configuration files~(see Figure~\ref{fig:code-samples-prose} (left)), which is often part of large code-bases. Each line in the configuration file can be thought of as trivial decision functions that return constant values as decisions. 

%The above examples motivate a new perspective/role program synthesis systems assume in modern software development --- (a) often the developers want to generate code for a function/subroutine that they find hard-pressed to write manually, rather than generating the end-to-end code that can carry out the entire task (think of big codebases, and not simple self-contained programs like SQL queries), (b) traditional kinds of specification fall woefully short -- and in many cases infeasible --- in these scenarios.

We propose a new framework, \textit{Programming by Rewards}, abbreviated as {\em PBR}, for automatically synthesizing and tuning decision functions in standard settings. A PBR specification consists of
\begin{enumerate}
    \item
    The input features $x$, with their data types, and the decision type. For example, in Figure~\ref{fig:code-samples-prose}, for the decision function {\tt ScoreLinesMap} the input features are {\tt selection} and {\tt lines}, both having type {\em double}, and the decision type is also {\em double}, which is the return type of the method.
    \item
    A reward function $r$, modeled as a black-box component, that consumes the output of the decision function, executes an arbitrarily complicated software module, and assigns a  reward value for the execution of the black-box software module with the provided decision value. For example, in Figure~\ref{fig:code-samples-prose}, the black-box is the PROSE engine~\cite{prosesdk} which takes the return values of {\tt ScoreLinesMap} and other such decision functions as decision values, and executes a complicated program synthesis engine, and assigns a reward like total execution time or accuracy of the synthesized program. Other examples of black-box engines could be communications software such as Skype or Zoom, or database engines such as SQLServer, where the reward value can be proportional to latency seen by the end-user or a combination of total execution time. % as well as the quality of the output produced in the case of PROSE.
\end{enumerate}
 The framework  relies on suitable programmer-defined rewards that indicate how well the decision value returned by the decision function eventually affects the success of the overarching software itself, represented by the the reward function $r$. We formalize the problem of synthesizing decision functions given only execution (or invocation) access to the reward function $r$. Furthermore, in practice, due to changes in the environment (such as load on the system), two invocations of the black-box with the same decision value can result in different reward values. Hence, $r$ need not be deterministic.  We optimize the expected value of the reward (see Definition 3 in Section 2), while allowing for randomness inside the reward function.
 
Prior work in this area follows one of the three approaches:
\begin{enumerate}
\item In the \textbf{rule-based approach}, which is widely used by practitioners, the programmer writes decision functions using custom code, which has domain-specific logic to compute decision values in terms of input features.
For example, the decision function ~\texttt{ScoreLinesMap} shown in Figure~\ref{fig:code-samples-prose} uses three manually chosen parameter values {\tt minscore,alpha} and {\tt beta}, to
compute one decision value as a function of the input features {\tt sel} and {\tt lines}, which is returned by the function. The programmer can tweak the custom code based on a few observed rewards, but in general, setting a larger number of parameters manually can lead to significantly sub-optimal rewards.  Further, the decision function ~\texttt{ScoreLinesMap} and the parameter values do not adapt automatically as the environment of the software changes, which is undesirable.
\item In the \textbf{sketch-based synthesis approach}, the idea is to specify the decision function as a template with ``holes''. For example, in the decision function~\texttt{ScoreLinesMap} shown in Figure~\ref{fig:code-samples-prose}, the programmer  can leave the values of the parameters {\tt minScore}, {\tt alpha} and {\tt beta} as holes and leave it to the synthesis engine to synthesize values for these holes such that a quantitative specification that optimizes the reward value is satisfied. Though Sketching has been primarily used with correctness specifications (which are Boolean) \cite{sketchtool}, prior work has explored sketching to optimize quantitative specifications such as reward values~\cite{chaudhuri2014bridging}. In this setting, the entire software $r \circ f$ is required to be white-box for program synthesis to work, which makes the approach challenging to scale.
\item In the \textbf{reinforcement learning approach}, the decision function is learned automatically using an ML formulation, which is trained from traces of executions of the code so as to optimize the expected value of the rewards \cite{agarwal2016multiworld,sutton2011reinforcement}. This approach has attracted much attention recently due to increasing popularity of machine learning. In this setting, the entire software $r \circ f$ is treated as a black-box, and the algorithms to learn $f$ need a large number of samples to learn when the number of features and parameters are large.
\end{enumerate}
PBR hits a sweet-spot between program synthesis and reinforcement learning approaches mentioned above. PBR takes a middle path, treating the decision code $f$ as a white-box, thereby exploiting the structure of $f$ to get better accuracy and faster convergence, and treating $r$ as a black-box, thereby scaling to real-world systems.
We consider a space of decision functions in a DSL of loop-free if-then-else programs, which can branch on the input features in a tree-structure and compute a linear function of the inputs in the leaves of the tree. We find that this DSL captures decision functions that are manually written in practice by programmers.

%So at a high level, the programmer provides a sketch from a if-then-else programs based DSL, which implies that it can instantiate some part of the program, and can leave some "holes" to be filled by the PBR system. Then, programmer sets a reward function which assigns goodness metric to the output decision values from the sketch, and then invoke our system to automatically learn the parameters/holes in the sketch. 
Our methods work in both the {\em offline} and {\em online} settings.
In the offline setting we have access to all the executions and reward values apriori. In the online setting, a reward is assigned online in response to a decision value, and PBR uses the reward to improve the decision function. In practical large-scale systems, online setting is more practical and general, so we mostly focus on this setting from algorithm development viewpoint. Our most substantial case study (Section ~\ref{sec:prose}) is also in the context of black-box reward functions in an online setting.
However,  we also compare with baselines (especially based on sketch-based synthesis methods) having offline access all the data, and having white-box access to the reward functions (Section ~\ref{sec:evaluation_sketch}).

Our key technical contribution is the use of continuous-optimization techniques, specifically gradient descent, to perform synthesis of  decision functions such as if-then-else programs, with small {\em update} and {\em sample} complexity, i.e., the execution time as well as the number of black-box reward executions required by our techniques are relatively small.  However, there are several challenges in applying gradient descent techniques in PBR setting. First, we only have restricted access to the reward function $r$ so computing gradient itself is challenging. Furthermore, the general decision tree functions are non-smooth, piece-wise linear functions which are challenging to capture with continuous optimization methods. 

In this work, we make two-fold contributions on this front. First, we build upon well-established approach in the optimization literature, that gradient based methods allow inexact, noisy but {\em unbiased} gradient estimates, which can be obtained by invoking reward function $r$ on random perturbation of decision values. However, such methods can lead to large sample complexity (i.e. number of executions of $r$) if we have a large number of parameters. If the decision function is a linear function or if it is an if-then-else program with {\em small} number of decision values, then we can provide significantly more efficient gradient estimation methods. 
%Hence, we need to estimate gradients using only one invocation to the reward function $r$. We build on the randomized gradient descent algorithm proposed by~\cite{flaxman2005online} in the context of online learning with bandit feedback, which estimates gradients by random perturbations on the decision value. Our technical insight is that this idea of perturbations can be generalized to linear functions as well as decision trees. If the decision function is a linear model, then we show that applying the random perturbation to the the result of the linear transformation on the inputs is provably more efficient than perturbing the inputs themselves. If the decision function is a tree, then there are additional challenges due to non-smooth and piece-wise linear nature of such functions. We propose a differentiable and shallow neural network model, and show that under some structural constraints, it implicitly represents decision trees. Then, we learn the parameters of the shallow neural network using gradient descent, with the perturbation trick. We also show that the framework is theoretically-founded ---in cases when the rewards satisfy nice properties, the synthesized code is optimal in a precise sense.
%\jg{I suggest introducing $m$ and $d$ already earlier so that we can point out (with citation) that previous work
%has complexity based on $m$ -- and why that is bad?}
%In this paper, we propose an operator called \selftune, which combines the advantages of rule based and data-driven approaches to produce a robust and adaptive solution to this problem.
Formally, let $d$ be the number of parameters in the system, and let $m$ be the number of decisions made in the system. Our core result is that we can automatically synthesize decision functions 
%jg01: Not clear what we are improving over here -- thus I commented this out
%with improved efficiency 
if they are loop-free and use linear operators.
Specifically, even though there may be many parameters in the system ($d \gg m$), we design ML 
algorithms with complexity proportional to $m$ rather than $d$. 
%\jg{We did not say which previous work has complexity based on $m$.}
Since $m$ is typically much smaller than $d$, our approach has significant
advantages over existing approaches, whose complexity scales with $d$.
As an example, the decision function {\tt ScoreLinesMap} (in Figure~\ref{fig:code-samples-prose}) is a linear model with $m = 1$ and $d = 3$.
The decision function {\tt IsLikelyDataRatio} (in Figure~\ref{fig:code-samples-prose}) can be encoded as a decision tree model with (on the order of) 10 parameters
representing the weights and predicates in a shallow tree. In this case, we have a decision tree with $d\sim 10$ and $m=1$.
In this work, we propose novel ML algorithms whose complexity depends on $m$ rather than $d$ for linear models and decision trees.
Furthermore, for certain cases, we provide rigorous bounds on the efficiency of the proposed methods.  

Second, we show how decision trees can be modeled using a continuous shallow network model under structural constraints, which enables usage of standard gradient descent type of methods for learning the parameters of decision trees with only black-box reward $r$ (see Section~\ref{sec:learning_tree}). 

Finally, we conduct extensive experiments to validate that our PBR based methods can indeed be used to efficiently synthesize programs in real-world codebases using only black-box reward function (Section~\ref{sec:prose}). In particular, for \prose \cite{prosesdk}, which is a complicated system with nearly 70 ranking related heuristics fine-tuned over several years, our method can synthesize all the heuristics while achieving competitive accuracy to hand-tuned systems, after training for a few days and after only about 250 calls to the reward function; in \prose each reward function call requires synthesizing programs for about $740$ benchmarks and then computing their accuracy, thus highlighting the need to optimize sample complexity of PBR methods. Furthermore, we show that standard reinforcement style learning methods when used in completely black-box manner indeed suffer from poorer sample complexity in such problems (Section~\ref{sec:structure}). Finally, we observe that even when we provide existing sketch based methods--either using CEGIS (Counterexample Based Inductive Synthesis) on top of SAT \cite{sketchtool} or numerical methods \cite{chaudhuri2014bridging}--with whitebox access to the entire reward function and also provide apriori access to all the execution traces (i.e. offline data), their computational and sample complexity is significantly higher than our techniques (Section~\ref{sec:evaluation_sketch}). 

In summary, the paper makes the following contributions:
\begin{enumerate}
    \item We observe \imp~(Definition~\ref{def:imp}) to be rich language that accurately models typical decision functions in practice and also identify some important sub-classes of \imp. 
    \item We formalize Programming by Rewards (PBR) with \imp~DSL and also present an equivalent formalism as that of learning decision trees with black-box rewards. 
    \item  We present novel algorithms for learning decision tree as well as linear models with rewards and provide rigorous guarantees for the latter. 
    \item We present strong empirical validation of our approach on an industrial strength codebase, demonstrating both sample and computational efficiency. 
\end{enumerate}
{\bf Paper Organization}: Section~\ref{sec:problem} formally specifies the PBR problem, provides the DSL that we consider and motivates it using existing work and inspection of a few real-world codebases. Then in Section~\ref{sec:equiv_tree}, we provide overview of how we set up the problem as a decision-tree learning problem and set up the key metrics to be considered while designing the algorithms. In Section~\ref{sec:learning} we provide specific PBR algorithms for three different DSLs and in certain cases, provide rigorous guarantees. In Section~\ref{sec:evaluation} we provide empirical evaluation of our method and compare it against relevant sketching and reinforcement learning based baselines. In Section~\ref{sec:related}, we survey related work, and finally conclude with Section~\ref{sec:conclusions}. 
%We find that existing approaches are unsuitable for learning such decision functions. Program synthesis approaches, such as Sketch~\cite{sketchtool}, attempts to fill in the "holes" in the templates (corresponding to parameters), by using combinatorial search or SAT solving methods, which do not scale for a large number of parameter and complex software black-boxes. 
%Reinforcement Learning~\cite{sutton2011reinforcement} techniques currently do not exploit the structure in our templates and tend to have high sample complexity.  
%The empirical results in Section 5 confirm the above mentioned challenges with existing techniques and also demonstrate effectiveness of our solution. 
%\begin{figure}
%\begin{minipage}[.49\textwidth]

\begin{figure}[t!]
  \begin{subfigure}[b]{.3\linewidth}
	\begin{lstlisting}[language=dsl,morekeywords={Resource1,Resource2},basicstyle=\footnotesize\ttfamily]
	[Resource1]
    RefreshInterval=00:01:00
    PriorityHighUnderloaded=97
    PriorityHighOverloaded=98
    PriorityLowUnderloaded=88
    PriorityLowOverloaded=90
    ...
	[Resource2]
    RefreshInterval=00:01:00
    PriorityHighUnderloaded=80
    PriorityHighOverloaded=90
    ...
    \end{lstlisting}
  \end{subfigure}
  \begin{subfigure}[b]{.6\linewidth}
	\begin{lstlisting}[language=dsl,morekeywords={public,static,double,return},basicstyle=\footnotesize\ttfamily]
    double ScoreLinesMap(double sel, double lines) {
        double minScore = 100.0;
        double alpha = 1.0; double beta = 1.0;
        return alpha * sel + beta * lines + minScore * 20;
    }
    bool IsLikelyDataRatio(int dataCount, int totalCount) {
        if (totalCount < 10) return dataCount >= 6;
        if (totalCount < 20) return dataCount >= 15;
        if (totalCount < 50) return dataCount >= 30;
        return dataCount / (double) totalCount >= 0.6;
    }
    \end{lstlisting}  \end{subfigure}\vspace*{-5pt}
    \caption{(left) A configuration file is a common coding pattern for hard-coding constants/settings in software, and (right) Implementations of a rule for scoring programs, and a control logic for determining the value of a boolean state variable; both part of the widely-deployed program synthesis SDK, \prose~\cite{prosesdk}.}
  \label{fig:code-samples-prose}
\end{figure}

\section{Problem Formulation}
\label{sec:problem}
This section formally introduces Programming by Rewards (PBR) problem for synthesizing {\em decision functions}. We start by
motivating and defining the class of decision functions considered in this work. We set up the PBR problem with an example, and discuss how it relates to the standard sketch-based synthesis problem. Next, we give an equivalent formulation of the problem using decision trees. This lets us formulate the problem as a learning problem which can leverage continuous-optimization based techniques and provide significantly more efficient algorithms. 

\subsection{Decision functions in real-world software}
We motivate our approach by showing various examples of decision functions that are currently written manually by programmers. We performed extensive studies
of such decision functions in two domains: 
\begin{itemize}
% \item{ \bf Efficient search and ranking.} There are many problem domains where we need to search through the space of solutions in a combinatorial space and return a ranked list of preferred solutions. This includes SAT and SMT solvers, as well as program synthesis engines. Software written for such domains often involve several heuristic decisions made in the code. We study decision functions from  PROSE~\cite{prosesdk}, which is an industrial strength program synthesis system. In PROSE, the search and ranking heuristics right are designed to optimize efficiency of synthesis and accuracy of the returned solution as two metrics. A significant fraction of developer time is spent on hand-crafting these heuristics for each new domain where PROSE is applied. Each new domain is characterized by a different DSL, and the nature of the DSL and the programs we seek determines which production rules we want to favor, which is in turn determined by the decision functions. The example decision function in Figure~\ref{fig:code-samples-prose} is from real-world ingestion software built on top of PROSE~\cite{prosesdk}. The function implements a heuristic for deciding header to data ratio in the input file; the examples in Figures~\ref{fig:code-regexpair} and~\ref{fig:code-formatdatetimerange} are ranking heuristics for spreadsheet processing widely used in commercial software products today. We observe that these decision functions either involve simple (linear) operations on the input variables, or compute decisions based simple branching decisions on the input variables.
\item{ \bf Efficient search.} There are many problem domains where we need to search through the space of solutions in a combinatorial space. This includes SAT and SMT solvers, as well as program verification and synthesis engines. Software written for such domains often involve several heuristic decisions made in the code. Figure~\ref{fig:search} shows a snippet of a hybrid branching heuristic implemented in the \textsf{MapleGlucose} SAT solver~\cite{liang2016mapleglucose}. We notice the rather obscure choice of constants and thresholds in the conditionals as part of the heuristic. 

\begin{figure}[t!]
	\begin{lstlisting}[escapechar=!,language=csh,morekeywords={public,static,double,return,string,foreach,void,bool,int},basicstyle=\footnotesize\ttfamily]
    bool Solver::search(int nof_conflicts)
    {
        ...
        for (;;) {
             ... 
             if (S > !\colorbox{yellow}{0.06}!)
                S -= !\colorbox{yellow}{0.000001}!; 
             if(conflicts % !\colorbox{yellow}{5000}! == 0 && var_decay < !\colorbox{yellow}{0.95}!)
                var_decay += !\colorbox{yellow}{0.01}!;
             ...
        }
    }
    \end{lstlisting}\vspace*{-5pt}
	\caption{Code snippet from \textsf{MapleGlucose} SAT solver~\cite{liang2016mapleglucose} that implements a hybrid LRB-VSIDS branching heuristic in the search procedure.}
	\label{fig:search}
\end{figure}

\item{ \bf Ranking heuristics.} Program synthesis and in particular programming-by-example engines, e.g., PROSE (which has found adoption in multiple mass-market products), need to deal with ambiguity in the user's intent expressed using a few input-output examples. This is done by carefully implementing ranking heuristics to select an intended program from among the many that satisfy the few input-output examples provided by the user. A significant fraction of developer time is spent on hand-crafting these heuristics for each new domain where PROSE is applied. Each new domain is characterized by a different DSL, and the nature of the DSL and the programs we seek determines which production rules we want to favor, which is in turn determined by the decision functions. We study decision functions from PROSE, which is an industrial strength program synthesis system. The example decision functions in Figure~\ref{fig:code-samples-prose} is from real-world ingestion software built on top of PROSE. The function implements a heuristic for deciding header to data ratio in the input file; the examples in Figures~\ref{fig:code-regexpair} and~\ref{fig:code-formatdatetimerange} are ranking heuristics for example-based string transformations widely used inside a mass-market spreadsheet product. We observe that these decision functions either involve simple (linear) operations on the input variables, or compute decisions based simple branching decisions on the input variables.
\item{\bf  Real-time services.} Large-scale services for workload management and scheduling, cloud database management servers (AzureSQL), etc. often have configuration settings and parameters to ensure optimal performance of the service in terms of latency, efficiency and cost. Often, these settings are defined in \textit{one-size-fits-all} manner relying on wisdom-of-the-crowd or other heuristics. Figure~\ref{fig:code-samples-prose} (left) shows the configuration snippet from a real-time scheduling service that is part of a mass-market commercial software system. These are the simplest type of decision functions, i.e. constants that one need to set optimally in a dynamic fashion to account for the continuously-changing system.  
\end{itemize}

\noindent
We observe that the decision function instances cited above, and more generally arising in these software domains, have the following characteristics:
\begin{enumerate}
    \item Operations are restricted to linear combinations of program variables; however, the parameters in the operation can be real numbers (expressed using floating point or decimal notation).
    \item There are nested if-then-else conditions, but the conditions are also expressible by linear combinations of input variables.
    \item There are no loops.
\end{enumerate}
This observation lets us define the following language that contains commonly-arising decision functions. Variants of this language have been used in program synthesis and in sketching~\cite{chaudhuri2014bridging,bornholt2016optimizing}.

\subsection{A DSL for decision functions}
\begin{defn} [Language \imp]
\label{def:imp}
We consider the language \imp~of programs with imperative updates and if-then-else statements without loops. Programs in \imp~operate over real valued variables; in particular, \imp~allows linear transformations over the input variables $x \in \mathbb{R}$ in both conditionals as well as assignments to local variables $o \in \mathbb{R}$. The function returns a tuple of local variables as decision values.
\begin{eqnarray*}
    P &::=& S; \return\ (o_1, o_2, \ldots, o_\numparam)\\
    S &::=& o = E \\
      &  & \big |~ \ifc~(E > 0)~\thenc~S_1~\elsec~S_2  \\
      &  & \big |~ S_1;S_2\\
    E & ::= &W_1 \cdot x_1 + W_2 \cdot x_2 +\cdots+W_\numcontext \cdot x_\numcontext+ W_{\numcontext+1} \\
    W & ::= & \mathbf{K} ~\big |~ ?? \\
\end{eqnarray*}
where~$\mathbf{K}$ is a numerical constant in $\mathbb{R}$, and "??" denotes a hole.
\end{defn}

Given the definition above, we can define decision functions formally as programs in \imp~ that take $\numcontext$ program variables as input and return $m$ decision values as output. It is easy to see that the functions in Figure~\ref{fig:code-samples-prose} are in \imp, with $\numcontext = 2$ and $m=1$. Furthermore, we observe that in some of the real-world settings, the decision functions that programmers write (including some of the aforementioned examples) \emph{do not} require the full expressiveness of~\imp, but can be even more concisely specified (as presented below) --- this facilitates provably-efficient algorithms in such settings (Section~\ref{sec:learning}).
\begin{defn} [Languages \implin~and~\impconst]~\\
%1. \textbf{\emph{$\imptree \subset \imp$}} consists of "tree functions" deciding on a single output value ($m = 1$), branching out on different states of the input program variables, using if-then-else style code; e.g. Figure~\ref{fig:code-samples-prose} (bottom), Figure~\ref{fig:DT} (left). \\
1. \textbf{\emph{$\implin \subset \imp$}} consists of linear functions with no if-then-else branches; e.g. Figure~\ref{fig:code-samples-prose} (right), Figures~\ref{fig:code-regexpair} and~\ref{fig:code-formatdatetimerange}.\\
2. \textbf{\emph{$\impconst \subset \implin$}} consists of constants, the most basic form of decisions, $W_1 = W_2 = \dots = W_\numcontext = 0$ in~\emph{\implin}; e.g. Figure~\ref{fig:code-samples-prose} (left).
\end{defn}
% \begin{remark} 
% While the above distinction between sub-languages of \imp\ may not make much of difference from the programmer's perspective as a user of PBR system, it turns out to be crucial in deriving provably sample-efficient algorithms, as we show in Section~\ref{sec:learning}.
% \end{remark}

\subsection{PBR formulation}
%Let $f \in \imp$ denote a decision function. In our PBR setup, the decision given by $f$ is fed into some black-box component $B$, which is a program in some arbitrary language $\lang$ (typically much more complex than $\imp$); we have no knowledge of $B$ or $\lang$, but we have black-box or invocation access to $B$. In other words, we can think of $B$ as the executable of the software that embeds the decision function of interest $f$ as a sub-routine. Finally, let us denote by $r$, a reward metric that quantifies some aspect of executing the black-box component $B$ using the decisions given by $f$. The reward may be assigned a value, which depends not only on correctness of the execution but also on nonfunctional aspects such as performance, latency, throughput of resource utilization. Formally, $r: B \circ f \mapsto \mathbb{R}_+$. The following example illustrates this set up.
Let $f \in \imp$ denote a decision function. In our PBR setup, the decision given by $f$ is fed into a black-box program (software) in some arbitrary language $\lang$ (typically much more complex than $\imp$); we have no knowledge of $\lang$. Let us denote by $r$, a reward metric that quantifies some aspect of executing the software using the decisions given by $f$. The reward may be assigned a value, which depends not only on correctness of the execution but also on nonfunctional aspects such as performance, latency, throughput of resource utilization. Formally, $r: f \mapsto \mathbb{R}$. The following example illustrates this set up.
\begin{example}
\label{ex:pbr}
\begin{figure}
	\begin{lstlisting}[language=dsl,morekeywords={public,static,double,return,string,foreach,void},basicstyle=\footnotesize\ttfamily]
    int ParseAndProcessFile(string fileName)
    {
        string contents = read(fileName);
        int[2] counts = getCounts(contents);
        if (PBR.DecisionFunction(PBRID_IsLikelyDataRatio, count[0], count[1]))
            preProcess(fileName);
        parseFile(FileName);
        return (processFile(fileName));
    }
    void main() 
    {
        double successes = 0.0;
        StopWatch sw = new StopWatch();
        sw.Start();
        foreach (string fileName in Config.Benchmarks):
            successes += ParseAndProcessFile(fileName) > 0 ? 1: 0;
        sw.Stop();
        double reward = (success/Config.Benchmarks.Length) - (sw.ElapsedMilliseconds/1000);
        PBR.AssignReward(reward);
    }
    \end{lstlisting}\vspace*{-5pt}
	\caption{A C\# program that uses the PBR system to automatically learn  a decision function in the place of the manually written function \texttt{IsLikelyDataRatio} from Figure~\ref{fig:code-samples-prose}.}
	\label{fig:synthex}
\end{figure}
Consider the program shown in Figure~\ref{fig:synthex}. Here, we show how to use the PBR system to automatically learn a decision function in place of the manually written function \texttt{IsLikelyDataRatio} from Figure~\ref{fig:code-samples-prose}.
The decision function learned by the PBR system is identified by a unique identifier {\tt PBRID\_IsLikelyDataRatio} for the function in the~\decfun~method. For each file, the function determines if we need to invoke an expensive pre-processing method before parsing the file.
A reward metric for the execution, which could potentially have multiple invocations of many decision functions, is assigned using the \assignr~method.

In this example, we have a single instance of a decision function invoked multiple times, once for each file.
The reward takes into account both the effectiveness of processing, which is the number of files processed successfully, and the efficiency of processing as measured by the time taken for processing all the files. Ideally, we want to balance the goals of successfully processing the files with spending less time on pre-processing. The goal of the PBR system is to learn a decision function which maximizes the expected reward.
\end{example} 

%begin{remark}1. Randomness in the rewards? (in features $\x$, in black-box $B$, ...) The above example doesn't bring it out. But we want to use that in the problem statement below.2. Also, it would be good to run through the entire synthesis flow with this example? like what do we mean by black-box access, as we have given a white-box program in the above Figure. With SelfTune API, the semantics was clear. No longer. How we iteratively synthesize 'f', perhaps compile the black-box program with the new 'f', invoke reward function, etc.\end{remark}

With this notation, we formally state the Programming By Rewards (PBR) problem.
\begin{defn}
[Programming By Rewards (PBR)] Given a specification consisting of:
\begin{enumerate}
    \item The input variables $\x \in \mathbb{R}^p$ to the decision function $f: \mathbb{R}^p \to \mathbb{R}^m$, with their data types, and the decision type,
    % \item
    % A black component $B$, which consumes the output of $f$ and executes an arbitrarily complicated software module, and
    % \item
    % Query-access to a black-box reward function $r: \mathbb{R}^m \to \mathbb{R}$, which assigns a reward value for the execution of the black-box $B$ with the provided decision values of $f$,
    \item
    Query-access to a black-box reward function $r: \mathbb{R}^m \to \mathbb{R}$, which assigns a reward value for the execution of the software with the provided decision values of $f$,
\end{enumerate}
the goal of PBR is to synthesize an \emph{optimal} $f \in \emph{\imp}$ such that:
% \begin{equation*}
%     f \in \arg \max_{f \in \emph{\imp}}~~\mathbb{E}[r(B \circ f)],
% \end{equation*}
\begin{equation*}
    f \in \arg \max_{f \in \emph{\imp}}~~\mathbb{E}[r(f)],
\end{equation*}
%where $\mathbb{E}$ denotes the expectation with respect to any randomness in the black-box component $B$ and $r$.
where $\mathbb{E}$ denotes the expectation with respect to the randomness in $r$.
\label{def:rgs}
\end{defn}
That is, the programmer provides input variables and reward function $r$, and expects a PBR system to synthesize the program from the \imp language (or a subset of \imp discussed in next section). Note that, our methods allow programmers to provide a sketch of the program from  \imp, and in general, that would lead to more efficiency. But as not providing any sketch is the extreme version of this problem, we focus on it   from our algorithmic development point of view, and leave further experiments on completing partial sketches via rewards for future work. 

We remark that the black-box reward functions $r$ can be expensive to execute (or even disruptive, in settings where the learning needs to happen in an online fashion in real-world deployments).  
Hence, the complexity of any algorithm to the PBR problem needs to be stated in terms of the total number of queries made to the
reward function to learn a "sufficiently good" solution---we refer to this as the {\bf sample complexity}.

The PBR problem can, in principle, be phrased as a program synthesis problem with the language \imp~being used to specify a sketch representing the space of programs we need to search for synthesizing decision functions.
However, such an approach runs into many difficulties.
The parameters we want to determine range over real values, which do not work well with enumerative search or symbolic methods using SMT solvers.
Further, the outputs of decision functions routinely feed into modules with millions of lines of code.
The reward function $r$ can be quantitative and the reward values can arrive \textit{much later in execution} after many large modules are executed with the decision values.
Consequently, if we want to model PBR using sketching, the whole program (which includes methods such as {\tt preProcess} and {\tt parseFile} in the example above) need to be modeled as a white-box. Moreover, it is unclear as to how concepts such as execution time can be modeled using sketching.

Aforementioned aspects distinguish our problem formulation from the standard sketch-based synthesis formulations. In particular, sketching techniques that take in boolean~\cite{sketchtool} or quantitative specification~\cite{chaudhuri2014bridging}, require complete knowledge of the program that contains the holes, and white-box access to some functional form of the reward function (which also needs to satisfy some constraints, and can not be arbitrary). Consequently, many practical instances of PBR can not be posed as sketches.

\section{Equivalence to Decision Tree Learning}\label{sec:equiv_tree}
\label{sec:equivalence}
We observe that the IMP programs are a form of {\em decision trees} with linear comparisons in nodes and linear computations in leaves.
Below, we establish the PBR learning problem as that of learning optimal decision trees in presence of a black-box reward function, and then describe algorithms for the latter. This allows us to make use of an elegant notation for describing our algorithms as well as makes our algorithms relevant to a broader community.

%We now state an equivalence between synthesizing decision functions from \imp\ and learning structures popularly known in machine learning parlance as decision trees. In the subsequent section, we will exploit this equivalence to derive algorithms for the PBR problem in Definition~\ref{def:rgs}. 

For ease of exposition, we will consider $m = 1$ case below (i.e. the decision function returns only one value); it is straight-forward to extend to the general setting. 

\begin{defn} [Decision Trees]
\label{def:DT}
A binary decision tree of height $h$ represents a piece-wise linear function $f_T(\x; \treemodel): \mathbb{R}^\numcontext \to \mathbb{R}$ parameterized by (a) weights $\treemodelW = \{ \weights_{ij} \in \mathbb{R}^\numcontext, b_{ij} \in \mathbb{R} \}$, with $\weights_{ij}$ at $(i,j)$th node ($j$th node at depth $i$) computing decisions of the form $\langle\weights_{ij}, \x\rangle + b_{ij} > 0$, and (b) $\treemodelTheta = \{ \btheta_j \in \mathbb{R}^\numcontext \}$, with parameters $\btheta_j \in \mathbb{R}^\numcontext$ that encode a linear model at each leaf node. The computational semantics of the function $f_T$ is as follows: An input $\x \in \mathbb{R}^\numcontext$ traverses a unique root-to-leaf path in the tree, based on the results of the binary decisions at each height, with the convention that the left branch is taken if the decision at the node is satisfied, or the right branch otherwise. The output of the decision tree is $\btheta_{*} \cdot \x$, where $*$ is the leaf node reached by $\x$ and $\cdot$ denotes the dot product. %The parameters of the tree, , $(\treemodel) = \big\{ (\w_{ij}, b_{ij}), i \in \{0,1,\dots,h-1\}, j \in \{0,1,\dots,2^i-1\} \big\} \cup \big\{\btheta_j, j \in \{0,1,\dots,2^h-1\}\big\}$. 
\end{defn}
\begin{remark}
The standard "regression tree" (decision tree for real-valued predictions) of height $h$ is a piece-wise constant function, i.e., $f_T$ realizes at most $2^h$ unique values (one per leaf node). We consider a more general and expressive tree, that represents a piece-wise linear function, as defined above.
\end{remark}

\begin{remark}
The above definition of decision tree can be extended to the general $m$ setting, by allowing the leaf node parameters to be $\btheta_j \in \mathbb{R}^{\numparam \times \numcontext}$; $\treemodelW$ remains the same as above. 
\end{remark}

\begin{figure}
  \begin{subfigure}[b]{.45\linewidth}
	\begin{lstlisting}[language=dsl,morekeywords={public,static,double,return,string,foreach,void},basicstyle=\footnotesize\ttfamily]
    double throttle(double latency, double load, double min)
    {
        double rate = 1.0;
        if (latency < 5) return load;
        if (load > 100)
            rate = 0.5;
        if (load * rate < min) return min;
        return load * rate;
    }   
    \end{lstlisting}
  \end{subfigure}
  \begin{subfigure}[b]{.45\linewidth}
    \centering
    \includegraphics[scale=0.3]{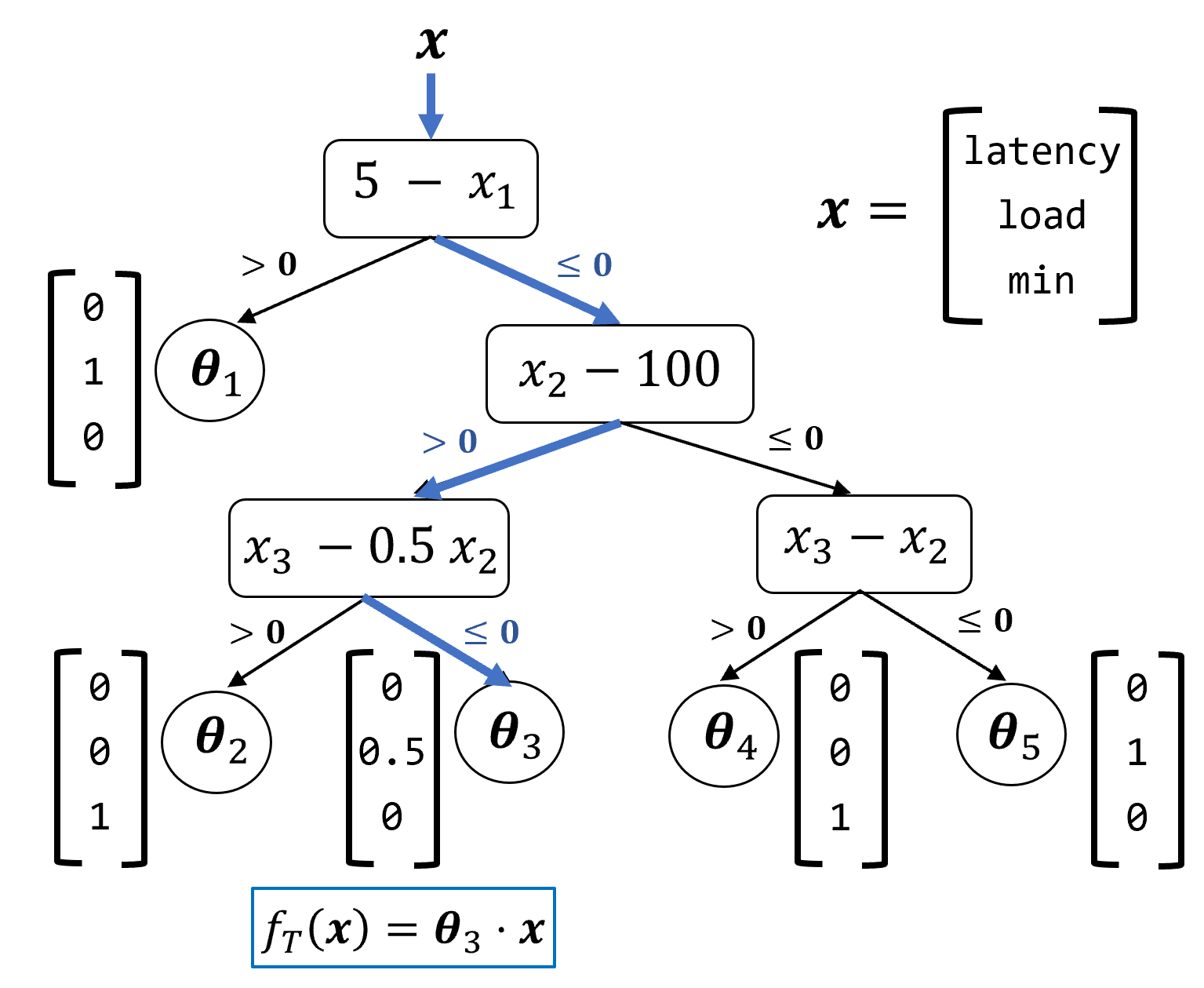}
  \end{subfigure}\vspace*{-5pt}
    \caption{(left) Example decision function in \imp, and (right) its equivalent binary decision tree with linear models in the leaf nodes.}
  \label{fig:DT}
\end{figure}

% \begin{figure}[h]
% \includegraphics[scale=0.4]{DT.png}
% \caption{A binary decision tree with linear models in the leaf nodes, capturing the computations of programs in \imp, such as in Figure~\ref{fig:code-samples-prose}.}
% \label{fig:DT}
% \end{figure}

Figure~\ref{fig:DT} illustrates such a decision tree of height 3 computing a decision function. We note the computational semantics of a decision tree is identical to that of if-then-else programs in \imp. Below, we make this observation precise.

\begin{lem}
\label{lem:equiv1}
For every program $f \in \emph{\imp}$ defined over $\numcontext$ variables, there is a decision tree $f_T$, as given in Definition~\ref{def:DT} that behaves identical to $f$ on every input $\x \in \mathbb{R}^\numcontext$. In other words, for every $f \in \emph{\imp}$, there is $f_T$ such that $f = f_T$. 
\end{lem}

In the worst case, a sequence of $h$ conditionals in $\imp$ can result in
a complete tree of depth $h$ and size $2^h$. However, in practice, we find
that our algorithms tend to learn sparse (i.e. with many empty nodes that can be pruned) trees. Also, the sample complexity of our
algorithms depend only on $m$ and not on the size of the tree.
For these reasons, we do not worry about the exponential 
blow-up, which can happen in the worst case
when representing \imp\ programs as decision trees.

The converse of the above lemma also holds, which means we can convert any $f_T$ into an appropriate \imp~program.

\begin{lem}
\label{lem:equiv2}
For every tree $f_T$, there is $f \in \emph{\imp}$ that computes the same function as $f_T$. Synthesizing the program $f$ given $f_T$ can be done in time proportional to the total number of nodes (which is at most $2^{h+1}$ for a tree of height $h$) in the tree corresponding to $f_T$. 
\end{lem}
\subsection{Learning overview}
\label{sec:overview}
In light of the equivalence between trees and \imp, we can pose the reward-guided synthesis problem in Definition~\ref{def:rgs} as one of learning optimal decision trees, i.e. learning a tree $f_T$ that maximizes the expected reward $\mathbb{E}[r(f_T)]$. %$\mathbb{E}[r(B\circ f_T)]$.

\begin{defn} [PBR via Tree Learning] Given a specification consisting of:
\begin{enumerate}
    \item The input features $\x^{(t)}, 1\leq t\leq T$ to the decision tree model $f_T$,
    \item Query-access to reward function $r$, which returns a scalar reward value $r(f_T(\x^{(t)}))$ for decision values $f_T(\x^{(t)})$,
\end{enumerate}
the goal of PBR can be re-stated as learning \emph{optimal} decision tree model $f_T$ parameters, i.e.,
\begin{equation*}
    \treemodelopt \in \arg \max_{\treemodel}~~\mathbb{E}\bigg[\frac{1}{T}\sum_t r(f_T(\x^{(t)}; \treemodel))\bigg],
\end{equation*}
where $\mathbb{E}$ denotes the expectation with respect to any randomness in $r$ and features $\x^{(t)}$. Note that once we have the optimal tree model, we can synthesize the corresponding program in \emph{\imp}~by Lemma~\ref{lem:equiv2}.
\end{defn}
Following the machine learning convention, the observed variable values $\x \in \mathbb{R}^\numcontext$ are referred to as "features" or "context". 

The above given reformulation casts the PBR problem as that of learning parameters of the decision tree with a black box access to the reward function $r$. Note that as defined, formulation covers a wide range of settings, e.g., {\em offline} setting where all input features $\x^{(t)}$ are available apriori, or {\em synthesis} setting where we can design synthetic features $\x^{(t)}$ learn a decision tree. However, in this work, we mainly focus on the more general and practical setting of {\em online learning} that we describe below. 

For ease, we will drop the subscript in decision tree $f_T$ and refer to the tree model by $\pred$, which also stands for the \imp~program it represents. We parameterize the model $f$ by a vector $\weights \in \mathbb{R}^d$, where $d$ is the total number of parameters, and we write $f(\x; \weights)$ for $\x \in \mathbb{R}^\numcontext$. For example, in case of decision tree model in Definition~\ref{def:DT}, $\weights$ is simply the parameters $\treemodel$ vectorized, and $\numweights$ here is the total number of parameters in $\treemodelW$ and $\treemodelTheta$. Note that $\numweights \gg \numcontext$ in general; in case of tree of height $h$, $\numweights$ can be as large as $(\numcontext+1) \cdot 2^{h+1}$, corresponding to $\numcontext$ linear weights and a bias at each node in the tree including the leaf nodes (Definition~\ref{def:DT}).  

The learning problem proceeds in rounds, and in each round we observe features $\x^{(t)}$ and are {\em required} to output {\em decision values} $\paramvals^{(t)}=f(\x; \weights)$ which is then consumed by a black-box to output the reward value $r^{(t)}(\weights)$. The weights $\weights$ can then be modified based on the latest reward value. In this setting, the learning algorithm for \selftune{} is an instantiation of Algorithm~\ref{alg:basic}. 

Now the goal of this learning problem is to learn a model parameterized by $\weights$ so that the total rewards are maximized. However, as $\weights$ themselves are updated after each round, we use the standard {\em regret} notion to study different algorithms. Regret of an algorithm is the loss in reward when compared to the best solution ``hindsight'', i.e., $\max_{\weights} \ \ \sum_{t=1}^{T} r^{(t)}(\weights)$. Hence, the goal is to minimize the {\em regret} of the algorithm defined as: 
\begin{equation}
\label{eqn:regret}
R_T := \ \ \max_\paramvals \frac{1}{T} \sum_{t=1}^{T} r(\paramvals) - \frac{1}{T}\mathbb{E}\bigg[\sum_{t=1}^{T} r(\paramvals^{(t)}) \bigg],
\end{equation}
% \begin{equation}
% \label{eqn:regret}
% R_T := \mathbb{E}[r(\paramvals) - r(\paramvals^*)],
% \end{equation}
where the expectation $\mathbb{E}[.]$ is with respect to any randomness in the algorithm and in the reward $r$. Note that the notion of regret is general, and supports {\em offline} setting where all $\x^{(t)}$'s are given apriori and the goal is to find $\weights$ that maximizes the cumulative reward function. 
Ideally, the regret for an algorithm should decrease significantly with larger $T$, i.e., as we observe more rounds and samples, the reward of the algorithm should be similar to the reward of an optimal model. The rate of decrease of regret determines the {\em sample complexity} of the method, as introduced in Section~\ref{sec:problem}. 

%Now, based on the above formulation, we can evaluate any synthesis/learning algorithm on two key quantities: 
%\begin{enumerate}
 %   \item \textbf{Sample complexity} refers to the total number of queries made to the reward function $r$ to learn a "sufficiently good" model, i.e., to ensure that the regret is bounded by a pre-specified accuracy parameter. This quantity is important because in many settings querying $r$ can be expensive (or even disruptive, in settings where the model is learnt in an online fashion in real-world deployments) in terms of computation/time taken. Furthermore, this gives an objective way to compare two different algorithms on the same PBR problem instance (the algorithm with a smaller sample complexity is preferred). 
Besides sample complexity, another important metric for a learning/synthesis algorithm that proceeds in rounds to update the weights is that of \textbf{update complexity}, which denotes the time complexity of the \textsc{UpdateModel} step of the algorithm. This determines the efficiency of learning and in turn synthesis. The time complexity of synthesis is proportional to the sum of update complexity and the expected time required to query the reward function once. 
 
In practice, sample complexity is a more critical metric as each call of the reward function requires running the entire system which can be quite expensive. In next section, we present algorithms that are efficient in terms of the sample complexity and update complexity if the reward function satisfies certain assumptions. 

\begin{algorithm}
\caption{Learning in~\selftune{}}\label{alg:basic}
\begin{algorithmic}[1]
\Procedure{LearnInRounds}{}
\State Initialize the model parameters $\weights^{(0)} \in \mathbb{R}^d$, choose $\delta > 0$, and learning rate $\eta$
\For{$t = 0,1,2,\dots$}
\State Observe features (i.e. variable arguments) $\featvals^{(t)} := [\featval_1, \featval_2, \dots, \featval_\numcontext]$  
\State Compute decisions $\paramvals^{(t)} := \pred\big(\featvals^{(t)}; \weights^{(t)}\big)$ 
\State Query reward $r^{(t)} := r (\paramvals^{(t)})$  %\algorithmiccomment{\stc{\ \textbf{\texttt{SetReward}}}}
\State Update $\weights^{(t+1)}$ = \textsc{UpdateModel}($\featvals^{(t)}, r^{(t)}, \delta, \eta)$ %\hspace*{120pt}\algorithmiccomment{Happens at the server-side}
\EndFor
\EndProcedure
\end{algorithmic}
\end{algorithm}

%In the next section, we develop efficient techniques for implementing Step 7 of this algorithm for different types of decision functions. We are interested in two key quantities defined below that determine the efficiency of the algorithm.

\section{Learning Algorithms}
\label{sec:learning}
%Give details of the learning algorithm, and state main results (including that sample complexity doesnt depend on $d$)
In this section, we present learning algorithms for the three types of decision functions introduced in Section~\ref{sec:problem}: constants, linear models and decision trees corresponding to languages~\impconst,~\implin~and~\imp~respectively. That is, we instantiate Algorithm~\ref{alg:basic} for PBR when restricted to the above mentioned languages and also in certain cases, present regret bounds.\\
\textbf{Notation.} In the following, bold small letters $\mathbf{u}, \paramvals$, etc denote vectors. The subscript $u_i$ denotes the entry at the $i$th index of vector $\mathbf{u}$. The norm of a vector is defined as $\|\mathbf{u}\|_2 = \sqrt{\sum_{i} u_i^2}$. In all the cases, the dimensionality of the vector will be clear from the context.

\begin{figure}[t!]
\includegraphics[scale=0.3]{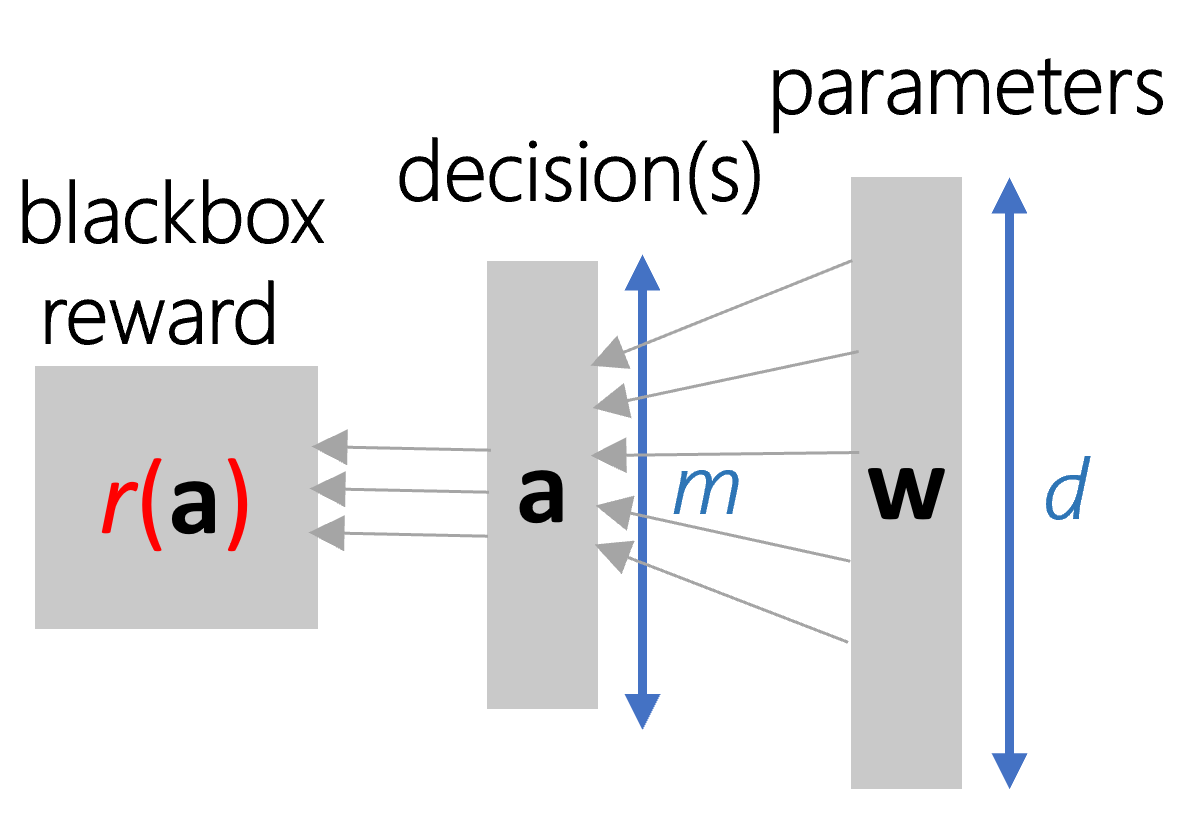}\vspace*{-5pt}
\caption{The black-box reward $r$ modeled as $r(\paramvals)$ where $\paramvals = f(\x; \weights) \in \mathbb{R}^m$ but $\weights \in \mathbb{R}^d$; in practice,  $d \gg m$.}
\label{fig:raw}
\end{figure}

\subsection{Intuition behind the Algorithms}
The key challenge with the problem is black-box access to the reward function, which disallows standard white-box optimization methods like gradient-descent. However, over the years, optimization literature has shown that for maximizing a given reward function, we do not require the {\em exact gradient}. Instead, as long as we can compute an {\em unbiased} estimate of the gradient, we can still allow strong optimization methods.  Several existing methods show that we can estimate unbiased gradient of a function by only black-box queries of the function at random points \cite{flaxman2005online,ghadimi2013stochastic,shamir2017optimal}. 

Our key insight is that we can produce more accurate estimates of the gradients by exploiting the structure of the {\em sketch}, i.e. of the function to be learned. Recall that, we get reward $r(\mathbf{a})$ for action/decision-values $\mathbf{a}\in \mathbb{R}^m$, where $\mathbf{a}=f(\x;\w)$ for some sketch function $f$ (like linear function or decision-tree function) parameterized by weights $\weights$ of dimension $d$ and the input features $\mathbf{x}$ are of dimension $p$. 
Now, if we use standard gradient estimates w.r.t. $\weights$ then the error depends on the dimensionality $d$ of $\weights$ due to random perturbation in each coordinate. 

However, as the black-box acts only on $m$-dimensional decision values $\mathbf{a}$, we can use chain-rule and the fact that we know sketch function $f$ completely to produce a significantly more accurate estimate of the gradient that is dependent only on dimensionality of $m$ of decision-values. In several real-world codebases we observe that $m\ll d$, and hence this technique is significantly more accurate.

%We build on the randomized gradient descent algorithm proposed by~\cite{flaxman2005online} in the context of online learning with bandit feedback, which estimates gradients using random perturbations on the decision values. Our technical insight is that this idea of perturbations can be generalized to linear functions as well as decision trees. 
%Figure~\ref{fig:raw} shows the dimensions of various quantities in our setting.
%we have a space of input features  $\mathbf{x}$, of dimension $p$, which is transformed using  a sketch/function (like linear function or decision-tree function ) parameterized by $\weights$ of dimension $d$ to a decision value $\mathbf{a}$ of dimension $m$, and the decision value is fed into the black-box to produce a reward.
%Our main intuition is that  since $m$ is much smaller than $d$, this structure can be exploited to estimate the gradients significantly more accurately.
%If the decision function is a linear model, then we show that applying the random perturbation to the result of the linear transformation, that is, on the decision value $\mathbf{a}$, is provably more efficient than perturbing the inputs $\mathbf{x}$ themselves. 

If the decision function is a tree, then there are additional challenges due to non-smooth and piece-wise linear nature of such functions. We propose a differentiable and shallow neural network model, and show that under some structural constraints, it implicitly represents decision trees. Then, we learn the parameters of the shallow neural network using gradient descent, with the perturbation trick. We also show that the framework is theoretically-founded ---in cases when the rewards satisfy nice properties, the parameters learnt, and in turn the synthesized code, is optimal in a precise sense.

\subsection{Learning constants ($f \in \impconst$)}
\label{sec:learning_constant}
We first consider synthesis with \impconst~language, which reduces to learning a set of constants, without any features. That is, in this case, the decision values given by the synthesized program ($\paramvals$) are just the model parameters, i.e., $\paramvals=\weights$, and hence $m=d$. See Figure~\ref{fig:code-samples-prose} (left) for an example of such a sketch.

For \impconst~language the problem reduces to the standard {\em online learning with bandit feedback} problem, which can be solved using a randomized gradient descent algorithm proposed by \cite{flaxman2005online} (see Algorithm~\ref{alg:constant}). Note that line $3$ of the algorithm first estimates the gradient at $\paramvals^{(t)}$ by using random perturbation $\mathbf{u}$ and then performs standard gradient ascent (because we are maximizing reward) over $\paramvals$ using the estimated gradient. Also, while Algorithm~\ref{alg:constant} mostly follows template of Algorithm~\ref{alg:basic}, there is a minor difference which is that the reward function is not queried at the predicted decision value $\paramvals^{(t)}$, instead it is queried at a perturbation of $\paramvals^{(t)}$. 

\begin{algorithm}[H]
\caption{Learning constants}\label{alg:constant}
\begin{algorithmic}[1]
\Procedure{UpdateConstantModel (\textbf{Input:} $\featvals^{(t)}$, $\eta$, $\delta$)}{}
%\State Initialize parameter values $\paramvals^{(0)}$  
%\For{$t = 1,2,\dots$}
\State Sample $\mathbf{u}$ uniformly from $\{ \mathbf{u}\ | \ \|\mathbf{u}\|_2 = 1 \}$
%\State $\paramvals^{(t+1)} = \paramvals^{(t)} - \eta \frac{1}{\delta} r_t(\paramvals^{(t)} + \delta \mathbf{u}) \mathbf{u}$
\State $\paramvals^{(t+1)} =  \paramvals^{(t)} + \eta \frac{1}{\delta} r(\paramvals^{(t)} + \delta \mathbf{u}) \mathbf{u}$
%\EndFor
\EndProcedure
\end{algorithmic}
\end{algorithm}

Note that in this language, there is no input context/features $\x$, hence $p=0$. Also, the number of decision values is same as the number of parameters $d$. It is easy to see that the update complexity of the method is only $O(d)$. Below we formally state the sample complexity as well under certain assumptions that we define first. 
\begin{defn} [Concavity] The reward function $r$ is said to be concave if, for all $\paramvals, \paramvals' \in \mathbb{R}^m$ and $\lambda \in [0,1]$,
\[r\big(\lambda \paramvals + (1-\lambda) \paramvals'\big) \geq \lambda r(\paramvals) + (1-\lambda) r(\paramvals')\ . \]
\label{def:concave}
\end{defn}
\begin{defn} [Lipschitz-continuity] The reward function $r$ is said to be $L$-Lipschitz continuous if for all $\paramvals, \paramvals' \in \mathbb{R}^m$,
\[ |r(\paramvals) - r(\paramvals')| \leq L\|\paramvals - \paramvals' \|_2\ . \]
\label{def:lipschitz}
\end{defn}
% \begin{thm}
% Assume the reward function $r$ is concave (Definition~\ref{def:concave}) and Lipschitz continuous (Definition~\ref{def:lipschitz}). Define $\bar{\paramvals} = \frac{1}{T}\sum_{t=1}^T \paramvals^{(t)}$, where $\paramvals^{(t)}$ are the decisions given by Algorithm~\ref{alg:constant} at the end of $T$ rounds. Then:
% \[ R_T \leq C \cdot \frac{\numparam}{T^{1/4}}\ , \]
% where $C>0$ is a global constant. 
% \end{thm}
\begin{thm}[\cite{flaxman2005online}]
Assume the reward function $r$ is concave (Definition~\ref{def:concave}) and Lipschitz continuous (Definition~\ref{def:lipschitz}). At the end of $T$ rounds, Algorithm~\ref{alg:constant} satisfies:
\[ R_T \leq C \cdot \frac{\numparam}{T^{1/4}}\ , \]
where $C>0$ is a global constant. 
\label{thm:constant1}
\end{thm}
% If the reward functions are concave and Lipschitz continuous, ~\citet{flaxman2005online} showed that the regret of Algorithm~\ref{alg:constant} converges at a rate proportional to $\frac{\numparam}{T^{1/4}}$, where $\numparam$ is the number of constants to tune and $T$ is the number of times we obtain a reward from the system. That is, 
% \begin{equation}
% R_T := \ \ \max_\paramvals \frac{1}{T} \sum_{t=1}^{T} r_t(\paramvals) - \frac{1}{T}E\bigg[\sum_{t=1}^{T} r_t(\paramvals^{(t)})\bigg] \leq \frac{C\cdot \numparam}{T^{1/4}},
% \end{equation}
That is, to ensure $R_T\leq \epsilon$, the {\em sample complexity} of the algorithm is $O(\numparam^4/\epsilon^4)$. The above bound can be further improved in settings (e.g. offline) where it is possible to obtain reward $r(\paramvals)$ at two different $\paramvals$ values. In that case, the gradient estimator in Step 4 of Algorithm~\ref{alg:constant} can be replaced with: 
\begin{equation}
\paramvals^{(t+1)} =  \paramvals^{(t)} + \eta \frac{1}{\delta} (r(\paramvals^{(t)} + \delta \mathbf{u})-r(\paramvals^{(t)} - \delta \mathbf{u})) \mathbf{u}.
\label{eqn:twopoint}
\end{equation} 
And the bound can be improved~\cite{shamir2017optimal} as stated below.
%given in Step 5, one can show that $R_T$ is at most a constant factor of %$\frac{\numparam}{\sqrt{T}}$, i.e.,  
\begin{thm}[\cite{shamir2017optimal}]
Assume the reward function $r$ is concave (Definition~\ref{def:concave}) and Lipschitz continuous (Definition~\ref{def:lipschitz}). At the end of $T$ rounds, Algorithm~\ref{alg:constant}, with the ``two-point'' gradient estimator~\eqref{eqn:twopoint} in Step 4, satisfies:
\[ R_T \leq C \cdot \frac{\numparam}{T^{1/2}}\ , \]
where $C>0$ is a global constant. 
\label{thm:constant2}
\end{thm}
% \begin{equation}
% R_T := \ \ \max_\paramvals \frac{1}{T} \sum_{t=1}^{T} r_t(\paramvals) - \frac{1}{T}E\bigg[\sum_{t=1}^{T} r_t(\paramvals^{(t)})\bigg] \leq \frac{C\cdot \numparam}{T^{1/2}}. 
% \end{equation}
That is, in the offline  setting,  the rate of convergence wrt $T$ improves significantly which is critical for deployment of such solutions as the reward computation in general is expensive. Furthermore, note that both the regret bounds suffer from a linear dependence on the number of constants to learn ($d=\numparam$) which is tight as the method is required to explore $r(\paramvals)$ in all the $\numparam$ coordinates to accurately estimate the gradient. So, if in the extreme case, if every parameter to tune in the system is treated as individual constants, irrespective of the decision outcome's structure, then $m = d$ which is expensive; typically number of outcomes $m$ is an order of magnitude smaller than the number of parameters $d$. The next two sub-sections discuss how one can exploit additional structure in certain templates, when applicable, to reduce the dependence on $d$.
\subsection{Learning linear models ($f \in \implin$)}
\label{sec:learning_tree}
Now let us consider the problem of synthesizing programs from the \implin language,  where the decisions $\paramvals$ are fixed to be a linear function of some observed features $\featvals$. That is, $\paramvals=\pred(\featvals; \weights)$ where $\pred$ is a fixed linear function and $\featvals$ are the observed features/context. Note that unlike in Section~\ref{sec:learning_constant} where $\paramvals$ were fixed to be constant despite changing context of the system ($\x$), in this case, we allow $\paramvals$ to change as a linear function of the features $\featvals$.

Following the online setting described in Section~\ref{sec:problem}, we observe features $\featvals^{(t)} \in \mathbb{R}^\numcontext$ in the $t$-th round and provide decision values $\paramvals^{(t)}\in\mathbb{R}^m$ given by $\paramvals^{(t)}=\weights^{(t)} \featvals^{(t)}$ where $\weights^{(t)}\in \mathbb{R}^{m\times \numcontext}$. The black-box system $B$ then provides reward $r^{(t)}(\paramvals^{(t)})$ for the predicted $\paramvals^{(t)}$, and the parameters $\weights$ are then updated based on the reward. Thus the regret is defined for this problem as: 
\begin{equation}
\label{eqn:regretlin}
    R_T := \ \ \max_\weights \frac{1}{T} \sum_{t=1}^{T} r\big(\pred(\featvals; \weights)\big) - \frac{1}{T}\mathbb{E}\bigg[\sum_{t=1}^{T} r\big(\pred(\featvals; \weights^{(t)})\big)\bigg].
\end{equation} 
Note that in this case, the number of parameters $d = m \times p$ which can be very large as in typical systems, we would require several features to capture the context of the system. This is illustrated in Figure~\ref{fig:raw}.
If we treat $\weights$ as constants to be learned and apply Algorithm~\ref{alg:constant} with rewards $\widetilde{r}(\weights^{(t)})=r(\weights^{(t)}\featvals^{(t)})$, the regret for such a method would scale as $d/T^{1/4} = m\cdot p/T^{1/4}$. On the other hand, $m$ which is the total number of decisions $\paramvals$ to be estimated tends to be significantly smaller (typically $<\sim 10$). So the question is if we can devise an algorithm that scales better with $\numfeat$ when $m\ll \numfeat$.

\begin{algorithm}[H]
\caption{Learning linear models}\label{alg:linear}
\begin{algorithmic}[1]
\Procedure{UpdateLinearModel (\textbf{Input:} $\featvals^{(t)}$, $\eta$, $\delta$)}{}
%\State Initialize parameter values $\weights^{(0)}$ 
%\For{$t = 1,2,\dots$}
%\State Choose  $\delta > 0$, and learning rate $\eta$
\State Define $\paramvals^{(t)} := \weights^{(t)}\featvals^{(t)} \in \mathbb{R}^\numparam$, where $\featvals^{(t)} \in \mathbb{R}^\numcontext$ and $\weights^{(t)} \in \mathbb{R}^{\numparam\times\numcontext}$
\State Sample $\mathbf{u} \in \mathbb{R}^m$ uniformly from $\{ \mathbf{u}\ | \ \|\mathbf{u}\|_2 = 1 \}$
\State {\small $\weights^{(t+1)} = \weights^{(t)} + \eta \frac{m}{\delta} r(\paramvals^{(t)}+\delta u)\mathbf{u} \cdot (\featvals^{(t)})^{\text{T}}$}
%\EndFor
\EndProcedure
\end{algorithmic}
\end{algorithm}
We exploit the linear structure of the template to significantly reduce the regret and hence the number of samples required to obtain good estimate of $\paramvals$. We make the following simple observation: $\nabla_\weights r(\featvals^{(t)}\weights)=\featvals^{(t)} (\nabla_\paramvals r(\paramvals))^{\text{T}}$ where $\paramvals:= \weights\featvals^{(t)} $ and $(b)^{\text{T}}$ is the transpose of $b$. Hence, the gradient wrt $\weights$ has a special form that requires estimating only a $m$ dimensional vector $\nabla_\paramvals r(\paramvals)$ instead of a $m\times \numcontext$-dimensional matrix. This leads to significantly cheaper gradient estimation step, implying a tighter regret bound and sample complexity that we discuss below. Also, note that the update complexity of the method is $O(mp)$, which is optimal.

Our novel algorithm for updating $\weights$ is given in Algorithm~\ref{alg:linear} and the regret of the algorithm is given by: 
%
%
%
%Note that in general $\numfeat$ can be significantly larger than the number of parameters $m$ to be estimated. 
%As in the previous section, we can use Algorithm~\ref{alg:constant} 
%The regret of Algorithm~\ref{alg:linear} can be bounded as follows. \naga{Rough statement, need to be made precise, and much more lucid, currently I don't define any of the technical terms in the statement.}
\begin{thm}
\label{thm:linear}
%	Let $\featvals^{(t)}\stackrel{i.i.d.}{\sim} \mathcal{D}$ for all $1\leq t\leq T$, i.e., all $\featvals^{(t)}$ are sampled from a fixed distribution $\mathcal{D}$. Also, 
Let $r$ be concave and Lipschitz continuous. Then the regret defined in Equation~\ref{eqn:regretlin} for Algorithm~\ref{alg:linear} can be bounded as:
\[ R_T = \max_\weights \frac{1}{T} \sum_{t=1}^{T} r\big(\weights \featvals^{(t)} \big) - \frac{1}{T}\mathbb{E}\bigg[\sum_{t=1}^{T} r\big(\weights^{(t)}\featvals^{(t)}\big)\bigg] \leq O\bigg(\frac{m}{T^{1/4}}\cdot \sqrt{\max_t \|\featvals^{(t)}\|_2}\bigg)\ .\]
\end{thm}
The above regret bound implies sample complexity bound of $O(m/\epsilon^4)$ to ensure regret of $\epsilon$. The proof of the theorem relies on Lemma 3.1 of~\cite{flaxman2005online}; but the key observation we make is that the norm of the noise in the gradient in Step 5 of Algorithm~\ref{alg:linear} is bounded by $O(m)$ rather than $O(d)$, if we appropriately choose $\delta$ and $\eta$ at line 4. See Appendix~\ref{app:proofs} for details. Note that the above regret bound is dependent only on $m$ and is completely independent of $d$ given $m$. Dependence on $d$ (through $p$) can creep through $\|\featvals^{(t)}\|_2$ but several practical problems tend to have small $\|\featvals^{(t)}\|_2$ and in general such a bound is considered to be ``dimension independent''. In contrast, the regret of Algorithm~\ref{alg:constant} is bound to suffer a linear dependence on $d = mp$. This implies significantly smaller regret for Algorithm~\ref{alg:linear} for $m\ll d$ which is a typical case in practical applications. Finally, similar to the previous section, in the offline deployment setting with two-point feedback, the regret bound decreases at $T^{-1/2}$ rate.  
%Finally, the above bound assumes each $\featvals^{(t)}$ is sampled from a fixed distribution. For arbitrary $\featvals^{(t)}$, that might not follow any distribution, we can get a similar bound but with $T^{1/4}$ rate. 
%\begin{remark}
%If we use the standard formulation, and update $\weights$ for the linear model, using gradient estimate in Step 5 of Algorithm~\ref{alg:constant}, then the regret will be:
%\[ R_T \leq O\bigg(\frac{d}{\sqrt{T}}\bigg). \]
%\end{remark}

\subsection{Learning tree models ($f \in \imp$)}
\label{sec:learning_tree_models}
In this section, we discuss our  method for synthesizing a program from the general \imp~language. As discussed in Section~\ref{sec:problem}, the problem is equivalent to that of learning a general decision tree with rewards, which is a challenging problem and has been relatively unexplored in both machine learning and programming languages literature. 

Recall that the decisions $\paramvals$ are fixed to be a function $\pred$ of the context ($\featvals\in \mathbb{R}^\numcontext$), where $\pred$ is a tree structured function, i.e.,  $\paramvals=\pred(\featvals;\weights)$ with $\pred$ being a decision tree parameterized by $\weights$. 
% For example, such a function can capture the following tree-based template: 

% \begin{lstlisting}[language=dsl,morekeywords={public,static,double,return,if,else},basicstyle=\small\ttfamily]
%     double ScoreLinesMap(double[] features)
%     {
%         double[4] w = {0,1,0.5,0.5};
%         double param = -1;
%         if(features[0] > w[0]):  param = w[3] * features[1] - w[1];
%         else:  param = w[2] * features[2];
%         return param;
%     }
% \end{lstlisting}
As in Algorithm~\ref{alg:basic}, the learner would observe $\featvals^{(t)}$ in the $t$-th round, propose $\weights^{(t)}$ to predict decisions $\paramvals^{(t)}$, receive reward $r(\paramvals^{(t)})$ and the goal is to optimize the reward. If we ignore the tree structure of $\paramvals$ and treat $\weights$ as the parameters to be learned using constant template and apply Algorithm~\ref{alg:constant}, it would lead to poor regret and hence many rounds for learning a good solution (as discussed in linear templates). The reason is that the size of $\weights$ typically increases exponentially with the height of the tree, which implies that the regret bound of Section~\ref{sec:learning_constant} would also increase exponentially with the height of the tree. 

Instead, we use the following observation, that we exploited in the previous section as well (for simplicity we provide this observation when $\paramval \in \mathbb{R}$ is a scalar): \[\nabla_\weights r(\pred(\featvals^{(t)}; \weights^{(t)}))=r'(\paramval^{(t)})\nabla_\weights \pred(\featvals^{(t)}; \weights^{(t)}),\]
where $\nabla_\weights r(\pred(\featvals^{(t)}; \weights^{(t)}))$ is the derivative of $r(\cdot)$ wrt $\weights$, evaluated at $\weights^{(t)}$. Now, $r'(\paramval)$ is the first derivative of $r$ evaluated at $\paramval^{(t)}=\pred(\featvals^{(t)}; \weights^{(t)})$ and $\nabla_\weights f(\cdot)$ is the derivative of $\pred(\cdot)$ wrt $\weights$. 

Note that we do not know function $r$ and can only receive feedback $r(\paramvals^{(t)})$, so $r'$ would be computed using the standard perturbation based technique (see Algorithm~\ref{alg:constant}). However, as we know function $\pred$, we can evaluate $\nabla_\weights \pred(\cdot)$  accurately assuming $\pred$ represents a differentiable tree function. Together, this can significantly reduce the number of prediction-reward feedback loops for learning the template, even though the number of parameters $\weights$ can be very large --- as $r'$ is a one-dimensional quantity (and in general $m$ dimensional where $\paramvals\in \mathbb{R}^m$) and only very few random perturbations are required for its accurate estimation. %On the other hand the number of parameters can be still be large (exponential in height of the tree), so a direct application of Algorithm~\ref{alg:constant} would require a significantly larger number of random perturbations and hence a large number of reward feedback loops.  
The key challenge here is that the algorithm requires computing gradient of $\pred(\cdot)$ wrt $\weights$ which in general is a challenging task due to the discrete structure of trees. In this work, we develop a novel differentiable model for trees.

% \begin{algorithm}[H]
% \caption{Learning tree models (stated for $m = 1$)}\label{alg:tree}
% \begin{algorithmic}[1]
% \Procedure{UpdateTreeModel (\textbf{Input:} $\featvals^{(t)}$, $\eta$, $\delta$, $h$)}{}
% %\State Initialize parameter values $\weights^{(0)}$ 
% %\For{$t = 1,2,\dots$}
% %\State Choose  $\delta > 0$, and learning rate $\eta$
% \State Define $\paramval^{(t)} := \pred(\featvals^{(t)},\weights^{(t)}) \in \mathbb{R}$, where $\featvals^{(t)} \in \mathbb{R}^\numcontext$ are input features and $\weights^{(t)} \in \mathbb{R}^d$ are the tree model parameters 
% \State Sample $u\in \mathbb{R}$ uniformly from $\{-1,1\}$
% \State Update $\weights^{(t+1)} = \weights^{(t)} -  \frac{\eta}{\delta} r(\paramval^{(t)}+\delta u)u \cdot \nabla_\weights \pred(\featvals^{(t)}, \weights^{(t)})$
% %\EndFor
% \EndProcedure
% \end{algorithmic}
% \end{algorithm}

\subsubsection{Decision Trees as \arch}
\label{sec:arch}
The core idea of our approach and the motivation are as follows. Learning decision tree models is intractable in general, and is hard even for well-behaved \textit{known} reward functions, because: (a) it is highly non-smooth and (b) the function class is piecewise-constant (or piecewise-linear). Existing approaches typically try to solve the problem greedily~\cite{carreira2018alternating} or try to come up with a smooth relaxation or upper bound of the loss function~\cite{norouzi2015efficient}. A major shortcoming of these techniques is that they are tied to a specific loss (reward) function or type of decisions made and therefore do not extend to more general settings. 

Our \textit{non-greedy} approach  uses a \textit{differentiable} and \textit{shallow} neural network model for learning decision trees and works with any general loss/reward function. Moreover, we can formally show that the neural network architecture under some structural constraints is {\em equivalent} for decision trees, thus allowing  gradient based training methods for learning tree parameters. We refer to this model as~\ent, denoted by $\fent$. The details of this approach are given in Appendix~\ref{app:tree}.

\begin{algorithm}
\caption{Learning trees in~\selftune{}}\label{alg:tree}
\begin{algorithmic}[1]
\Procedure{LearnInRounds (\textbf{Input:} $h$)}{}
\State Initialize \ent~$\fent$ parameters (vectorized) $\netmodelw^{(0)} \in \mathbb{R}^d$ for the given height $h$
\State Choose $\delta > 0$, and learning rate $\eta$
\For{$t = 0,1,2,\dots$}
\State Observe features $\featvals^{(t)} := [\featval_1, \featval_2, \dots, \featval_\numcontext]$  
\State Define $\paramval^{(t)} := \fent(\featvals^{(t)},\weights^{(t)}) \in \mathbb{R}$, where $\featvals^{(t)} \in \mathbb{R}^\numcontext$ are input features 
\State Sample $u\in \mathbb{R}$ uniformly from $\{-1,1\}$
\State Update $\netmodelw^{(t+1)} = \netmodelw^{(t)} +  \frac{\eta}{\delta} r(\paramval^{(t)}+\delta u)u \cdot \nabla_{\netmodelw} \fent(\featvals^{(t)}, \netmodelw^{(t)})$
\EndFor
\State \textbf{return} Tree model (Definition~\ref{def:DT}) $\treemodel = \textsc{InferTree}(\netmodelw, h)$ (Algorithm~\ref{algo:infertree} in Appendix~\ref{app:proofs})
\EndProcedure
\end{algorithmic}
\end{algorithm}

\textbf{Algorithm.} We provide the pseudo-code for learning trees, for the case when $\paramval\in \mathbb{R}$, in Algorithm~\ref{alg:tree} (which also subsumes the basic skeleton in Algorithm~\ref{alg:basic}) but it can be easily extended to the general decision functions in \imp~(that return $m$ values instead of 1). Note that the algorithm learns parameters for the equivalent function $\fent$ and eventually returns the intended decision tree function (which can be easily inferred given $\fent$ via \textsc{InferTree} procedure stated in the Appendix~\ref{app:proofs}). While intuitively exploiting the tree structure should lead to significantly smaller regret bound, it is difficult to provide a rigorous analysis of the same. Due to tree structure $\pred(\cdot)$ being a non-convex  function, the standard online learning techniques~\cite{flaxman2005online} do not apply in this case. While certain novel techniques like~\cite{agarwal2019learning} have been designed for non-convex optimization, we leave further investigation into the regret bound and hence the sample complexity of this method for future work.  

\subsection{PBR usage in practice}
Here we briefly discuss how our PBR method can be applied in real-world codebases. At a high level, the programmer must first decide the set of decision values that are critical to the performance of the system, and also figure out the context/features important for setting the decision values. Then, the programmer sets up a reward function based on critical metrics that needs to be optimized. Finally, the programmer can either specify the language (\impconst, \implin, \imp) from which a program should be synthesized. Our tool takes care of storing the feature values and the corresponding reward functions, and learning the appropriate decision function. See Appendix~\ref{sec:API} for more details about the front-end that enables using our PBR method easier for developers. Note that our exposition focus only on setting where programmer does not provide any partial sketch; further empirical validation of our methods in partial sketching settings is left for future work. 

While our sample complexity bounds hold only for Lipschitz continuous and concave functions, in practice the functions might have discontinuities. However, we observe empirically that our method is indeed able to learn  effective  function/sketch parameters. We attribute this to the fact that in practice, the reward function would have a few points of discontinuities~\cite{chaudhuri2012continuity}, thus the sample complexity is small in large portions of the parameter space. Furthermore, perhaps the first order or second order optima (instead of global optima) are also reasonably good solutions, which can be ensured by techniques similar to our method \cite{agarwal2019learning}. We leave rigorous analysis of practical reward functions and our methods in those settings to future work.

%Here again, we aim to utilize a gradient descent style method with the following observation: 

%\section{Deployment Modes}
%Discuss offline vs online

\section{Implementation and Evaluation}
\label{sec:evaluation}
In this section, we discuss implementation and present empirical evaluation of our algorithms. Our algorithms for the~\selftune{} framework operate in both online and offline settings requiring only black-box access to the reward function, and the key benefit of our algorithms is in settings where the structure of the decision function/sketch (linear or tree) can be exploited. We design evaluation studies that bring out some of these aspects and merits under different settings. In particular, we seek answers to the following questions.
\begin{enumerate}
   
    \item In practice, \textit{how well do our algorithms help learn decision functions in codebases with complex reward functions?} In this study, we apply our algorithms to synthesize search and ranking heuristics in the widely-used~\cite{prosesdk} codebase for synthesizing data formatting programs in the \textsc{FlashFill} DSL. %Here, we compare with the standard reinforcement-learning techniques~\cite{agarwal2016multiworld}.
    
    \item \textit{How do our (black-box)~\selftune~algorithms compare with white-box synthesis techniques?} As it is difficult to construct strong real-world examples of white-box reward functions, we consider somewhat artificial programs proposed by prior work \cite{chaudhuri2014bridging}, where the entire system, including reward function, is available
    as a white-box, and the reward function can be invoked as many times
    as desired. In this artificial setting, we compare our approach with
    ~\sketch{}~\cite{sketchtool}, which uses SAT solvers to synthesize integer valued parameters, and ~\Fermat~\cite{chaudhuri2014bridging}, which synthesizes continuous valued parameters using numerical methods.
    
    \item \textit{Does exploiting structure in the decision functions really help?} Here, we want to compare our algorithms that exploit the decision function structure presented in Section~\ref{sec:learning} against treating the synthesis as a sketch with missing numerical parameters, which can be addressed using black-box reinforcement learning style methods.
%    \item \textit{Can we learn decision functions that are partly known, with smaller sample complexity?} Here, we show we can leverage part knowledge of weights in the linear model and some tree structure in case of tree models, with fewer reward queries than otherwise.
\end{enumerate}
\textbf{Evaluation settings and baselines.} We work with both online (where we need to take decisions at every round) and offline settings, and black-box rewards. We evaluate and compare ~\selftune~algorithms with different baselines, each applicable only to certain settings unlike our algorithms (and therefore selectively used for comparison in the aforementioned three studies as applicable): 
(1) general-purpose (evolutionary) optimization algorithms in the popular open-source \Nevergrad~platform~\cite{nevergrad}, which is also applicable to black-box, offline and continuous parameter settings; and
(2) the multi-arm bandit formulation~\cite{bietti2018contextual} that is currently used in the popular~\decisionservice~\cite{agarwal2016multiworld}, an enterprise-scale reinforcement learning framework, that works in black-box and online learning settings, but is restricted to discrete parameters/decisions (such as recommending ads or news articles to users); and 
(3) the~\sketch{} synthesis tool~\cite{sketchtool} which is applicable to white-box and offline settings, and can synthesize integer-valued parameters; and
(4) the \Fermat tool~\cite{chaudhuri2014bridging}, which is also applicable to white-box, offline settings, and can synthesize continuous-valued parameters.
%\red{naga: a short para on settings (black-box vs white-box, online vs offline) and motivate our choice of baselines that are applicable under different settings.}
\subsection{Implementation}
\label{sec:implementation}
We have implemented the PBR framework (\decfun~and \assignr~API introduced in Section~\ref{sec:problem}), and the learning algorithms presented in Section~\ref{sec:learning} as a utility library (with support for C\# and Python languages) for software developers. Our implementation also provides flexibility and customization capabilities to the developer in terms of explicitly providing domain knowledge (for example, the developer can suggest one of \imp,~\impconst~or~\implin~to synthesize an appropriate decision function) and ability to inspect and debug the synthesized code for the decision function (instead of working with the tree or linear model as a black-box). Details of the API are given in Appendix~\ref{sec:API}. All experiments are conducted on a standard desktop machine with 16GB RAM, 2GHz processor and 4 cores.

\subsection{How well do our algorithms help learn decision functions in real codebases with complex reward functions?}
\label{sec:prose}
PROSE \cite{prosesdk} is a well-known framework for programming by examples (PBE). It has been instrumental in developing software for many practical applications around data ingestion~\cite{raza2018disjunctive,iyer2019synthesis}, formatting, spreadsheet processing~\cite{gulwani2011automating}, web extraction, program repair and transformation~\cite{le2017s3,rolim2017learning}, and more. PROSE provides a meta-framework~\cite{polozov2015flashmeta} for (a) defining a domain-specific language (DSL) for programs of interest, (b) synthesizing programs from a given input-output specification in a divide-and-conquer fashion, (c) pruning the search space, and (d) ranking the (sub)programs to narrow down to one or a few programs intended by the user. Often, software applications built over the PROSE engine require coming up with their own DSL (e.g. spreadsheet processing vs web extraction) and ranking function. 

Several critical decision making points exist in the resulting software; in particular, developers write several complex heuristics for determining how to score different operators in the DSL and ranking the sub-programs that these operators compose, and in turn, for choosing the ``best programs'' capturing the user intent with very few input-output examples. There is a growing line of research in the intersection of programming languages and AI~\cite{gulwani2017programming}. In this study, we consider the popular spreadsheet processing application DSL~\cite{gulwani2011automating} that has been commercially deployed~\cite{msflash}. The corresponding ranker (available here~\cite{prosesdk}, details are provided in~\cite{pmlr-v89-natarajan19a}) has several heuristic rules, each implemented as a function involving multiple hard-coded constants and features. For instance, the \texttt{RegexPair} rule shown in Figure \ref{fig:code-regexpair} has three hard-coded constants.

\begin{figure}
	\begin{lstlisting}[escapechar=!,language=csh,morekeywords={public,static,double,return}]
    public double RegexPair(double bias_RegexPair, double score_RegexPair_r, double score_RegexPair_r2) {
        return !\colorbox{yellow}{-0.205}!*bias_RegexPair +
        !\colorbox{yellow}{1}!*score_RegexPair_r + !\colorbox{yellow}{1}!*score_RegexPair_r2;
    }
    \end{lstlisting}\vspace*{-10pt}
	\caption{The \texttt{RegexPair} heuristic in \prose~\cite{prosesdk,pmlr-v89-natarajan19a} with parameters of interest highlighted.}
	\label{fig:code-regexpair}
\end{figure}

We consider the problem of learning these typically hand-tuned parameters to improve the performance of PROSE as measured on a set of curated benchmark tasks~\cite{pmlr-v89-natarajan19a}. Here, the reward function is defined as the fraction of tasks for which the software synthesizes a correct program for. This reward function is highly discontinuous and complicated, and computing the function involves running the entire software with complex recursive algorithms interleaving synthesis and ranking. But, we know that the constants in the ranking functions (e.g., Figure~\ref{fig:code-regexpair}) heavily influence the performance of PROSE on most of the tasks, and thus, the reward. Note that evaluating the reward function is expensive in terms of time --- a single evaluation takes nearly 10 minutes. So we evaluate algorithms based on the number of reward computations (sample complexity) as well as the time taken to converge to a good solution (i.e. one that achieves good performance on the benchmark).

Recall that due to black-box nature of the reward function, existing sketch methods that require whitebox access do not apply \cite{sketchtool,chaudhuri2014bridging}. Instead, we compare our method against a multi-arm bandit method (UCB, as implemented in the open-source framework SMPyBandits \cite{SMPyBandits})-- a popular reinforcement learning style method. However, UCB requires discretization of the real-valued parameters into a small number of actions, which means that the number of actions grows exponentially with the number of parameters and can be tricky when tuning parameter values of high precision. In this case, we discretize each parameter value into 9 bins around 0 (which is a good initial solution). This implies that just for one of the heuristics (\texttt{RegexPair} mentioned above), the total number of actions (from which UCB chooses one set of decision-value/action) turn out to be $729$. So, to ensure applicability of the UCB method, we initially focus only on learning {\em one} heuristic function in PROSE codebase. 

\begin{figure}[t]
\begin{tabular}{cc}
\hspace*{-10pt}\includegraphics[width=.48\linewidth]{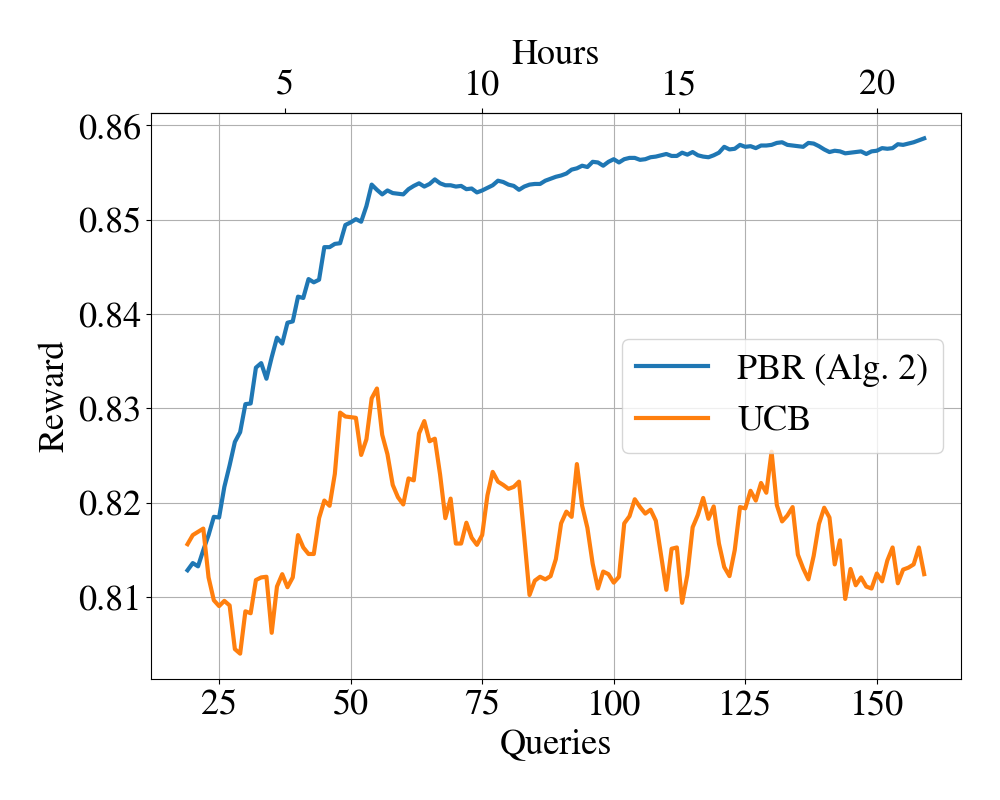}&
\hspace*{-10pt}\includegraphics[width=.48\linewidth]{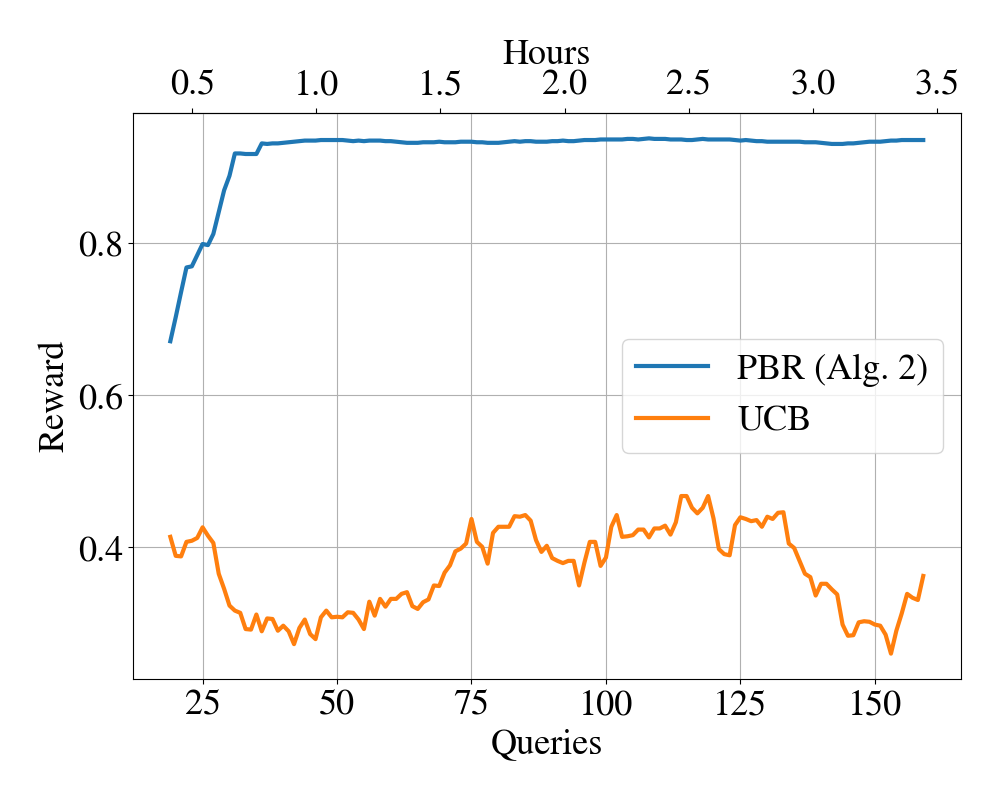}\vspace*{-5pt}\\
(a)&(b)
\end{tabular}
\caption{(Smoothed) Reward over time wrt number of calls to reward oracle to learn different functions in \prose. For the multi-arm bandit formulation (UCB), we discretize each parameter before execution. {\bf (a)}: Learning \texttt{RegexPair} with $3$ parameters (Figure~\ref{fig:code-regexpair}). For UCB, we discretize each parameter into $9$ values resulting in {$9*9*9 = 729$} actions. The sample complexity of UCB is quite high and the solution it obtains is relatively poor even after nearly 1 day. {\bf (b)} Learning \texttt{FormatDateTimeRange} with $10$ parameters (Figure~\ref{fig:code-formatdatetimerange} in Appendix~\ref{app:evaluation}). For UCB, we discretize each parameter into $5$ values resulting in $5^{10} \sim 9$ million actions.}
%\red{mention reason for smaller time/query?}}
\label{fig:prose_zo_ucb_m22_mean_reward}
\end{figure}

Figure \ref{fig:prose_zo_ucb_m22_mean_reward} (a) shows the comparison of~\selftune{} and UCB algorithms for learning the three parameters associated with the \texttt{RegexPair} function (Figure~\ref{fig:code-regexpair}).  We observe that (a) \selftune{} quickly ramps up the reward in about 50 queries (in about 6 hours), whereas the UCB algorithm spends a lot of time and queries performing ``explore-exploit'' of multiple independent actions, without gaining enough confidence on any particular action. Even if we reduce the action space of UCB to 125 actions (i.e. discretize each parameter into 5 bins instead of 9), the performance of UCB is still significantly worse than~\selftune{} as observed from Figure~\ref{fig:prose_zo_ucb_m11_mean_reward} (in Appendix~\ref{app:evaluation}). The observations are similar in Figure \ref{fig:prose_zo_ucb_m22_mean_reward} (b) that shows a comparison for synthesizing the \texttt{FormatDateTimeRange} function (Figure \ref{fig:code-formatdatetimerange} in Appendix~\ref{app:evaluation}) which has a larger number of parameters to learn. 

An important aspect of evaluation of parameter learning is how often the software or the system that is tuned is exposed to ``bad'' rewards. We find that~\selftune{} spends much smaller fraction of time with low rewards as against UCB (Figures~\ref{fig:prose_zo_ucb_m22_hist} and~\ref{fig:prose_large_zo_ucb_hist} in Appendix~\ref{app:evaluation}).

\textbf{Deployment.} Next, we deployed our~\selftune{} framework to jointly learn the parameters ($d = 490$) for {\em all} the ranking heuristics/decisions ($m = 70$) in the~\prose\ codebase for spreadsheet processing DSL. At convergence (after nearly 100 hours, 250 queries), we observed that the parameters learnt by~\selftune{} improved the correctness of the system by nearly 8\% compared to the state-of-the-art results~\cite{pmlr-v89-natarajan19a} on the benchmark consisting of 740 tasks. In particular, with the learnt parameters,~\prose system achieved an accuracy of 668/740 compared to 606/740 obtained by~\cite{pmlr-v89-natarajan19a}, and 703/740 obtained by domain experts over multiple years by hand-tuning heuristics in the~PROSE codebase today. A key factor for success of~\selftune{} here is that it optimizes for the metric of interest directly, in contrast to the ML approach in~\cite{pmlr-v89-natarajan19a}. Finally, note that we couldn't apply UCB method for learning all the $70$ decision functions as it requires discretization of the entire decision-value space which leads to exponential blow-up. 

% \hspace*{-10pt}\includegraphics[width=.51\linewidth]{exps_imgs/xor_smart_vs_naive.png
%  Reward obtained vs number of calls to reward evaluation for learning the ~\texttt{XOR} function in Figure~\ref{fig:xor.png}, using the tree template (Algorithm~\ref{alg:tree}) and the na\"ive constants template (Algorithm~\ref{alg:constant}).}

%\input{validation}
\subsection{How do our algorithms compare with white-box synthesis techniques?}\label{sec:evaluation_sketch}
Here we compare \selftune{} to program synthesis techniques such as ~\sketch{}~\cite{sketchtool}, which synthesizes parameter values for integer holes using CEGIS techniques implemented on top of SAT/SMT solvers, 
and \Fermat~\cite{chaudhuri2014bridging}, which synthesizes continuous-valued parameters using numerical methods.
We also compare with a optimization algorithm implemented in the open-source \Nevergrad platform~\cite{nevergrad}. Specifically, we use TBPSA, an evolutionary algorithm that can perform well in continuous, stochastic settings~\cite{tbpsa}. While our methods allow for realstic settings of online and blackbox rewards, but to ensure fair comparison against existing methods, we restrict this set of experiment to restricted and artificial contexts where we have white-box access to the entire system, including the reward function, and the reward function can be invoked as many times as desired.

First, we generate problems of the following type, where the goal is to learn a~\implin~decision function $f(\cdot; \weights)$, with integral $\weights \in \mathbb{Z}^d$, such that the following expected reward is maximized:
% \begin{equation}
%     \min_{\weights \in \mathbb{Z}^\numfeat} L(\weights) = \sum_{i=1}^\numex \ell\big(\langle \weights, \featvals^{(i)}\rangle\big),
% \label{eqn:toyproblem}
% \end{equation}
\begin{equation}
    \max_{\weights \in \mathbb{Z}^\numfeat} \mathbb{E}[r\big( \weights \cdot \featvals \big)],
\label{eqn:toyproblem}
\end{equation} 
where the expectation is with respect to randomness in the features $\featvals$ and the reward $r(.)$ is set to be negative of loss $\ell$ (minimizing the loss is equivalent to maximizing the reward), defined below:
\begin{align}
\label{eqn:loss}
\ell_{\text{sq}}(y) := \big(y - y^{*}\big)^2, \text{ or } \ell_{\text{abs}}(y) := \big|y - y^{*}\big|. 
\end{align}
The values $y^{*}$ are chosen to be such that $y^{*} =  \weights^* \cdot \featvals$ for some fixed $\weights^* \in [0,1,2,\dots,10]^\numfeat$ which is the optimal solution to the problem~\eqref{eqn:toyproblem}. We assume a uniform distribution over $n$ features $\x$ to compute the expectation in ~\eqref{eqn:toyproblem}. We fix $n = 2d$ in all cases (which is sufficient for learning).

%The pseudo-code for implementing the optimization problem~\eqref{eqn:toyproblem} in \sketch{} is given in Algorithm~\ref{alg:sketch}. 
The \sketch{} implementation is given in Figure~\ref{fig:code-sketch} (Appendix~\ref{app:evaluation:sketch}). For~\selftune{}, we use Algorithm~\ref{alg:linear} (fixing $\delta = 0.5$ and $\eta = 2\times10^{-3}$) meant for~\implin~decision functions. %\red{change UCB to \Nevergrad/TBPSA here}
%and for the multi-arm bandit formulation we use the Upper Confidence Bound algorithm (\textsc{UCB}) with the action space $[0,1,\dots,10]^d$. 

\begin{remark}
We restrict the search space for~\sketch{} further by ensuring that $\weights^*$ values are non-negative (requirement of the tool), small and bounded; we explicitly add these bound constraints in the ~\sketch{} problem specification. See the \texttt{assert}ion in Figure~\ref{fig:code-sketch}. Furthermore, in all the experiments, we work with 4-bit integers for holes, which we know is sufficient, using the flag \texttt{bnd-cbits} in the~\sketch{} tool. Using larger-sized integers will only increase the computation time.
\end{remark}

\begin{figure}[t]
\begin{tabular}{cc}
\hspace*{-10pt}\includegraphics[width=.51\linewidth]{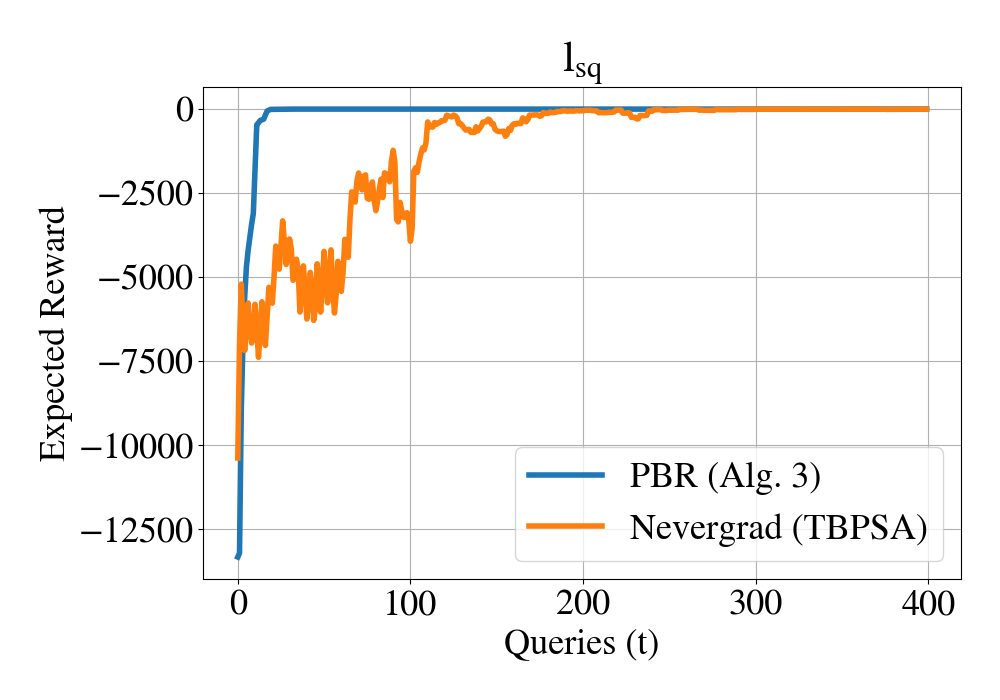}&
\hspace*{-10pt}\includegraphics[width=.51\linewidth]{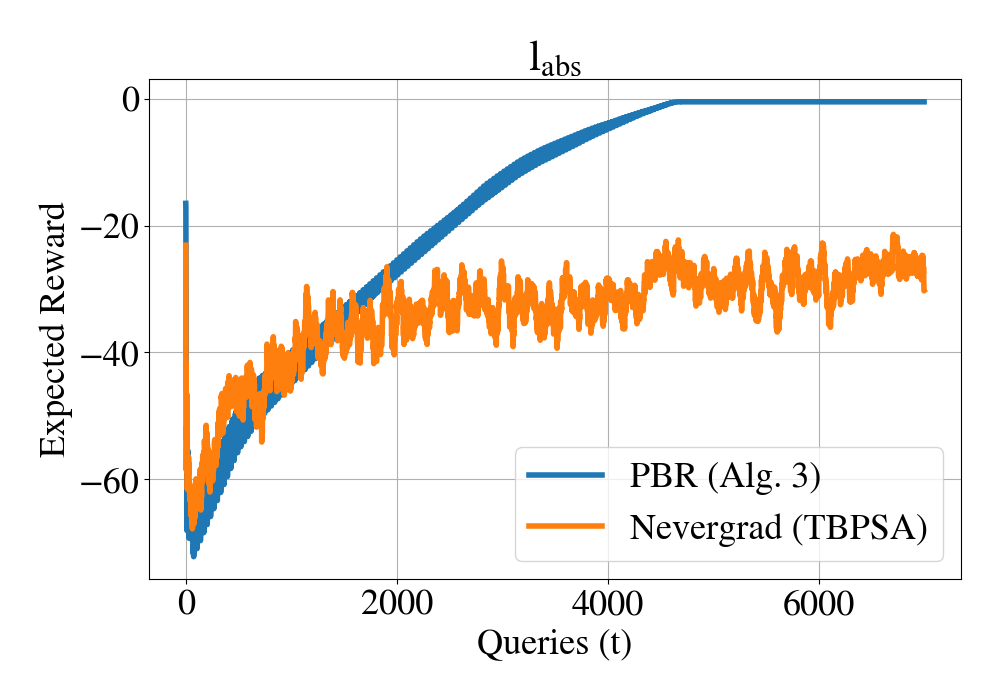}\\
(a) Squared loss $\ell_{\text{sq}}$ & (b) Abs. deviation loss $\ell_{\text{abs}}$
\end{tabular}
\caption{Expected reward (in problem~\eqref{eqn:toyproblem}) vs. number of queries to reward $r$ for loss functions in Equation~\eqref{eqn:loss}.}
\label{fig:sketchcomp}
\end{figure}

We create multiple problem sets varying number of parameters $\numfeat$ in Equation~\eqref{eqn:toyproblem}. To account for randomness in algorithms as well as in problem sets themselves, we repeat each experiment 10 times and report (a) accuracy, i.e. how often does the method solve the problem~\eqref{eqn:toyproblem} exactly, and (b) mean and standard deviation of time taken to solve. Results are presented in~Table~\ref{tab:sketch}, for the two loss functions given in~\eqref{eqn:loss}. It is clear that the search algorithm of sketching takes prohibitively long time, and fails to solve the problem even for small values of $d$ within the budget of 1 hour (and in many cases, we find that the tool prematurely fails well before the timeout despite multiple restarts with random seeds). In case of squared loss $\ell_{\text{sq}}$, sketching totally fails because it involves using multiplication circuits in the back-end SAT problem. Our algorithm solves every problem instance (as guaranteed by Theorem~\ref{thm:linear}). Even the general-purpose, continuous, black-box optimization algorithm of~\Nevergrad{} fails to compute the exact solution in most of the cases within 1 hour. The progress of \Nevergrad{} and~\selftune{} algorithms against \# reward queries is shown in Figure~\ref{fig:sketchcomp}. 

\begin{table*}[t]
\caption{Comparison of techniques on parameter learning problem in Equation~\eqref{eqn:toyproblem}, for the two losses defined in Equation~\eqref{eqn:loss}, for increasing number of parameters (holes) $\numfeat$. The implementation of~\sketch{} is given in Figure~\ref{fig:code-sketch}. For each $d$, we create 10 problem sets and report the mean and the standard deviation of the time taken (in seconds) to solve the problems. %For each configuration of $d$ and $n$, we create 10 problem sets and report the mean and the standard deviation of the time taken (in seconds) to solve the problems. 
The number of cases successfully solved out of 10 is indicated in parentheses. Timeout is set to 1 hour.}
\label{tab:sketch}
\begin{center}
\begin{footnotesize}
\begin{tabular}{c|r|r|r|r|r|r}
\toprule
$d$ & \multicolumn{3}{c|}{$\ell_{\text{abs}}$} & \multicolumn{3}{c}{$\ell_{\text{sq}}$}\\ \hline
 & \sketch & \selftune (Alg.~\ref{alg:linear})& \Nevergrad & \sketch & \selftune (Alg.~\ref{alg:linear}) & \Nevergrad \\
\hline
 2 & 0.49 $\pm$ 0.04 (\good{10}) & 0.04 $\pm$ 0.01 (\good{10}) & 298.66 $\pm$ 555.73 (\bad{5}) & 70.8 $\pm$ 73.95 (\good{10}) & 0.01 $\pm$ 0.00 (\good{10}) & 0.41 $\pm$ 0.15 (\bad{3}) \\ 
 4 & 9.65 $\pm$ 7.72 (\good{10}) & 0.04 $\pm$ 0.01 (\good{10}) & - (\bad{0}) & - (\bad{0}) & 0.01 $\pm$ 0.00 (\good{10}) & - (\bad{0}) \\ 
 6 & 582.1 $\pm$ 425.0 (\bad{5}) & 0.07 $\pm$ 0.03 (\good{10}) & - (\bad{0}) &- (\bad{0}) & 0.01 $\pm$ 0.00 (\good{10}) & - (\bad{0})\\ 
 8 & - (\bad{0})  & 0.06 $\pm$ 0.02 (\good{10}) & - (\bad{0}) & - (\bad{0})& 0.02 $\pm$ 0.01 (\good{10}) & - (\bad{0})\\ 
% 10  & 10 & - (\bad{0})  & $\pm$ & $\pm$ &- (\bad{0})& & \\
%   & 20 & - (\bad{0})  & $\pm$ & $\pm$ &- (\bad{0}) & & \\
\bottomrule
\end{tabular}
\end{footnotesize}
\end{center}
\end{table*}

We tried to relax the sketch specification to not require exact solution to problem~\eqref{eqn:toyproblem} but approximate to a small additive error; however, the sketch tool still fails (See Appendix~\ref{app:evaluation:sketch}).

\label{sec:whitebox}
\begin{figure}[ht]
\begin{lstlisting}[escapechar=!,language=csh,morekeywords={public,double,return,bool,assert,??},numbers=left,basicstyle=\footnotesize\ttfamily,xleftmargin=5.0ex]
double Thermostat(double lin, double ltarget) {
    double h = ??(0, 10);    double tOn = ltarget + ??(-10,0);
    double tOff = ltarget + ??(0, 10);    bool isOn = false; double K = 0.1;
    assert(tOn < tOff; 0.9); assert(h > 0; 0.9);    assert(h < 20; 0.9);
    for (int i=0; i<40; i = i + 1) {
        if (isOn) {
            curL = curL + (h - K * (curL - lin)); if (curL > tOff) isOn = false;
        } else {
            curL = curL - K * (curL - lin); if (curL < tOn) isOn = true;
        }
        assert(curL < 120; 0.9);
    }
    return abs(curL - ltarget);
}
\end{lstlisting}
\vspace{-0.5cm}\caption{Thermostat sketch.}
\label{fig:sketch-thermostat}
\end{figure}

Next, we compare \selftune{} with the \Fermat tool \cite{chaudhuri2014bridging} for numerical parameter synthesis with quantitative as well as boolean specification, on the synthesis benchmarks studied in~\cite{chaudhuri2014bridging}. In particular, their technique has white-box access to the cost function that is part of the sketch. The input to \Fermat is (1) a sketch implementing a cost function that returns a real value,
with some holes for constants (that need to be learned)
and probabilistic assertions,
and (2) a distribution of the input values.
The goal is to learn values for the constants minimizing: (a)~the probability that the assertions fail,
and (b)~the expected value of the function (which computes some notion of \emph{error}).

Figure~\ref{fig:sketch-thermostat} shows example of such a function sketch;
the \Thermostat function takes two probabilistic inputs (\texttt{lin} and \texttt{ltarget}),
has three holes at lines 2-4 ($\texttt{??}(c_1,c_2)$ denotes a hole with additional insight that the parameter is likely to lie in the interval $[c_1,c_2]$~\cite{chaudhuri2014bridging}),
contains four assertions at lines 6, 7 and 16,
and returns a double value representing the error.
Given the distribution of the input variables,
the goal is to find the values for the three constants, such that probability for assertions failing and
the expected error are minimized.

\textbf{Setting up the problem in~\selftune.} We model this setting in \selftune, using Algorithm~\ref{alg:constant}, as follows.
We generate $n = 10,000$ inputs for the cost function by sampling the input distribution. We define the loss as the squared error (the value returned by the sketch cost function) plus a very high additive cost for any assertion violation (we use 1000), and set the reward $r$ to negative of this loss.
% Next, we iterate the cycle: (1) predict constant values using Algorithm~\ref{alg:constant}~\footnote{
%     For each parameter we pick the initial values by uniformly sampling its likely
%     interval ($[c_1, c_2]$ discussed above).
% },
% (2) run the function with given predicted constants (over all sampled inputs),
% (3) 
We consider Algorithm~\ref{alg:constant} converged when there is no improvement in the reward for 100 iterations.
Following the experiment in \citet{chaudhuri2014bridging}, we run both \Fermat~and~\selftune{}~80 times per problem, varying the initial random seed where the search starts.

\begin{table}
    \caption{Comparison of \selftune{} with \Fermat on two synthesis tasks from~\cite{chaudhuri2014bridging}.\vspace*{-5pt}}
    \begin{center}
    \begin{footnotesize}
    \begin{tabular}{c|cc|cc|cc}
        \toprule
        Benchmark
            & \multicolumn{2}{c|}{Time (minutes)}
            & \multicolumn{2}{c|}{\# iterations}
            & \multicolumn{2}{c}{Error}
            \\ \hline
         & \Fermat & \selftune (Alg.~\ref{alg:constant}) & \Fermat & \selftune (Alg.~\ref{alg:constant})
            & \Fermat & \selftune (Alg.~\ref{alg:constant})  \\ \hline
        \Thermostat &
        119.29 $\pm$ 98.07 & 5.15 $\pm$ 3.62 & 2999.43 $\pm$ 91.93 & 781.11 $\pm$ 546.42 & 4.61 $\pm$ 2.32 & 2.74 $\pm$ 0.71 \\
        \Aircraft &
        333.57 $\pm$ 4.53 & 6.66 $\pm$ 1.98 & 4232.59 $\pm$ 313.65 & 994.50 $\pm$ 273.29 & 18.09 $\pm$ 2.89 & 25.22 $\pm$ 1.56 \\
        \bottomrule
    \end{tabular}
    \end{footnotesize}
    \end{center}
    \label{tab:whitebox-results}
\end{table}

\textbf{Results.}
We compare the tools on two problems from \citet{chaudhuri2014bridging}:
\Thermostat (in Figure~\ref{fig:sketch-thermostat}) and \Aircraft (details in Appendix~\ref{app:fermat-sketch}).
We look at the time taken and the number of reward queries required for the algorithms to converge to some good solution, the mean error over all 80 runs (the error of a run is expected error over all inputs). The results are given in Table~\ref{tab:whitebox-results}. We observe that \Fermat (a) takes much longer time to converge than \selftune{}, and (b) uses many more reward queries to converge. Our algorithm outperforms~\Fermat in the~\Thermostat sketch in terms of the quality of the constants learnt, but is worse in the~\Aircraft sketch. On the other hand, \Fermat requires white-box access to the code,
and its applicability is very limited.

\subsection{Does exploiting structure in the decision functions really help?}
\label{sec:structure}
We now present examples to show that our algorithms can indeed exploit structure in decision functions, in order to learn a good solution to the synthesis problem with small number of queries to the reward function $r$, even though the number of parameters $\numweights$ may be very large. For the case of \implin~decision functions, Theorem \ref{thm:linear} provides a theoretical guarantee for how exploiting the linear structure helps in terms of sample complexity. On the contrary, for decision trees (i.e. general~\imp~decisions), such a rigorous analysis is difficult. To this end, we consider three different ~\imp~decision function instances, and empirically evaluate the tree learning Algorithm~\ref{alg:tree} against posing the tree learning problems as sketches, treating every numerical parameter in the tree as a hole, and applying the Algorithm~\ref{alg:constant} that disregards any structure.

\begin{figure}
% \begin{tabular}{cc}
% \hspace*{-10pt}\includegraphics[width=.51\linewidth]{exps_imgs/xor_smart_vs_naive.png}&
% \hspace*{-10pt}\includegraphics[width=.51\linewidth]{exps_imgs/slates_smart_vs_naive.png}\\
% (a)&(b)
% \end{tabular}\vspace*{-10pt}
    \begin{subfigure}{.45\textwidth}
        \includegraphics[width=\linewidth]{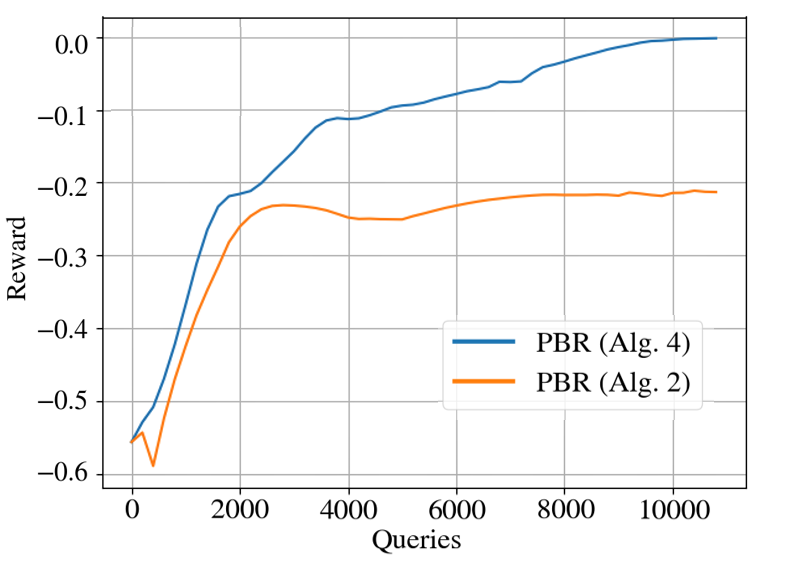}
    \end{subfigure}
    \begin{subfigure}{.49\textwidth}
        \includegraphics[width=\linewidth]{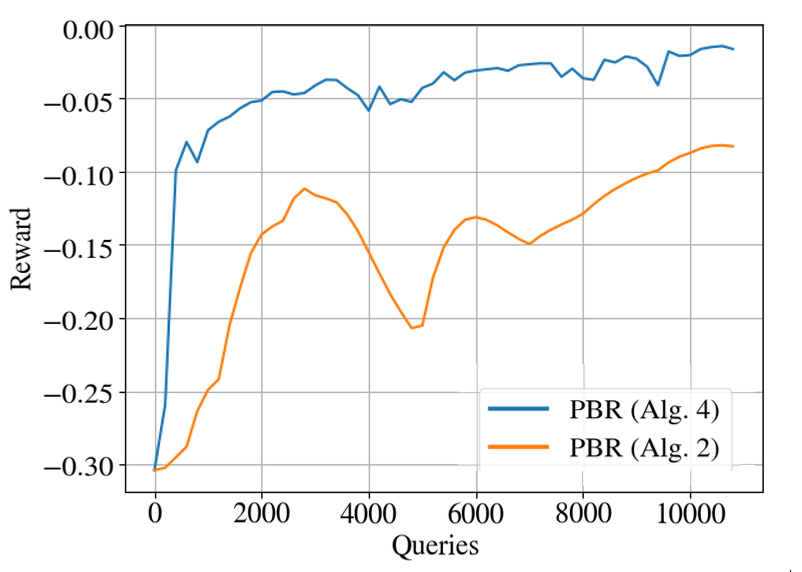}
    \end{subfigure}
\caption{Reward vs number of queries to reward $r$ for learning two different tree decision functions using Algorithm~\ref{alg:tree} and   Algorithm~\ref{alg:constant}. {\bf (a)}: (\textsc{Xor} in Figure~\ref{fig:xor.png}) The learnt decision function is given in Figure~\ref{fig:code-xor} (Appendix \ref{app:evaluation}); {\bf (b)}: (\textsc{Slates} in Figure~\ref{fig:slates.png}) The learnt decision function is given in Figure~\ref{fig:code-slates} (Appendix \ref{app:evaluation}).}
\label{fig:constant_vs_tree_synth}
\end{figure}

First, we consider an~\imp~decision function that has the~\textsc{Xor} structure, modeled on two features distributed as given in Figure~\ref{fig:xor.png} (in Appendix~\ref{app:evaluation}). Next, we consider a more complex hypothetical tuning problem \textsc{Slates}, where we wish to learn a piece-wise constant threshold function based on the values of two observed features, that decides what the threshold for the input should be (these type of heuristics are common in systems). In the setup, we have 6 different possible thresholds that depend on the features as shown in Figure \ref{fig:slates.png} in Appendix \ref{app:evaluation}. In both the cases, the reward is negative of squared loss $\ell_{\text{sq}}$ defined in Equation~\eqref{eqn:loss} between the actual and predicted value. Figure \ref{fig:constant_vs_tree_synth} compares the performance of learning constants directly vs. exploiting the tree structure to learn the decisions. In both cases, we observe that indeed exploiting structure helps converge to a significantly better solution using much fewer reward queries. The height of the tree learnt in \textsc{Xor} is 2, hence we learn $d=13$ parameters using both algorithms. For \textsc{Slates} we learn a tree of height 3, where $d=29$. However, in both problems, the sample complexity of Algorithm \ref{alg:tree} is proportional to $m=1$ since only one decision is made.

\begin{figure}[ht]
  \begin{subfigure}[b]{.45\textwidth}
    \begin{lstlisting}[escapechar=!,language=csh,morekeywords={float,return,cos,sin,acos,asin},basicstyle=\footnotesize\ttfamily]
float inversek2j(float x,float y) {
  float th2 = acos(((x*x) + (y*y) - 0.5) /
                    0.5);
  return asin((y*(0.5 + 0.5*cos(th2)) -
                 0.5*x*sin(th2)) /
               (x*x + y*y));
}
    \end{lstlisting}%
  \caption{}%
  \label{fig:parrot-code}%
  \end{subfigure}%
  \quad
  \begin{subfigure}[b]{.49\textwidth}
  \includegraphics[width=\linewidth]{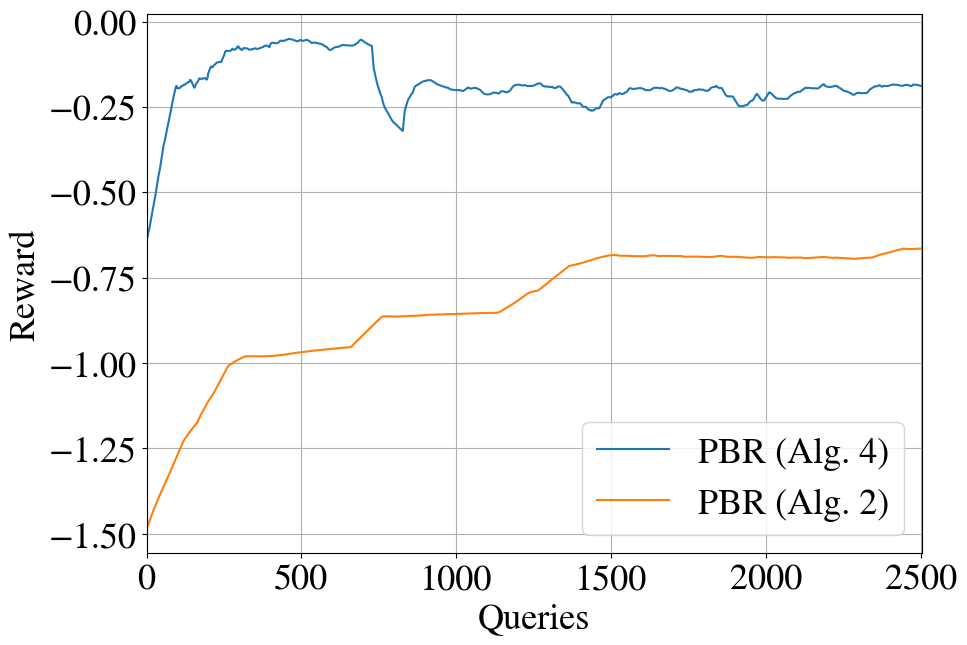}%
  \caption{}%
  \label{fig:parrot-plot}%
  \end{subfigure}%
  \caption{(a) The function to approximate from the \textsf{Parrot} benchmark.
  (b)~The rewards obtained for learning the function approximations using tree model (Alg.~\ref{alg:tree}) and constants (Alg.~\ref{alg:constant}).}
\end{figure}

Next, we consider the \textsc{Parrot} synthesis benchmark studied by~\citet{bornholt2016optimizing,esmaeilzadeh2012parrot}, where the goal is to learn a piece-wise polynomial approximation for the complex function in Figure~\ref{fig:parrot-code} (which is more efficient to execute than the trigonometric functions). %We learn two different tree models: based on the tree template (Algorithm~\ref{alg:tree}) and simple constant template (Algorithm~\ref{alg:constant}). 
Here, we compute 16 features $\{ x^i \cdot y^j \}$ for $0 \leq i,j \leq 3$ based on the input $x$ and $y$ to the function that needs to be approximated. The feature distribution is uniform over $100$ pairs $(x_i, y_i)$ sampled from the range $[-1, 1] \times [-1, 1]$. We then learn a decision tree of height $h=4$ using the same $\ell_{\text{sq}}$ loss as in the above instances. We also learn the $d$ tree parameters treating them as constants, where $d = (2^{h+1} - 1) * 16 = 496$. Figure~\ref{fig:parrot-plot} shows expected reward against the number of queries for the two algorithms. Again, we observe that exploiting structure results in the algorithm converging to much better rewards quicker. Additionally, we observe $33\%$ and $212\%$ median percent (relative) approximation error, for Algorithm~\ref{alg:tree} and Algorithm~\ref{alg:constant}, respectively.

% \subsection{Can we learn decision functions that are partly known, with smaller sample complexity?}
% \label{sec:partholes}
% \red{last priority: ajay to re-do the examples in 4.5 treating the decision functions as part holes and part knowns.}

\section{Related Work}
\label{sec:related}
\textbf{A/B and Multiworld Testing.} A/B testing methodology is commonly used in many disciplines and domains (medicine, especially) for understanding the effects of treatments in a controlled setting. It has been increasingly adopted in understanding the effects of parameters in systems and software~\cite{kohavi2017online,optimizely}. Performing randomized experiments on real systems, gathering data from control and treatment groups, analyzing how the parameters affect the metrics of interest, and making decisions can be disruptive, laborious, and prohibitively expensive in terms of time (especially when the number of parameters is large and continuous-valued). Multiworld testing (MWT) significantly improves upon A/B testing methodology by reusing data and context collected from a deployed system to estimate the outcome of many A/B tests without running separate tests for each; however, it still does not scale well with respect to number of parameters. The state-of-the-art MWT framework known as~\decisionservice~\cite{agarwal2016multiworld}, and is primarily intended for making a single categorical decision (such as recommending ads or news articles to users), using contextual bandit algorithms~\cite{bietti2018contextual}. As we demonstrated in Section~\ref{sec:prose}, the multi-arm bandit algorithms scale poorly when the multi-dimensional action space is discretized. There is recent work on extending the techniques to continuous-valued parameters~\cite{krishnamurthy2019contextual} and multiple parameters. However, these algorithms are yet to be incorporated into tools for efficient parameter tuning.

\textbf{Reinforcement Learning (RL)} is a popular and widely-used approach~\cite{sutton2011reinforcement} for online learning of actions/decisions in order to maximize some long-term reward. \smartchoices\ by Google~\cite{carbune2018smartchoices} is a recent example of an RL-based framework, which is primarily intended for tuning parameters in software. Similar to~\selftune, their interface is developer-friendly and intuitive. Since they rely on standard RL techniques and complex neural network models, the sample complexity and reward computations required are generally high and the framework forgoes interpretability which is a key differentiating aspect in our work. Another recent line of RL research involves learning interpretable policies (as programs)~\cite{verma2019imitation,verma2018programmatically} that exploit gradient-based techniques for finding a policy that optimizes some expected reward function and combinatorial search techniques for inferring the program corresponding to the policy.

\textbf{Configuration Optimization in Software.} An important line of related work in empirical software engineering involves automatic parameter tuning and large-scale configuration optimization~\cite{sayyad2013scalable,siegmund2015performance,guo2018data,kaltenecker2019distance,bao2019actgan}. Our work chiefly differs from a majority of approaches in this line of work in that (a) we focus on synthesizing interpretable code for computing decisions, via learning parameters, in software, and (b) our algorithms work with arbitrary and complex reward functions in practice. In contrast, for example, ~\citet{siegmund2015performance} is much more restrictive in applicability --- they assume a certain functional form of cost function/rewards, whereas our \selftune~ framework supports non-functional developer-defined rewards. Pure ML based techniques such as~\cite{bao2019actgan} use rather complex neural network models to determine optimal parameters/decisions given certain workload to the system; whereas we focus on synthesizing interpretable decisions (i.e. \textit{how the decision is made given the workload}) by observing and exploiting the structure in heuristics written by programmers.

\textbf{Program Synthesis, Sketching.} Automatic synthesis of interpretable programs from specification, such as input-output examples~\cite{gulwani2011automating,gulwani2017programming,padhi2017flashprofile,polozov2015flashmeta,gulwani2019quantitative}, or sketches (partial programs)~\cite{bornholt2016optimizing,chaudhuri2014bridging,solar2006combinatorial} is a flourishing line of research. Data extraction/formatting applications significantly benefit from these techniques~\cite{iyer2019synthesis}. Of particular relevance is the work by~\citet{chaudhuri2014bridging} and by~\cite{bornholt2016optimizing} that involves optimizing quantitative specification (reward/cost function) besides satisfying boolean specification. However, unlike~\selftune, these approaches need white-box access to reward function and are often limited in the type of reward functions they can handle (as discussed in Section~\ref{sec:whitebox}). In particular, the synthesis framework of~\cite{bornholt2016optimizing} needs defining careful metasketch specification (with additional information about the cost function like the gradient) for efficient synthesis, otherwise it reduces to the standard sketching problem.

\textbf{Differentiable programming} is a related field in that it studies programs that are end-to-end differentiable, and therefore can be optimized via automatic differentiation techniques. An example of such a class of differentiable programs are the neural networks and several tools exist to learn these models (e.g.~\textsf{TensorFlow}~\cite{tensorflow2015-whitepaper}). However, these models are not interpretable and require a lot of training data. Developing general purpose differentiable languages, and techniques to train such models with smaller amounts of training data are open areas of research. 
\section{Conclusion and Future Work}\label{sec:conclusions}
We formalized a novel problem -- Programming By Rewards (PBR) -- where the goal is to synthesize decision functions from an imperative language that optimize programmer-specified reward metrics. Our technical contribution is the use of continuous optimization methods to perform this synthesis with low sample complexity and low update complexity. Section 5.2 showed that the approach is able to efficiently synthesize decision functions in the PROSE code base (an industrial-strength program synthesis engine) with only $\approx 250$ reward function calls and is competitive with respect to hand-tuned heuristics developed over many man-years, and outperforms prior approaches designed specifically to tune these heuristics. 

We see several directions for future work. First, our current implementation of PBR accepts user guidance at a very coarse granularity --the user can say that the function is a constant, linear function or decision tree. While this has been sufficient for the case studies we have done so far, we believe that we can further 
improve the sample complexity if the user can give us a sketch of the tree they expect to synthesize and initial values of parameters they expect. We plan to pursue this direction both in terms of theoretical guarantees as well as empirically, with case studies. Next, while Theorem 3 gives a precise bound for Algorithm 3 in the context of learning linear functions, a corresponding bound for learning trees is still open. Finally, as stated in Section 4.5, while PBR works well in practice for the case studies we have tried, even when the reward functions are not continuous, we would like to understand the nature of reward functions that arise in practice and characterize formally the assumptions under which PBR is guaranteed to work.

\clearpage
\newpage
\bibliography{main}

\clearpage
\newpage
\appendix
\section{Appendix to Section~\ref{sec:learning}: Learning Decision Trees}\label{app:tree}

In this section, we will give the neural network architecture and observe that, under some structural constraints, it implicitly represents decision trees. This equivalence forms the basis of implementing Algorithm~\ref{alg:tree} efficiently in practice, which will be discussed subsequently.
%This equivalence forms the basis of our algorithm to learn decision trees in both supervised and black-box settings (see Section~\ref{sec:learning}).  

Given a decision tree $f(.; \treemodel)$ of height $h$, we want to be able to capture the sequence of \textit{binary tests} performed at the nodes of the tree along a root-to-leaf path. It is challenging and non-trivial to formulate the navigation function that encodes sequential decision making. A simple and key observation is that we can encode the tree paths via a three-layer neural network, which is referred to as \ent{} (the term was coined in ~\cite{sethi1990entropy}), in a way that the network computes the same function as the decision tree $f$. 

\begin{figure*}
	\begin{center}
		\includegraphics[scale=0.48]{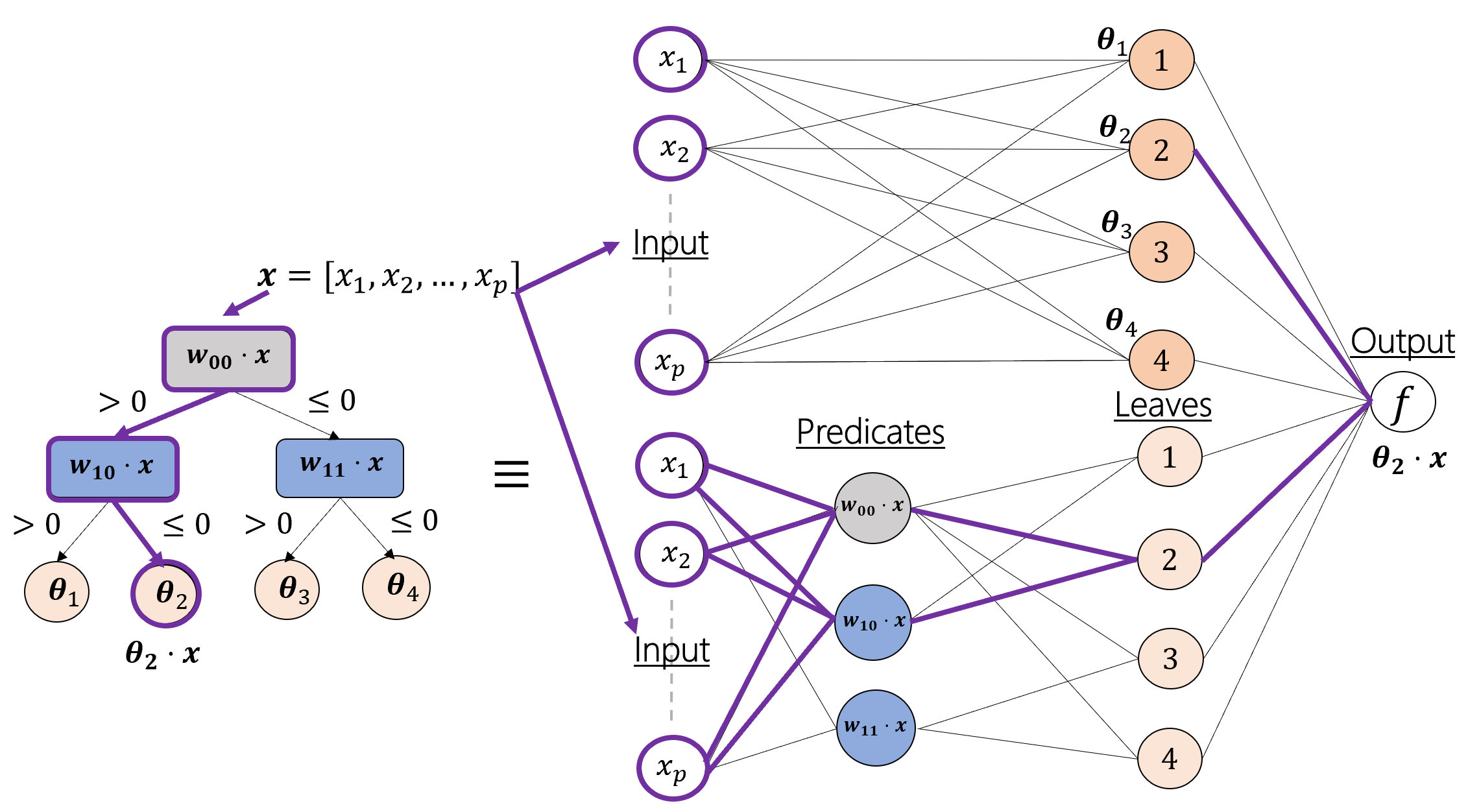}
		\caption{On the left is a decision tree of height 2. On the right is the equivalent \ent{} with activation functions and weights as per the description in Definition~\ref{def:entnet}. The path highlighted on the tree traversed by a given input example $\x$ has a one-to-one correspondence to a set of neurons that fire in the \ent{}.}
		\label{fig:net}
	\end{center}
\end{figure*}

Figure~\ref{fig:net} gives an illustration of \ent{}. Informally, the first layer of the neural network encodes all the decisions (binary tests) made in the tree; thus it has a total of $2^h-1$ neurons ($=$ number of internal nodes of a complete binary tree of height $h$). The second layer encodes the \textsc{And} of the decisions capturing the root-to-leaf paths; it has two neurons per leaf, where one neuron is to encode the path and the other neuron is to encode the linear model $\btheta$ in the leaf node. The final layer encodes the output of the tree. The activation functions of the neurons are chosen suitably so that there is a one-to-one mapping between root-to-leaf paths and the set of neurons that fire in the network for any given input example (see highlighted path and neurons in Figure~\ref{fig:net}). We give this construction precisely in Definition~\ref{def:entnet}. %(Appendix \ref{app:learning}).
In the following, we give the construction of~\ent{}~precisely. Assume without loss of generality that the tree is a complete binary tree (otherwise, we can extend the tree with $\langle\mathbf{0}, \x\rangle \geq 0$ nodes, and appropriately shift the leaves down to make it a complete binary tree).
\begin{defn} [\ent{}]
\label{def:entnet}
Let $\mathbbm{1}_\ell$ denote the path indicator function for a decision tree of height $h$ defined as $\mathbbm{1}_\ell(i,j) = 1$ if $(i,j)$th node lies in the path leading from the root to the $\ell$th leaf node, for $\ell = \{0,1,\dots,2^h-1\}$.  For each non-root $(i,j)$th node in the tree, define $g_{ij} = +1$ if the node is the left child of its parent node, $g_{ij} = -1$ otherwise. The \emph{\ent{}} architecture takes as input $\x \in \mathbb{R}^\numcontext$ and comprises: \\
(1) The ``predicate layer'' with $2^{h}-1$ neurons corresponding to the internal nodes of the tree. Let $\netw_{ij}^{(1)} \in \mathbb{R}^{\numcontext}$ denote the weights of the $(i,j)$th neuron in this layer corresponding to the $(i,j)$th internal node. Set the activation function as: \[z^{(1)}_{ij} = \texttt{sign} (\netw_{ij}^{(1)}\cdot \x + b_{ij})\ .\] Let $\z^{(1)} \in \{+1,-1\}^{2^h-1}$ denote the vector-valued output from this layer. \\
(2) The ``leaf layer'' has two parallel sets of neurons, each with $2^h$ neurons corresponding to the leaves of the tree:
\begin{enumerate}
\item[(a)] for the first set, let $\netw^{(2,1)}_{k} \in \mathbb{R}^{2^h-1}$ denote the weights of the $k$th neuron in this set (corresponding to the $k$th leaf node). For $i = \{0, 1, \dots, h-1\}, j \in \{0,1,\dots,2^{i}-1\}, $ set:
\[ 
  \netw^{(2,1)}_{k}(2^{i}+j-1) = \sum_{j' \in \{0,1,\dots,2^{i+1}-1\}} g_{i+1,j'}\ .\ \mathbbm{1}_k(i+1,j'), 
\]
and set the activation function of the neuron to: 
\begin{equation}
z^{(2,1)}_{k} = \max(\netw_{k} \cdot \z^{(1)} - h + \epsilon, 0),
\label{eqn:relu}
\end{equation} for some $\epsilon > 0$. Let $\z^{(2,1)} \in \mathbb{R}_+^{2^h}$ denote the vector-valued output from this layer.
\item[(b)] for the second set, let $\netw^{(2,2)}_{k} \in \mathbb{R}^{p}$ denote the weights of the $k$th neuron in this set (corresponding to the $k$th leaf node), where each neuron is connected to the $p$ input variables. Set the activation function as: \[z^{(2,2)}_{k} = \netw^{(2,2)}_k \cdot \x + b'_{k}\ .\] Let $\z^{(2,2)} \in \mathbb{R}^{2^h-1}$ denote the vector-valued output from this layer.
\end{enumerate}
(3) The output layer has one neuron whose output is set to the following non-linear activation: 
    \[ z^{(3)} = \frac{1}{\epsilon} (\z^{(2,1)} \cdot \z^{(2,2)}) \in \mathbb{R}. \]
\end{defn}

Next, we formally show that the~\ent{} architecture encodes the decision tree computation exactly. 
%that $\weights_{ij} \in \mathbb{R}^{\numcontext}$ and $b_{ij} \in \mathbb{R}$ are the linear weights and bias of the node $j$ at depth $i < h$.

% \begin{proposition} [Every Decision Tree is an \ent{}]
% For any given decision tree $f(.; \treemodel)$, let \netmodel{} denote the parameters of the \emph{\ent{}} architecture described in Definition~\ref{def:entnet} (or in Definition~\ref{def:entnetreg} for regression). Choose $M = 1$, and $\beta_{ij0} = 1, \forall i, j$. Let $\fent(.; \netmodel)$ the function it computes. We have, 
% \[f(\x; \treemodel) = \fent(\x; \netmodel), \forall \x.\]
% \end{proposition}

\begin{lem} [Every Decision Tree is an \ent{}]
\label{prop:treetonet}
For any given decision tree $f_T(.; \treemodel)$ (in Definition~\ref{def:DT}), there is an \emph{\ent{}} architecture described in Definition~\ref{def:entnet} with a particular choice of model parameters, say $\netmodel$, that computes the same function as the tree. Let $\fent(.; \netmodel)$ denote the function computed by the \emph{\ent{}}. We have, 
\[f_T(\x; \treemodel) = \fent(\x; \netmodel), \forall \x \in \mathbb{R}^p.\]
\end{lem}

% \begin{proof}
% In the~\ent{} architecture described in Definition~\ref{def:entnet}, consider the path function $\mathbbm{1}_p$ and $w_{ij}$ corresponding to the given decision tree. For $i \in \{0,1,\dots,h-1\}, j \in \{0,1,\dots,2^i-1\}$, set the predicate layer weights (Definition~\ref{def:entnet} (1)) as follows. \[\netw^{(1)}_{ij} = \w_{ij}, \]
% where $\w_{ij}$ are the weights of the internal nodes of the given tree. In the activation function, choose $b_{ij}$ to be the corresponding biases of the nodes $\forall i, j$. 

% Now, for classification, consider the output layer (3) in Definition~\ref{def:entnet}. For $j = \{0,1,\dots,2^h-1\}$, set:
% \[
%     \netw^{(3)}_k(j) = \btheta_j (k)
% \]
% where $\btheta_j (k)$ denotes the function value/score/probability for the $k$th class realized at the $j$th leaf node of the given tree. It is straight-forward to verify that the output of the network is %(a) $\z^{(3)} \in \{0,1\}^\numclasses$, and (b) $\z^{(3)}(k) = 1$ if and only if $f(\x; \treemodel) = k$. 
%  $\fent(\x; \netmodel) = \btheta_j$ if and only if  $f(\x; \treemodel) = \btheta_j$.

% For regression, consider the output layer (3) in Definition~\ref{def:entnetreg}. Set the weights $\netw^{(3)} \in \mathbb{R}^{2^h}$ of the neuron as:
%  \[
% \netw^{(3)} (j) = \theta_j, \ j \in \{0,1,\dots,2^h-1\},
% \]
% where $\theta_j \in \mathbb{R}$ are the corresponding predictions of the decision tree. The proof is complete.
% \end{proof}

Not only is every decision tree encoded by some \ent{}, but the vice versa also holds. Thus, the class of \entdesc{}s and the class of decision trees (and therefore ~\imp, from Lemmas~\ref{lem:equiv1} and~\ref{lem:equiv2}) are identical. 
\begin{lem} [Every \ent{} is a Decision Tree]
\label{prop:nettotree}
	For every neural network with 3 layers, with constraints on the weights and activations as described in Definition~\ref{def:entnet}, there is a corresponding decision tree which represents the same function.
\end{lem}
% \begin{proof}
% This holds by arguments similar to the proof of Proposition~\ref{prop:treetonet}. The corresponding tree parameters are precisely given in Algorithm~\ref{algo:infertree}.
% \end{proof}
\subsection{Updating $\fent$ model}
\label{sec:treeupdate}
Computing the gradient in the update Step 5 of Algorithm~\ref{alg:tree}, with $\fent$ in lieu of $f$ given the equivalence above, and ensuring that the updates maintain a valid decision tree, is still challenging for the following reasons.
\begin{enumerate}
\item The activation function in the first (predicate) layer given in Definition~\ref{def:entnet} (1) is the \texttt{sign} function, which is discontinuous and non-differentiable. We need a relaxation of the activation function to allow learning.
\item The structural constraints on the weights of the leaf neurons given in Definition~\ref{def:entnet} (2) impose that the weights be \textit{integral}, and in particular, $\{1,-1\}$-valued (see~Figure~\ref{fig:net}). This is an inviolable constraint, to ensure that the root-to-leaf paths are preserved in the learned neural network model.
\item Finally, we need only one of the leaf neurons to fire in the leaf layer -- this is ensured in Definition~\ref{def:entnet} (3), by the activation that thresholds the output at $h$. This essentially implements the \textsc{AND} of the predicates, which theoretically holds only when we use \texttt{sign} function in the predicate layer. Thus this requirement is at odds with any relaxation that we might want to use to mitigate the first challenge.
\end{enumerate}

%A feasible approach to tackle these challenges is to work with a less constrained space of ~\ent{}s, and devise a way to convert the learned parameters~$\netmodel${} into corresponding decision tree parameters~$\treemodel$. 
%This way, we can leverage stochastic gradient descent-style algorithms as well as common tricks such as batching and momentum to learn in the space of neural networks. Now, the main issue is to ensure that the conversion of model parameters does not compromise the performance/generalization of the resulting tree model. Even in simple problems, na\"ive relaxations can fail badly; in the following, we develop an annealing technique for gradually moving from the space of neural networks to decision trees for solving~\eqref{eqn:opt2} using SGD algorithm.
We address these challenges as follows:\\
\textbf{(I)} First, we relax the sign function using a scaled sigmoid function as the activation function for the predicate layer. In particular, for appropriately chosen $s > 0$, we define:
\begin{align}
\sigma_s (a) := \frac{1}{1 + \exp(-s\ . \ a)},~\text{and}  \nonumber\\
z^{(1)}_{ij} = 2\sigma_s (\langle\netw_{ij}^{(1)}, \x\rangle + b_{ij}) - 1.  
\label{eqn:sigmoid}
\end{align}
Note that for sufficiently large $s$, the above activation behaves like the \texttt{sign} function. In practice, selecting $s$ can be tricky. If we choose $s$ very large, then it will be hard for the optimization to proceed and it will likely get stuck in a poor solution. On the other hand, using a small $s$ will lead to violation of constraints as discussed in challenge~(3) above. 
%We observe in our experiments that such a relaxed activation function can fail poorly for any fixed choice of scale factor $s$, even on a simple \textsc{XOR} problem (Figure~\ref{fig:xor}) that can be \textit{perfectly} classified with a decision tree of height 2.

Even though the optimization tends to converge to a good $\netmodel$ with a small $s$ (when the model is effectively a neural network), the resulting decision tree model $\treemodel$ after conversion can be quite poor. On the other hand, we also observe that it is important for $s$ to be not too large (when the model is effectively a decision tree), otherwise the optimization cannot proceed. This motivates us to devise a careful scheduling of choices of $s$ from smaller values to larger values, letting the activation function approach the \texttt{sign} function, moving gradually from the space of neural networks to decision trees in the process.

\textbf{(II)} Second, we satisfy the integral constraints on the weights of the leaf neurons by simply anchoring the weights during training. Note that the number of non-zero weights in the second layer of \ent{} is only $h\ . 2^h$, as against standard dense neural network which has $O(2^{2h})$ connections.

\textbf{(III)} The above two ideas by themselves still do not guarantee convergence of the optimization to a good decision tree solution. In fact, the third challenge remains unaddressed. In particular, when we start with a small $s$ in the sigmoid activation in the beginning of the training, there is no guarantee that any ReLU activated neuron in the third layer, given in Eqn.~\eqref{eqn:relu}, will fire at all, if we set $\epsilon$ to be very small. On the other hand, if we set $\epsilon$ to be large, many neurons will fire, thus violating the basic property of a decision tree (each example traverses a unique root-to-leaf path). Therefore, we follow the opposite schedule for $\epsilon$; we start with a large $\epsilon$ ensuring that gradient information propagates through these neurons, and gradually decrease $\epsilon$. 
\section{Proofs}
\label{app:proofs}
\subsection{Proof of Lemma~\ref{lem:equiv1}}
Consider the parse tree of a given decision function $f: \mathbb{R}^\numcontext \to \mathbb{R}^m$ shown below in Figure~\ref{fig:parsetree}, denoted by $\mathcal{P}$, where each internal node in the tree corresponds to a non-terminal of type $E$ (expression) or $S$ (statement), and leaf nodes correspond to real numbers (i.e. choices for $W$ in the expansion of $E$). We start by constructing an equivalent parse tree $\mathcal{P}'$ for $f$, by transforming $\mathcal{P}$ in way that ensures that the computational semantics of the program is preserved. 

\begin{figure}[h]
\includegraphics[scale=0.4]{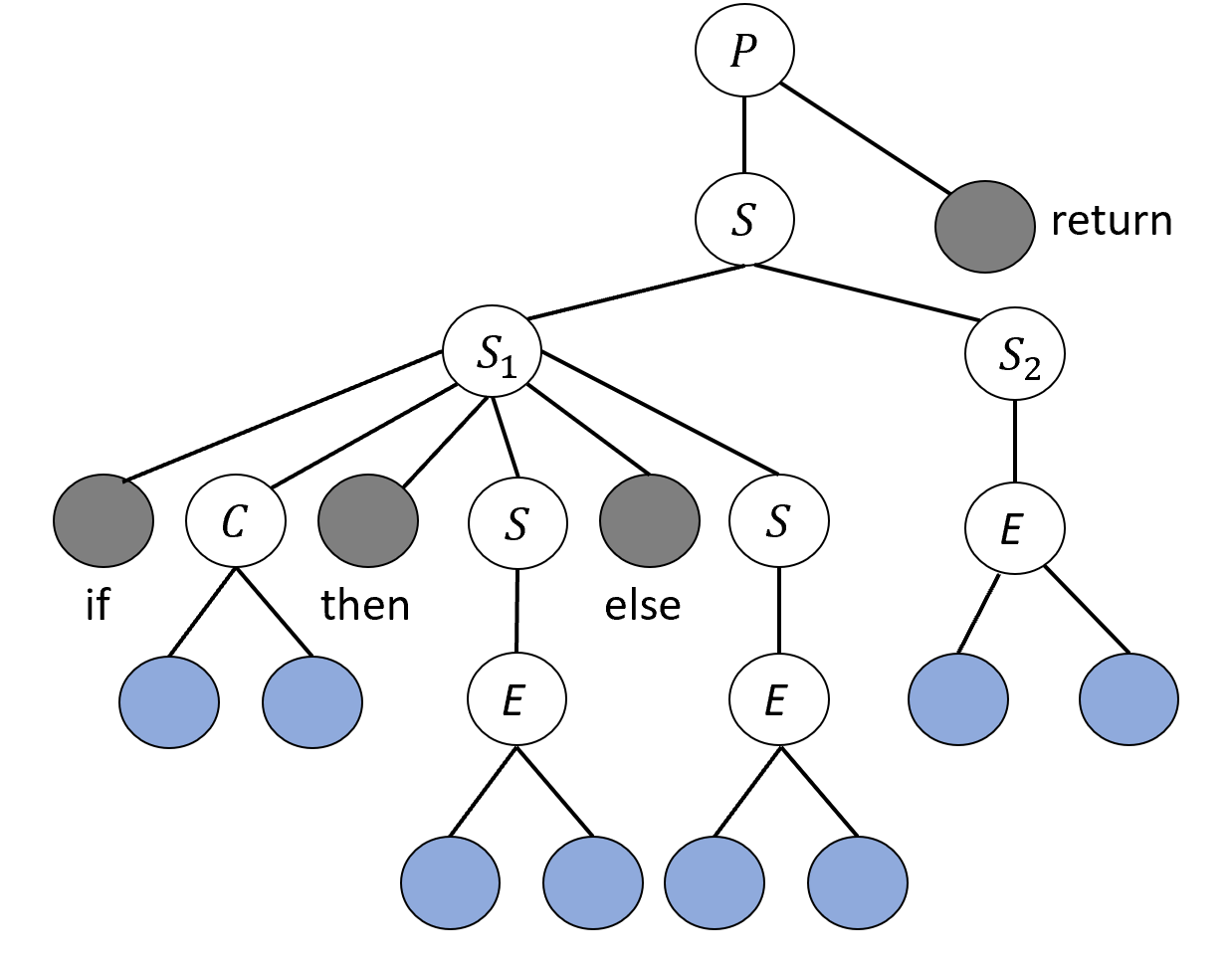}
\caption{Parse tree for a decision function in~\imp~with $p = 1$ (the blue terminal nodes correspond to choices for $W_1$ and $W_2$). The non-terminal $C$ is a short-hand for the conditional $E > 0$.}
\label{fig:parsetree}
\end{figure}

\begin{figure}[h]
\includegraphics[scale=0.4]{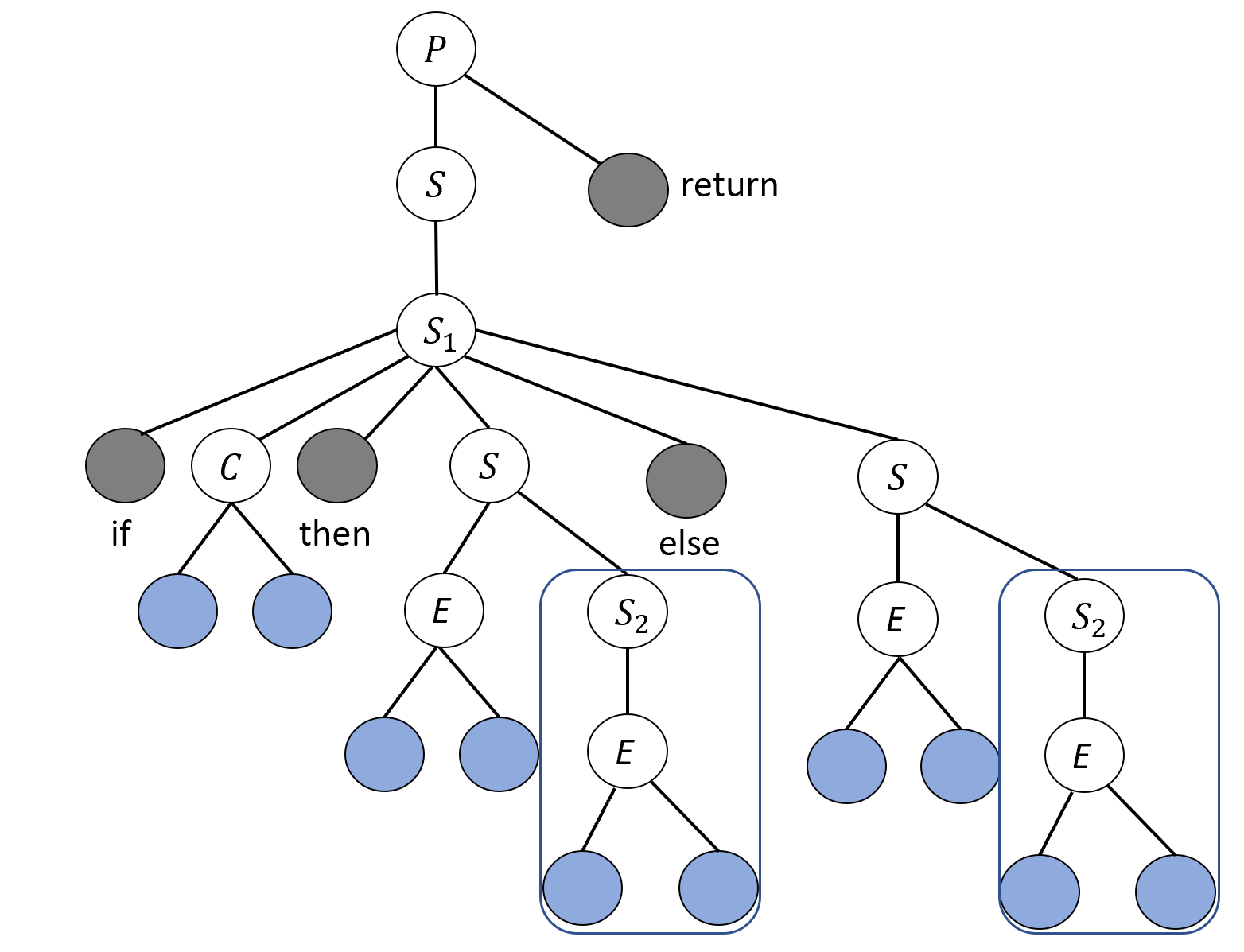}
\caption{Applying \textsc{Expand} transform to the parse tree in Figure~\ref{fig:parsetree}.}
\label{fig:parsetree2}
\end{figure}

The key transformation, called \textsc{Expand}, is as follows. For any internal node $S$ such that both of its children are of type $S$ (call them $S_1$ and $S_2$) (Figure~\ref{fig:parsetree} has one instance of this), let $\mathcal{P}_{S_2}$ denote the sub-tree of $\mathcal{P}$ rooted at $S_2$. Now, consider every root-to-leaf path in $\mathcal{P}_{S_1}$ starting from $S_1$ to the last internal node of type $S$. say $S'$,  along the path. We attach $\mathcal{P}_{S_2}$ as the second child of $S'$ (note that $S'$ has only child, because it is the last internal node). We now apply \textsc{Expand} to every such internal node $S$ such that both of its children are of type $S$. The resulting transformed parse tree $\mathcal{P}'$ after applying \textsc{Expand} to the example in Figure~\ref{fig:parsetree} is shown in Figure~\ref{fig:parsetree2}. 

Observe that $\mathcal{P}'$ and $\mathcal{P}$ are equivalent, i.e. a) every assignment statement of type $o_j = E$ for some $1 \leq j \leq m$ executed in $\mathcal{P}$, along any execution path of $f$, is also executed in $\mathcal{P}'$, b) the order of the assignments is also identical (attaching as the right child in the above transform ensures this).

With this, we will now give a straight-forward construction of decision tree $f_T:\mathbb{R}^\numcontext \to \mathbb{R}^m$ (Definition~\ref{def:DT}) that computes the given decision function $f$. The root of the decision tree of $f_T$ corresponds to the first non-terminal $S$ in the parse-tree $\mathcal{P}'$ (note that any program in \imp~ has at least one non-terminal $S$ in its parse tree). For this single-node tree $f_T$ (as yet), initialize parameters $\treemodelTheta = \{\btheta_j\} = \emptyset$. Now consider the sub-tree rooted at this $S$ in $\mathcal{P}'$ referred to as $\mathcal{P}'_S$. Note that there is no node in $\mathcal{P}'$ both of whose children of type $S$. That leaves us with the following cases:

\begin{enumerate}
    \item if the children node of $S$ in $\mathcal{P}'_S$ correspond to \textsf{if ($C$) then $S_1$ else $S_2$} then set the parameters $\weights \in \mathbb{R}^{\numcontext+1}$ of the current root node of $f_T$ to the corresponding terminals in the conditional expression $C$. Then:\\
    a) create a left child to the current root in $f_T$, copy $\{\btheta_j\}$ of the current root node, make this the current root, and repeat step (1) with $S = S_1$.\\
    b) create a right child to the current root in $f_T$, copy $\{\btheta_j\}$ of the current root node, make this the current root, and repeat step (1) with $S = S_2$.
    \item if the first (or only) child node of $S$ corresponds to assignment $o_j = E$, then set $\btheta_j \in \mathbb{R}^{\numcontext+1}$ of the current root node to the corresponding terminals in the expression $E$.
    \item if the other child node of $S$ exists and is of type $S$ (call it $S'$), then go to step (1) with $S = S'$.
\end{enumerate}

The height of the resulting tree $f_T$, say $h$, is proportional to the number of $S$ nodes in $\mathcal{P}'$ that expand to a conditional statement (note that the height of the tree grows only in step (1) above). The size of $f_T$ is exponential in $h$.

\subsection{Proof of Lemma \ref{lem:equiv2}}
This direction is straight-forward. We can synthesize the desired $f \in~\imp$ by doing a pre-order traversal of $f_T$. When we visit a leaf node, all the assignment statements of the form $o_j = E$ for $1 \leq j \leq m$ will be synthesized corresponding to $\{ \btheta_j \}$ parameters of the node.

\subsection{Proof of Theorem \ref{thm:constant1}}
We refer the reader to the proof of Theorem 3.3 of~\citet{flaxman2005online}. Note that $n$ in their notation is the same as $T$ in ours.

\subsection{Proof of Theorem \ref{thm:constant2}}
We appeal to the result in Corollary 2 of~\citet{shamir2017optimal}, that uses the two-point gradient estimator given in~\eqref{eqn:twopoint}, and matches the setting in our algorithm. 

\subsection{Proof of Theorem~\ref{thm:linear}}
We start by recalling a key lemma of \cite{flaxman2005online} that uses the online gradient descent analysis by \cite{zinkevich2003online} with unbiased random gradient estimates. We restate the result below for clarity, for the case when the reward is concave (the original result is stated for loss being convex). Let $\mathcal{B}_d(R)$ denote ball of radius $R$ in $\mathbb{R}^d$, and $\Pi_S(\weights) = \arg\min_{\weights' \in S} \|\weights - \weights'\|$.

\begin{lem}[Lemma $3.1$, \cite{flaxman2005online}]
\label{lem:kalai}
Let $S \subset \mathcal{B}_d(R) \subset \mathbb{R}^d$ be a convex set, $c_1, c_2, \dots, c_T : S \mapsto \mathbb{R}$ be a sequence of concave, differentiable functions. Let $\weights^{(1)}, \weights^{(2)}, \dots, \weights^{(T)} \in S$ be a sequence of predictions defined as $\weights^{(1)} = 0$ and 
$\weights^{(t+1)} = \Pi_S(\weights^{(t)} - \eta h^{(t)})$, where $\eta > 0$, and $h^{(1)}, h^{(2)}, \dots, h^{(T)}$ are random variables such that $\mathbb{E}[h^{(t)} \big| \weights^{(t)}] =  c_t(\weights^{(t)})$, and $\|h^{(t)}\|_2 \le G$, for some $G > 0$ then, for $\eta = \frac{R}{G\sqrt T}$, the expected regret incurred by above prediction sequence is:
\[
\max_{\weights \in S}\sum_{t = 1}^T c_t(\weights) - \mathbb{E}\bigg[ \sum_{t = 1}^T c_t(\weights^{(t)}) \bigg] \le RG\sqrt T.
\]
\end{lem}
%In our problem setting, let us first denote $\h\pred^{(t)}(\weights) = \hl_t(\pred_t(\weights;\x_t))$, for all $\weights \in \cW, \, t \in [T]$ (recall from Lem. \ref{lem:gradest}, 
We need the following assumptions:\\
(1) reward function $r$ is bounded, i.e. $r: \mathbb{R}^m \mapsto [-C,C]$, for some numerical constant $C$.\\
(2) $r$ is concave in $\weights \in \mathbb{R}^{\numparam \times \numcontext}$ (note that $r(\paramvals) = r(\weights\x)$, where $\weights \in \mathbb{R}^{\numparam \times \numcontext}$ and $\x \in \mathbb{R}^\numcontext$).\\
(3) $\max_t \|\x^{(t)}\|_2 \leq D$.

In the following, we will use the short-cut $\paramvals = f(\x; \weights) = \weights\x$ and $\paramvals^{(t)} = f(\x^{(t)}; \weights^{(t)}) = \weights^{(t)}\x^{(t)}$.

Also, in Algorithm~\ref{alg:linear}, we work with parameters $\weights \in \mathcal{W} = \mathcal{B}_d(mW)$, for some, possibly large number, $W$.

Let $\mathbf{U}$ denote the uniform distribution and $\mathcal{S}_m(1)$ denote the sphere in $\mathbb{R}^m$ of radius $1$. Now, define $\hat{r}: \mathbb{R}^m \mapsto [-C,C]$ such that $\hat{r}(\paramvals) = \mathbb{E}_{\mathbf{u} \sim \mathbf{U}(\mathcal{S}_m(1))}\big[r(\paramvals + \delta \mathbf{u})\big]$, for any $\paramvals \in \mathbb{R}^m$.
Then applying the above Lemma \ref{lem:kalai} in the setting of Algorithm~\ref{alg:linear}, on the concave function $\hat{r}$, with $h^{(t)}  = \frac{m}{\delta}\big(r(\paramvals^{(t)})\mathbf{u}\big)(\x^{(t)})^{\text{T}}$, and $\mathbf{u} \sim \mathbf{U}(\mathcal{S}_m(1))$ (note that this implies $\mathbb{E}[h^{(t)}\big| \weights^{(t)}] = \nabla_\weights  \mathbb{E}_{\mathbf{u}}\big[ r(\paramvals^{(t)} + \delta \mathbf{u}) \big]$, which can be seen from Lemma 2.1 of~\cite{flaxman2005online}), we get:

\begin{equation}
\label{eq:prf_algadv1}
\max_{\weights \in \mathcal{W}}\sum_{t = 1}^T \hat{r}(\paramvals) - \mathbb{E}\bigg[ \sum_{t = 1}^T \hat{r}(\paramvals^{(t)}) \bigg] \le \frac{m^2WDC \sqrt T}{\delta},
\end{equation}

as $\hat{r}(\cdot)$ is concave (in argument $\weights \in \mathbb{R}^{\numparam \times \numcontext}$) due to the assumption $(ii)$ above, and in this case $R \le W$, and $\|h^{(t)}\| = \|\frac{m}{\delta}\big({r(\paramvals^{(t)})}\mathbf{u}\big)(\x^{(t)})^{\text{T}}\| \le \frac{mDC}{\delta}$, so $G = \frac{mDC}{\delta}$ and hence we need to choose $\eta = \frac{W\delta}{DC\sqrt{T}}$. 

Further since $r(\cdot)$ is assumed to be $L$-Lipschitz continuous (Definition~\ref{def:lipschitz}), it follows from \eqref{eq:prf_algadv1} that, 

\begin{eqnarray*}
& \max_{\weights \in \mathcal{W}}\sum_{t = 1}^T  \big(r(\paramvals) - \delta L\big) - \mathbb{E}\bigg[ \sum_{t = 1}^T  \big(r(\paramvals^{(t)}) + \delta L\big) \bigg]  \le \frac{m^2WDC \sqrt T}{\delta},\\
\implies & \max_{\weights \in \mathcal{W}}\sum_{t = 1}^T  r(\paramvals) - \mathbb{E}\bigg[ \sum_{t = 1}^T  r(\paramvals^{(t)}) \bigg]  \le \frac{m^2WDC \sqrt T}{\delta} + 2 \delta LT 
\end{eqnarray*}
% setting $\alpha = \delta$.
We want to minimize the RHS above with respect to $\delta$, which can be achieved by setting $\delta = m\Big( \frac{WDC}{2L \sqrt T} \Big)^{1/2}$. This finally gives:
\begin{align*}
R_T = \max_{\weights \in \mathcal{W}}\frac{1}{T}\sum_{t = 1}^T  r(\paramvals) - \frac{1}{T}\mathbb{E}\bigg[ \sum_{t = 1}^T  r(\paramvals^{(t)}) \bigg]   \le \frac{2m\sqrt{2WLDC}}{T^{1/4}},
\end{align*}
which concludes the proof.

\subsection{Proof of Lemma~\ref{prop:treetonet}}
In the~\ent{} architecture described in Definition~\ref{def:entnet}, consider the path function $\mathbbm{1}_\ell$ and $g_{ij}$ corresponding to the given decision tree. For $i \in \{0,1,\dots,h-1\}, j \in \{0,1,\dots,2^i-1\}$, set the predicate layer weights (Definition~\ref{def:entnet} (1)) as follows. \[\netw^{(1)}_{ij} = \weights_{ij}, \]
where $\weights_{ij}$ are the weights of the internal nodes of the given tree. In the activation function, choose $b_{ij}$ to be the corresponding biases of the nodes $\forall i, j$. 
Set the weights of the second set of neurons in the leaf layer (2) to the linear models in the tree:
\[\netw^{(2,2)}_{j} = \btheta_j, \]
% Consider the output layer (3) in Definition~\ref{def:entnet}. Set the weights $\netw^{(3)} \in \mathbb{R}^{2^h}$ of the neuron as:
%  \[
% \netw^{(3)} (j) = \theta_j, \ j \in \{0,1,\dots,2^h-1\},
% \]
where $\btheta_k \in \mathbb{R}^\numcontext$ are the linear model coefficients in the $k$th leaf node; in the activation function, choose $b'_{k}$ to be the bias in the corresponding linear model. By construction, the resulting~$\fent$ computes the given tree function $f_T$, i.e. the weights chosen for the~\ent{} in the middle layer (Definition~\ref{def:entnet} (2 (a))) and the activation function ensure that for a path from root to some leaf node $j$, the one and only neuron that is activated (i.e. outputs $\epsilon$) in the leaf layer is the $j$th neuron (all other neurons output 0). The proof is complete.

\subsection{Proof of Lemma~\ref{prop:nettotree}}
This holds by arguments similar to the proof of Lemma~\ref{prop:treetonet}. The tree parameters corresponding to a given $\fent$ (and $h$ and $\epsilon$ in Definition~\ref{def:entnet}) are precisely given in the following Algorithm~\ref{algo:infertree}. Note here $\w(p)$ denotes $p$th entry in the vector $\w$.

\begin{algorithm*}
  \begin{algorithmic}[1]
    \Functionx{InferTree}{$\netmodel, h, \epsilon$}
    \For{$i \in \{0, 1,\dots, h-1\}$} \algorithmiccomment{predicates at internal nodes {\color{black} $\treemodelW$}}
    \For{$j \in \{0, 1,\dots, 2^{i}-1\}$}
    \State $\w_{ij} = [\netw^{(1)}_{ij}(0),\netw^{(1)}_{ij}(1), \dots, \netw^{(1)}_{ij}(\numcontext-1)]$
    \State $b_{ij} = \netw^{(1)}_{ij}(\numcontext)$
    \EndFor
    \EndFor    
    \For{$j \in \{0, 1,\dots, 2^h-1\}$} \algorithmiccomment{linear models at the leaves {\color{black} $\treemodelTheta$}}
    \State $\btheta_j (k) = \netw^{(2,2)}_j(k)$, for $k \in \{0,1,\dots,\numcontext\}$
    \State $\btheta_j (p) = b'_j$ (bias of the activation function, Definition~\ref{def:entnet} (2 b))
    \EndFor
    \State \textbf{return} $\treemodelW = \{ \weights_{ij}, b_{ij} \}, \treemodelTheta = \{ \btheta_j \}$
    \EndFunction
    \end{algorithmic}
    \caption{Inferring decision tree from \ent}
\label{algo:infertree}
\end{algorithm*}

%\section{Appendix to Section~\ref{sec:learning}}
%\label{app:learning}

\section{Appendix to Section~\ref{sec:evaluation}}
\label{app:evaluation}
% \begin{algorithm}[h!]
% \caption{\sketch{} pseudocode for Problem~\eqref{eqn:toyproblem}} \label{alg:sketch}
% \begin{algorithmic}[1]
% \Procedure{FillParams}{$n$, $d$, $\ell$}
% \State Randomly sample features $\featvals^{(i)} \in \mathbb{Z}^\numfeat$, for $1 \leq i \leq \numex$, such that $x^{(i)}_j \in [-10,10]$, for $1 \leq j \leq d$.
% %\State Choose $\weights^{*} \in \mathbb{Z}^\numfeat$, such that $\weight_j \in [0,10]$ for $1 \leq j \leq d$. \algorithmiccomment{\sketch{}\cite{sketchtool} requires holes to be non-negative.}
% \State Declare holes $\weight_j = \textbf{??}$, for $1 \leq j \leq d$
% \State Compute $\mathbb{E}[r\big( \weights \cdot \featvals \big)]$ according to~\eqref{eqn:toyproblem} for the given loss $\ell$
% \State \algorithmiccomment{$\weights^*$ is known, so we can compute $f$.}
% \State \textbf{assert} $\mathbb{E}[r\big( \weights \cdot \featvals \big)] == 0$
% \EndProcedure
% \end{algorithmic}
% \end{algorithm}
\subsection{PROSE details and additional results~(for Section~\ref{sec:prose})}
\begin{figure}
\includegraphics[scale=0.4]{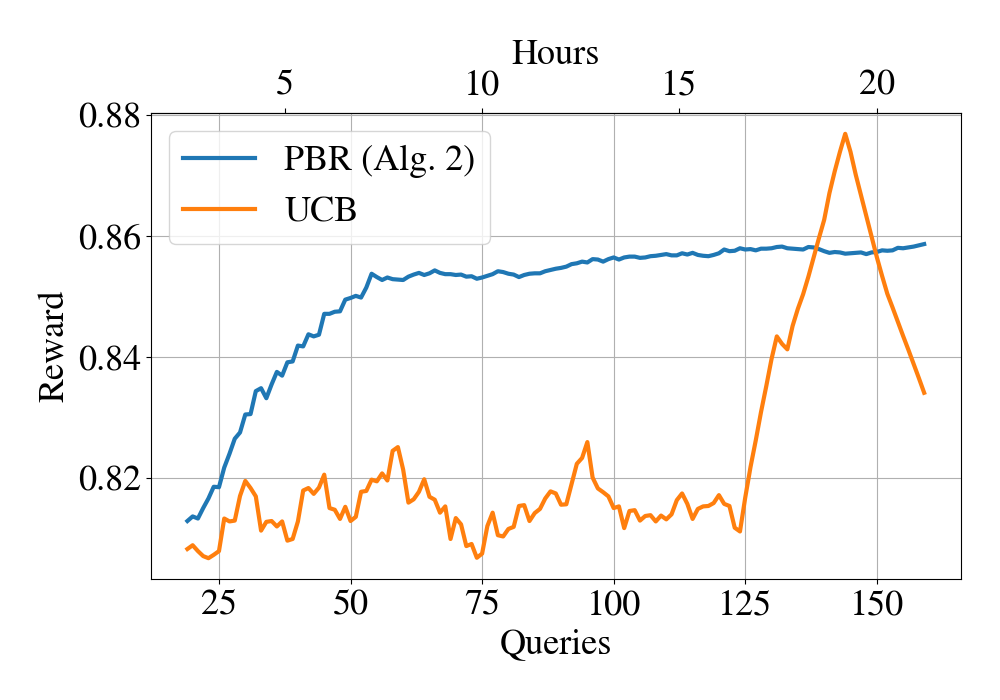}
\caption{For the same setting as Figure~\ref{fig:prose_zo_ucb_m22_mean_reward}, rewards obtained by \selftune{} and UCB for tuning \texttt{RegexPair} with $3$ parameters (Figure~\ref{fig:code-regexpair}); for the UCB algorithm, 3 parameters are discretized into {$5*5*5=125$} actions. }
\label{fig:prose_zo_ucb_m11_mean_reward}
\end{figure}

\begin{figure}
\includegraphics[scale=0.4]{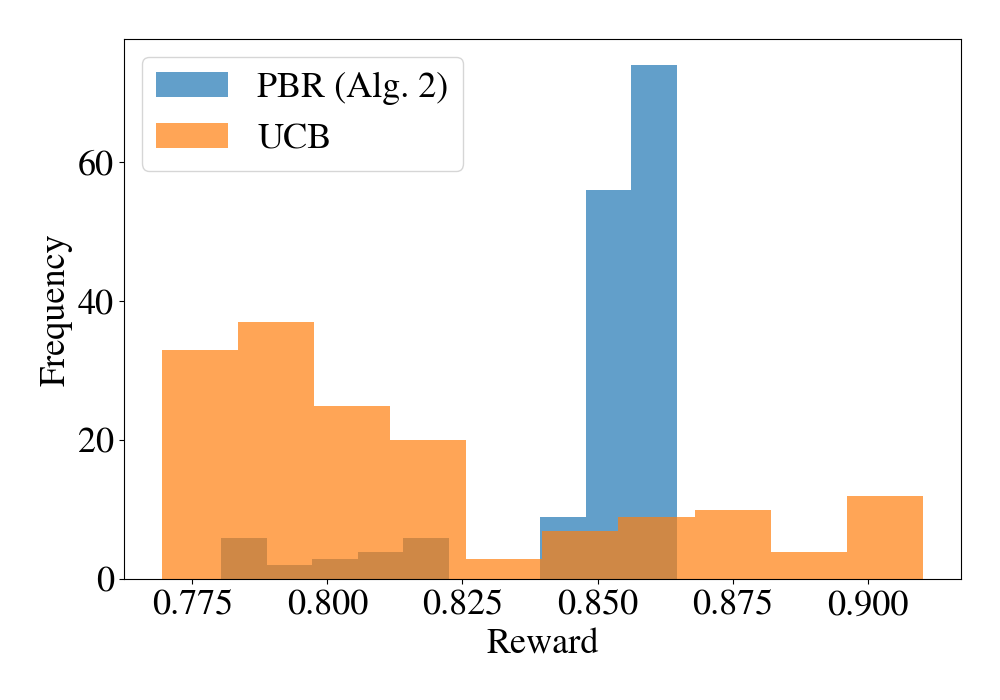}
\caption{Distribution of rewards received by \selftune{} and UCB when tuning \texttt{RegexPair} parameters (Figure~\ref{fig:code-regexpair}), corresponding to the result in Figure~\ref{fig:prose_zo_ucb_m22_mean_reward} (a).}
\label{fig:prose_zo_ucb_m22_hist}
\end{figure}

Here, we include additional results for the PROSE case study presented in the main text, Section~\ref{sec:prose}, in Figures~\ref{fig:prose_zo_ucb_m11_mean_reward},~\ref{fig:prose_zo_ucb_m22_hist} and~\ref{fig:prose_large_zo_ucb_hist}. Figure~\ref{fig:code-formatdatetimerange} shows the \texttt{FormatDateTimeRange} heuristic that is part of \prose~\cite{prosesdk,pmlr-v89-natarajan19a}, with parameters of interest highlighted.

\begin{figure} 
	\begin{lstlisting}[escapechar=!,language=csh,morekeywords={public,static,double,return}]
double FormatDateTimeRange(double base_SeparatorIsCommonDateTimeSeparator, ...) {
    return  !\colorbox{yellow}{-0.25}!*base_SeparatorIsCommonDateTimeSeparator +
        !\colorbox{yellow}{0.07}!*base_SeparatorIsOnlySymbolsAndPunctuation +
        !\colorbox{yellow}{-0.52}!*base_SeparatorIsOnlyWhitespace +
        !\colorbox{yellow}{-0.12}!*base_SeparatorIsWrappedByWhitespace +
        !\colorbox{yellow}{0.06}!*bias_FormatDateTimeRange +
        !\colorbox{yellow}{1}!*score_FormatDateTimeRange_dtRoundingSpec +
        !\colorbox{yellow}{1}!*score_FormatDateTimeRange_dtRoundingSpec2 +
        !\colorbox{yellow}{1}!*score_FormatDateTimeRange_inputDateTime +
        !\colorbox{yellow}{1}!*score_FormatDateTimeRange_outputDtFormat +
        !\colorbox{yellow}{1}!*score_FormatDateTimeRange_s;
}
    \end{lstlisting}
	\caption{The \texttt{FormatDateTimeRange} heuristic in \prose~\cite{prosesdk,pmlr-v89-natarajan19a} with parameters of interest highlighted.}
	\label{fig:code-formatdatetimerange}
\end{figure}

% \begin{figure}
% \includegraphics[scale=0.4]{exps_imgs/prose_zo_sc0p5_vs_prose_ucb_m11_hist.png}
% \caption{Distribution of rewards received by \selftune{} and UCB when tuning \texttt{RegexPair} parameters (Figure~\ref{fig:code-regexpair}), corresponding to the result in Figure~\ref{fig:prose_zo_ucb_m11_mean_reward}.}
% \label{fig:prose_zo_ucb_m11_hist}
% \end{figure}

\begin{figure}
\includegraphics[scale=0.4]{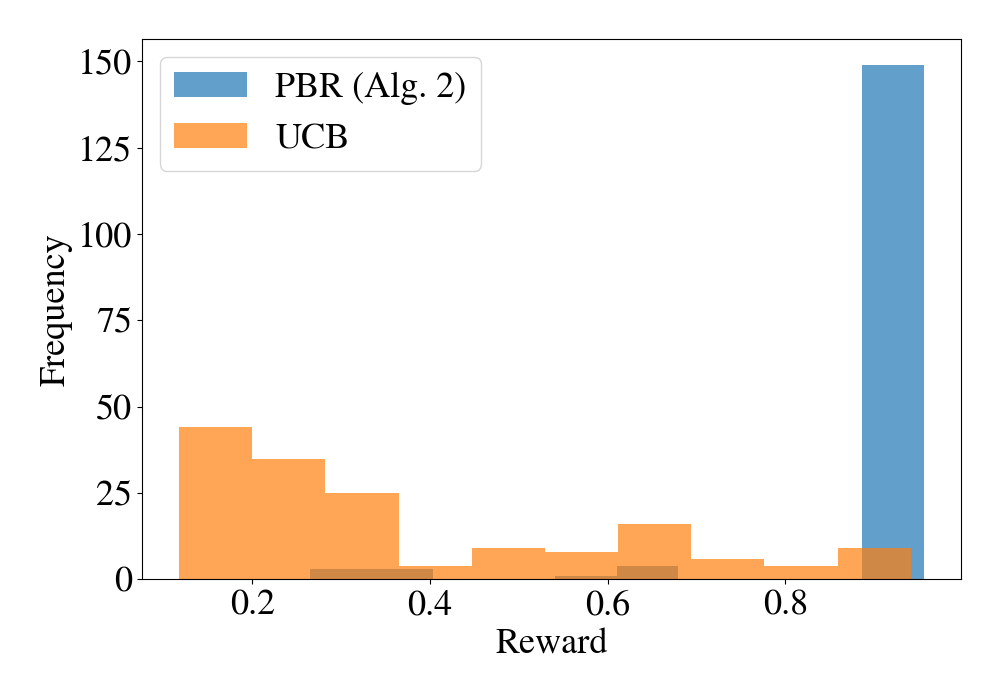}
\caption{Distribution of rewards received by \selftune{} and UCB when tuning \texttt{FormatDateTimeRange} (Figure~\ref{fig:code-formatdatetimerange}) parameters, corresponding to the result shown in Figure~\ref{fig:prose_zo_ucb_m22_mean_reward} (b).}
\label{fig:prose_large_zo_ucb_hist}
\end{figure}

\subsection{Sketch details (for Section~\ref{sec:evaluation_sketch})}
\label{app:evaluation:sketch}
Example code for~\sketch{} specification given as input to the tool~\cite{sketchtool} is shown in Figure~\ref{fig:code-sketch}. We vary \texttt{n, d, features, w\_opt} values in this code to produce the results presented in Section~\ref{sec:evaluation_sketch}.

\textbf{Relaxing Sketch specification.} We know that there is a solution to problem~\eqref{eqn:toyproblem} that achieves the objective value of 0. But, in some cases, it is still useful to fill the holes with values that yield a small enough objective value ``close to'' the optimal --- i.e. \textit{fill the holes $\weights$ so that $L(\weights) \leq L(\weights^*) + \epsilon = \epsilon$}, where $L(\weights) := \mathbb{E}[r\big( \weights \cdot \featvals \big)]$. Thus we could relax the sketch specification, i.e. last line 5 of Figure~\ref{fig:code-sketch}, to \texttt{assert (w[0] <= 10) \& (w[1] <= 10) \& (losses <= epsilon)}. As we increase $\epsilon$ away from 0, the search problem becomes easier, and therefore one expects to find \textit{some} feasible solution, though it need not be optimal. Unfortunately, this intuition doesn't hold for sketching --- in some cases, it makes the computation even longer, because it takes more bits to encode the slackness constraint and ends up increasing the problem complexity. On the other hand, the algorithms used by~\selftune{} can yield a solution, close to the optimal, that improves with the time budget.

\begin{figure}[h]
	\begin{lstlisting}[language=dsl,morekeywords={public,static,double,return}]
int sqloss (int score1, int score2) {
    return (score1 - score2) * (score1 - score2);
}

int absloss (int score1, int score2) {
    return score1 > score2 ? score1 - score2: 
                             score2 - score1;
}

harness void paramSketch () {
    int n = 2;
    int d = 2; 
    int[n][d] features = {{-2,3}, {-3,-1}};
    int[d] w_opt = {1,2};
    int[d] w;
    int losses = 0;
    int[n] y;
    for (int i=0; i < n; i++) {
        y[i] = 0;
        for (int j=0; j < d; j++) {
            y[i] += w_opt[j] * features[i][j];
        }
    }            
    w[0] = ??;
    w[1] = ??;
    for (int i=0; i < n; i++) {
        int score = 0;
        for (int j=0; j < d; j++) {
            score += w[j] * features[i][j];
        }
        losses += absloss(score, y[i]);
    }
    minimize (losses);
    assert (w[0] <= 10) & (w[1] <= 10) & (losses == 0);
}
    \end{lstlisting}
	\caption{Example implementation of a sketch problem given to the \sketch{} tool~\cite{sketchtool} discussed in Section~\ref{sec:evaluation_sketch}. Here, number of examples $n = 2$ and number of holes to fill is $d = 2$. The values for \texttt{features} are random integers sampled from the range [-10,10], and those for \texttt{w\_opt} are random integers sampled from the range [0,10]. } %The rule-based implementations can be challenging to engineer and tune from scratch, and can quickly become obsolete in light of new data, changes to software such as feature addition. }
	\label{fig:code-sketch}
\end{figure}

% \section{Appendix: Evaluation}
% \label{app:evaluation}

\subsection{Fermat sketch (for Section~\ref{sec:whitebox})}
\label{app:fermat-sketch}
In Section~\ref{sec:whitebox} (Figure~\ref{fig:sketch-thermostat}) we have given the \Thermostat sketch;
Figure~\ref{fig:sketch-aircraft} shows the \Aircraft sketch, the other function sketch discussed in Section~\ref{sec:whitebox}.
Both of the sketches originate from~\citet{chaudhuri2014bridging}, where they are also explained in more detail.

\begin{figure}[ht]
\begin{lstlisting}[escapechar=`,language=csh,morekeywords={public,double,return,bool,assert,??},numbers=left,basicstyle=\footnotesize\ttfamily,xleftmargin=5.0ex]
double Aircraft(double v1, double v2) {
    `$\dots$`
    double criticalDist = ??(6, 9);
    double safetyDist = 3.0;
    double delay = ??(10, 15);
    double delay2 = ??(9, 14);
    assert(delay > 0.0; `$\theta$`);
    assert(delay2 > 0.0; `$\theta$`);
    assert(criticalDist > safetyDisty; `$\theta$`);
    assert(criticalDist < 10; `$\theta$`);
    for (int i=0; i<50; i=i+1) {
        if (stage == CRUISE) {
            move_straight(x1, y1, x2, y2, v1, v2);
            if (`$\delta$`(x1, y1, x2, y2) < criticalDist) {
                stage = LEFT;
                assert(!haveLooped; `$\theta$`);
                steps = 0;
            }
        }
        if (stage == LEFT) {
            move_left(x1, y1, x2, y2, v1, v2);
            steps = steps + 1;
            if (delay - steps < 0) {
                stage = STRAIGHT;
                steps = 0;
            }
        }
        if (stage == STRAIGHT) {
            move_straight(x1, y1, x2, y2, v1, v2);
            steps = steps + 1;
            if (delay - steps < 0) {
                stage = RIGHT;
                steps = 0;
            }
        }
        if (stage == RIGHT) {
            move_right(x1, y1, x2, y2, v1, v2);
            steps = steps + 1;
            if (delay - steps < 0) {
                stage = CRUISE;
                haveLooped = true;
            }
        }
        assert(`$\delta$`(x1, y1, x2, y2) < safetyDist; `$\theta$`);
    }
    return 2 * delay + delay2;
}
\end{lstlisting}
\caption{\Aircraft sketch~\cite{chaudhuri2014bridging}.}
\label{fig:sketch-aircraft}
\end{figure}

\subsection{Decision tree problems~\textsc{Xor} and \textsc{Slates} (for Section~\ref{sec:structure})}
\begin{figure} [h!]
\includegraphics[scale=0.45]{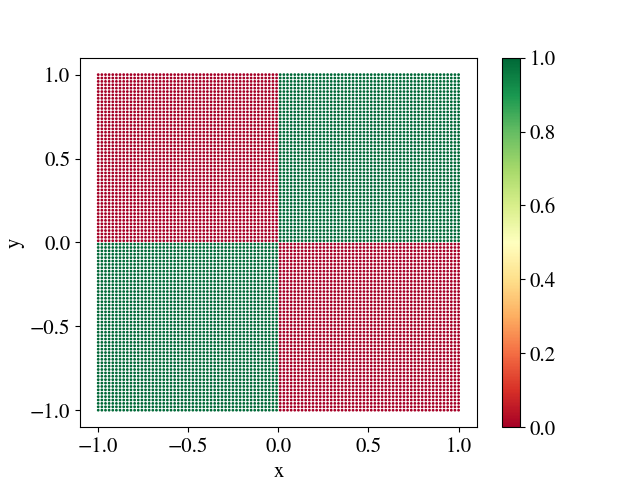}
\caption{(\textsc{Xor} problem) Expected values as a function of the two features \texttt{x} and \texttt{y} is given by the~\texttt{XOR} function, that can be modeled by a tree of height 2. Best viewed in color.}
\label{fig:xor.png}
\end{figure}

\begin{figure} [h!]
\includegraphics[scale=0.45]{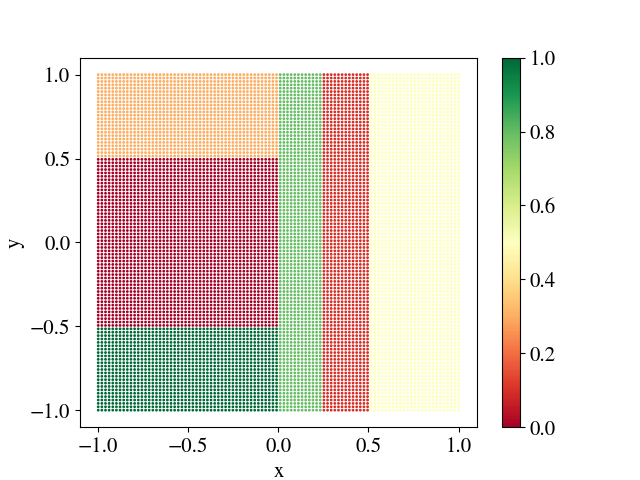}
\caption{(\textsc{Slates} problem) Expected values (thresholds) as a function of the two features \texttt{x} and \texttt{y} is given by a piece-wise constant function, that can be modeled by a tree of height 3. Best viewed in color.}
\label{fig:slates.png}
\end{figure}

The~\textsc{Xor} problem instance discussed in the main paper is presented in Figure~\ref{fig:xor.png}. \textsc{Slates} is a more complex hypothetical tuning problem, where we wish to learn a piece-wise constant threshold function based on the values of 2 features, to decide what the threshold for the input should be (these type of heuristics are common in systems); here, we have 6 different possible thresholds that depend on the 2 features as shown in Figure \ref{fig:slates.png}. The decision functions learnt by our Algorithm~\ref{alg:tree} for the~\textsc{Xor} and the~\textsc{Slates} problem instances are given in Figures~\ref{fig:code-xor} and~\ref{fig:code-slates}.
%As done previously in Section~\ref{sec:learning_tree_models} we compare our tree learning Algorithm \ref{alg:tree} against Algorithm \ref{alg:constant} which treats the tree parameters as constants, in Figure \ref{fig:slates_smart_vs_naive.png}.

\begin{figure} [h!]
	\begin{lstlisting}[escapechar=!,language=csh,morekeywords={public,static,double,return}]
    double XOR(double x1, double x2) 
    {
        if (x2 <= -0.02)
            if (x1 >= 0.04) 
                return 0; 
            else return 1;
        else
            if (x1 <= (-0.01 * x2 - 0.03)) 
                return 0.03; 
            else return 0.98;
    } 
\end{lstlisting}
	\caption{The tree heuristic learnt by \selftune{} for the \textsc{Xor} problem in Figure~\ref{fig:xor.png}.}
	\label{fig:code-xor}
\end{figure}

\begin{figure} [h!]
	\begin{lstlisting}[escapechar=!,language=csh,morekeywords={public,static,double,return}]
 double ThresholdMap(double x, double y) {
    if (x >= -0.01*y)
        if (x >= 0.25)
            if (x >= 0.41) return 0.5;
            else return 0.1;
        else
            return 0.81;
    else
        if (y >= -0.01*x - 0.41)
            if (y >= 0.01*x + 0.43) return 0.3;
            else return 0;
        else
            if (0.94*x >= 0.34*y + 2.58) return 0.47;
            else return 1;
}
    \end{lstlisting}
	\caption{The tree heuristic learnt by \selftune{} for deciding the threshold parameter described in Figure~\ref{fig:slates.png}.}
	\label{fig:code-slates}
\end{figure}
\section{Appendix: \selftune~Specification API}
\label{sec:API}
%Give the syntax and semantics of the SelfTune operator API
%At the time of creation of a \selftune\ problem instance, the programmer specifies \textit{what} parameters needs to be learned or tuned. 
%For example, in Table 1, Section 2, we want \selftune\ to learn a linear function with two 
%parameters.\\
Here we give the details of the API we have built for developers to use~\selftune and our algorithms.\\  
\noindent
\textbf{1. Creation.} The \Create\ API creates an instance of the synthesis/parameter
learning problem for \selftune. This API allows 
several optional arguments that essentially helps the programmer encode domain knowledge for the problem: \\
(a) the name of the decision function to learn, \\
(b) optional features, which the programmer uses to model the decision as a function of the context,\\ 
%\jg{Why only these three model types? }
(c) the model type (``template'') for computing decisions, as a function of feature values and parameter values (we currently support constant or linear model or decision tree as discussed in the main paper),\\
%\jg{Why don't we show this in Table 1?}
(d) optional initial values for the parameters, set to 0 by default \\
(e) optional constraints on the parameters to be tuned; the API supports range constraints (\texttt{min} and \texttt{max}), and type constraints (e.g., \texttt{isInt} is \textsc{True} if the parameters takes only integral values).
%\jg{This next piece of code comes a bit out of thin air -- should we delete it?}

\texttt{Constraints(\keywordc{double} min, \keywordc{double} max, \keywordc{bool} isInt)}

\texttt{\keywordc{int} \stc{\textbf{Create}}(\keywordc{string} param,} 

$\ \ \ \ \ $ \texttt{\keywordc{double} initValue = 0,} 

$\ \ \ \ \ $  \texttt{\keywordc{Dictionary}<\keywordc{string}, Constraints> constraints = null}, 

$\ \ \ \ \ $  \texttt{\keywordc{string[]} features = null}, 

$\ \ \ \ \ $  \texttt{\keywordc{enum} template})
\\

\noindent
\textbf{2. Connection.} 
%\jg{What is global about that data store? Should we drop global?}
The \Create\ API sets up a data store instance in the back-end for tuning the specified parameters, initializes the necessary background services to maintain/update this store. A unique identifier to this store instance is returned by the call to~\texttt{Create}.
The next step is to connect the client to this instance. 
The \Connect\ API connects
a parameter learning instance to a \selftune\ object.

\texttt{\keywordc{void}   
\stc{\textbf{Connect}} (\keywordc{int} problemId)
}
\\
Note that if a store already exists (for the parameter(s) of interest), then the client can directly connect to the instance by referencing the unique identifier to the instance, because  store instances are persistent. This also enables multiple clients (distributed spatially and/or temporally) to query the latest decisions for as well as give feedback to the same learning problem.  The creation of a store instance can also be performed through a separate GUI or a plugin.
\\

\noindent
\textbf{3. Prediction.} This interface encapsulates~\decfun~introduced in Example~\ref{ex:pbr}. With the \Predict~interface, the programmer can query the decision outcome using the learnt model. It takes the feature values as input from the context and returns the decision outcome:

\texttt{\keywordc{(int, double)} \stc{\textbf{Predict}}} (\texttt{\keywordc{Dictionary}<\keywordc{string}, \keywordc{double}> features})

Internally, it works by (1) retrieving the current version of the learnt model and parameters as an expression tree, (2) converting the expression tree into a lambda expression, and (3) running the lambda expression using the given feature values as arguments. 
The lambda expression is cached in the client as as to perform further predictions without needing to contact the server. If the model is refreshed, then the cache is cleared in the client, and the subsequent call to \Predict\ retrieves the updated expression from the model and recomputes the corresponding lambda expression.
Note that \Predict{} returns a pair of values -- a unique identifier which identifies the particular invocation of \Predict{}, and the predicted decision value.

If the client wishes to retrieve the expression tree to inspect the learnt model, it can use the \GetExprTree\ API:

\texttt{\keywordc{Expression<TDelegate>} \stc{\textbf{GetExprTree}}} (\texttt{\keywordc{Dictionary}<\keywordc{string}, \keywordc{double}> features})
\\
\noindent \textbf{4. Reward.} This is the \assignr~interface introduced in Example~\ref{ex:pbr}, which allows the client to specify a reward and also associate
it with a particular invocation of \Predict{} (the default is the last invocation, as in Figure~\ref{fig:synthex}).

\texttt{\keywordc{void} \stc{\textbf{AssignReward}}(\keywordc{int} invocationId, \keywordc{double} reward)}\\
We note that a decision is associated with many parameters (from a linear model or decision tree) and the reward specified is associated with all of these parameter choices (see Figure~\ref{fig:raw}). This aspect of \selftune\ distinguishes it from other solutions like~Decision Service~\cite{agarwal2016multiworld} and \smartchoices~\cite{carbune2018smartchoices}; they rely on providing explicit reward for each parameter and decision value, whereas we do not require that disambiguation --- our problem formulation and learning enables working with a single reward (i.e. value of the key system level metric) for a decision (and can be generalized to set of decisions as well), and in turn tuning several parameters together.

% As mentioned earlier, in some settings, the reward computation and passing it to the backend may have to happen asynchronously and there may not be a natural place in the code to call~\AssignReward. To this end, \selftune\ runs a service in the background (on the client, where the underlying software is also running) at a determined frequency to compute rewards and push them into the store. The service is initialized at the time of \Connect{} on the \selftune\ instance. Reward tracker runs in loop (at a frequency that can optionally be configured by the developer), invoking the developer-supplied implementation of \texttt{Reward} interface, and updating the reward in the data store against the particular problem instance.
\noindent
\textbf{5. Refresh.}  The \Refresh\ API learns an updated
model based on the data collected in the store instance, using the algorithms presented in Section~\ref{sec:learning}:
\texttt{\keywordc{void} \stc{\textbf{Refresh}()}}

The client can choose to perform either online learning or offline learning by how frequently it calls \Refresh.
Calling \Refresh\, say once a day, results in offline learning, and calling it every few minutes results in online learning. As a side-effect, \Refresh\ clears the client side cache used by \Predict, as discussed above.

\end{document}

%%
%% End of file `sample-sigconf.tex'.